\documentclass[acmtog, nonacm]{acmart}
\AtBeginDocument{%
  }

\acmSubmissionID{Siggraph 2025 Papers 770 }

\usepackage{bm}
\usepackage{multirow}
\usepackage{subcaption}
\usepackage{tikz}
\usetikzlibrary{spy,calc}
\usepackage{xcolor}
\usepackage{stmaryrd}
\usepackage{wrapfig}
\usepackage{mathtools}
\usepackage{makecell}

\newif\ifblackandwhitecycle
\gdef\patternnumber{0}

\pgfkeys{/tikz/.cd,
    zoombox paths/.style={
        draw=orange,
        very thick
    },
    black and white/.is choice,
    black and white/.default=static,
    black and white/static/.style={ 
        draw=white,   
        zoombox paths/.append style={
            draw=white,
            postaction={
                draw=black,
                loosely dashed
            }
        }
    },
    black and white/static/.code={
        \gdef\patternnumber{1}
    },
    black and white/cycle/.code={
        \blackandwhitecycletrue
        \gdef\patternnumber{1}
    },
    black and white pattern/.is choice,
    black and white pattern/0/.style={},
    black and white pattern/1/.style={    
            draw=white,
            postaction={
                draw=black,
                dash pattern=on 2pt off 2pt
            }
    },
    black and white pattern/2/.style={    
            draw=white,
            postaction={
                draw=black,
                dash pattern=on 4pt off 4pt
            }
    },
    black and white pattern/3/.style={    
            draw=white,
            postaction={
                draw=black,
                dash pattern=on 4pt off 4pt on 1pt off 4pt
            }
    },
    black and white pattern/4/.style={    
            draw=white,
            postaction={
                draw=black,
                dash pattern=on 4pt off 2pt on 2 pt off 2pt on 2 pt off 2pt
            }
    },
    zoomboxarray inner gap/.initial=5pt,
    zoomboxarray columns/.initial=2,
    zoomboxarray rows/.initial=2,
    subfigurename/.initial={},
    figurename/.initial={zoombox},
    zoomboxarray/.style={
        execute at begin picture={
            \begin{scope}[
                spy using outlines={%
                    zoombox paths,
                    width=\imagewidth / \pgfkeysvalueof{/tikz/zoomboxarray columns} - (\pgfkeysvalueof{/tikz/zoomboxarray columns} - 1) / \pgfkeysvalueof{/tikz/zoomboxarray columns} * \pgfkeysvalueof{/tikz/zoomboxarray inner gap} -\pgflinewidth,
                    height=(\imageheight ) / \pgfkeysvalueof{/tikz/zoomboxarray rows} - (\pgfkeysvalueof{/tikz/zoomboxarray rows} - 1) / \pgfkeysvalueof{/tikz/zoomboxarray rows} * \pgfkeysvalueof{/tikz/zoomboxarray inner gap}-\pgflinewidth,
                    magnification=3,
                    every spy on node/.style={
                        zoombox paths
                    },
                    every spy in node/.style={
                        zoombox paths
                    }
                }
            ]
        },
        execute at end picture={
            \end{scope}
            \gdef\patternnumber{0}
        },
        spymargin/.initial=0.5em,
        zoomboxes xshift/.initial=1,
        zoomboxes right/.code=\pgfkeys{/tikz/zoomboxes xshift=1},
        zoomboxes left/.code=\pgfkeys{/tikz/zoomboxes xshift=-1},
        zoomboxes yshift/.initial=0,
        zoomboxes above/.code={
            \pgfkeys{/tikz/zoomboxes yshift=0.87},
            \pgfkeys{/tikz/zoomboxes xshift=0}
        },
        zoomboxes below/.code={
            \pgfkeys{/tikz/zoomboxes yshift=-0.87},
            \pgfkeys{/tikz/zoomboxes xshift=0}
        },
        caption margin/.initial=4ex,
    },
    adjust caption spacing/.code={},
    image container/.style={
        inner sep=0pt,
        at=(image.north),
        anchor=north,
        adjust caption spacing
    },
    zoomboxes container/.style={
        inner sep=0pt,
        at=(image.north),
        anchor=north,
        name=zoomboxes container,
        xshift=\pgfkeysvalueof{/tikz/zoomboxes xshift}*(\imagewidth+\pgfkeysvalueof{/tikz/spymargin}),
        yshift=\pgfkeysvalueof{/tikz/zoomboxes yshift}*((\imageheight )+\pgfkeysvalueof{/tikz/spymargin}+\pgfkeysvalueof{/tikz/caption margin}),
        adjust caption spacing
    },
    calculate dimensions/.code={
        \pgfpointdiff{\pgfpointanchor{image}{south west} }{\pgfpointanchor{image}{north east} }
        \pgfgetlastxy{\imagewidth}{\imageheight}
        \global\let\imagewidth=\imagewidth
        \global\let\imageheight=\imageheight
        \gdef\columncount{1}
        \gdef\rowcount{1}
        
    },
    image node/.style={
        inner sep=0pt,
        name=image,
        anchor=south west,
        append after command={
            [calculate dimensions]
            node [image container,subfigurename=\pgfkeysvalueof{/tikz/figurename}-image] {\phantomimage}
            node [zoomboxes container,subfigurename=\pgfkeysvalueof{/tikz/figurename}-zoom] {\phantomimage}
        }
    },
    color code/.style={
        zoombox paths/.append style={draw=#1}
    },
    connect zoomboxes/.style={
    spy connection path={\draw[draw=none,zoombox paths] (tikzspyonnode) -- (tikzspyinnode);}
    },
    help grid code/.code={
        \begin{scope}[
                x={(image.south east)},
                y={(image.north west)},
                font=\footnotesize,
                help lines,
                overlay
            ]
            \foreach \x in {0,1,...,9} { 
                \draw(\x/10,0) -- (\x/10,1);
                \node [anchor=north] at (\x/10,0) {0.\x};
            }
            \foreach \y in {0,1,...,9} {
                \draw(0,\y/10) -- (1,\y/10);                        \node [anchor=east] at (0,\y/10) {0.\y};
            }
        \end{scope}    
    },
    help grid/.style={
        append after command={
            [help grid code]
        }
    },
}

\newcommand\phantomimage{%
    \phantom{%
        \rule{\imagewidth}{\imageheight}%
    }%
}
\newcommand\zoombox[2][]{
    \begin{scope}[zoombox paths]
        \pgfmathsetmacro\xpos{
            (\columncount-1)*(\imagewidth / \pgfkeysvalueof{/tikz/zoomboxarray columns} + \pgfkeysvalueof{/tikz/zoomboxarray inner gap} / \pgfkeysvalueof{/tikz/zoomboxarray columns} ) + \pgflinewidth
        }
        \pgfmathsetmacro\ypos{
            (\rowcount-1)*( \imageheight / \pgfkeysvalueof{/tikz/zoomboxarray rows} + \pgfkeysvalueof{/tikz/zoomboxarray inner gap} / \pgfkeysvalueof{/tikz/zoomboxarray rows} ) + 0.5*\pgflinewidth
        }
        \edef\dospy{\noexpand\spy [
            #1,
            zoombox paths/.append style={
                black and white pattern=\patternnumber
            },
            every spy on node/.append style={#1},
            x=\imagewidth,
            y=\imageheight
        ] on (#2) in node [anchor=north west] at ($(zoomboxes container.north west)+(\xpos pt,-\ypos pt)$);}
        \dospy
        \pgfmathtruncatemacro\pgfmathresult{ifthenelse(\columncount==\pgfkeysvalueof{/tikz/zoomboxarray columns},\rowcount+1,\rowcount)}
        \global\let\rowcount=\pgfmathresult
        \pgfmathtruncatemacro\pgfmathresult{ifthenelse(\columncount==\pgfkeysvalueof{/tikz/zoomboxarray columns},1,\columncount+1)}
        \global\let\columncount=\pgfmathresult
        \ifblackandwhitecycle
            \pgfmathtruncatemacro{\newpatternnumber}{\patternnumber+1}
            \global\edef\patternnumber{\newpatternnumber}
        \fi
    \end{scope}
}

\definecolor{tabfirst}{rgb}{0.96, 0.77, 0.77} %
\definecolor{tabsecond}{rgb}{0.98 , 0.93, 0.77} %
\definecolor{tabthird}{rgb}{1, 1, 0.7} %
\definecolor{lime}{rgb}{0.75, 1.0, 0.0}
\definecolor{theirs}{HTML}{FDAE61}
\definecolor{ours}{HTML}{2B83BA}

\newcommand{\sect}[1]{Section~\ref{#1}}

\newif\ifconfidential
\confidentialtrue

\newif\ifdraft
\draftfalse
\ifdraft

\newcommand{\hl}[1]{\textcolor{ForestGreen}{{#1}}}
\newenvironment{hlbreakable}%
  {\color{ForestGreen}}%
  {}

\else

\newcommand{\hl}[1]{\textcolor{black}{{#1}}}
\newenvironment{hlbreakable}%
  {\color{black}}%
  {}

\newcommand{\todo}[1]{}
\newcommand{\TODO}[1]{}
\newcommand{\amirc}[1]{}
\newcommand{\ericc}[1]{}
\newcommand{\nadavc}[1]{}
\newcommand{\arikc}[1]{}
\newcommand{\yedidc}[1]{}
\newcommand{\alexc}[1]{}

\fi

\newcommand{\rgbx}{RGB $\leftrightarrow$ X}

\newcommand{\ourmethod}{LightLab} %
\newcommand{\handcrafted}{Org.}
\newcommand{\procedural}{Rand.}
\newcommand{\synthetic}{Synth.}
\newcommand{\myvspace}[1]{}

\newcommand{\ourpapertitle}{LightLab: Controlling Light Sources in Images with Diffusion Models}

\copyrightyear{2025}
\acmYear{2025}
\setcopyright{cc}
\setcctype{by}
\acmConference[SIGGRAPH Conference Papers '25]{Special Interest Group on Computer Graphics and Interactive Techniques Conference Conference Papers }{August 10--14, 2025}{Vancouver, BC, Canada}
\acmBooktitle{Special Interest Group on Computer Graphics and Interactive Techniques Conference Conference Papers (SIGGRAPH Conference Papers '25), August 10--14, 2025, Vancouver, BC, Canada}
\acmDOI{10.1145/3721238.3730696}
\acmISBN{979-8-4007-1540-2/2025/08}
\begin{document}

\title{\ourpapertitle}

\author{Nadav Magar}
\affiliation{
 \institution{Tel Aviv University and Google}
 \country{Israel}}
\email{nadavm243@gmail.com}

\author{Amir Hertz}
\affiliation{
 \institution{Google}
 \country{United States of America}}
\email{hertzamir@gmail.com}

\author{Eric Tabellion}
\affiliation{
 \institution{Google}
 \country{United States of America}}
\email{etabellion@google.com}

\author{Yael Pritch}
\affiliation{
 \institution{Google}
 \country{Israel}}
\email{yaelp@google.com}

\author{Alex Rav-Acha}
\affiliation{
 \institution{Google}
 \country{Israel}}
\email{ravacha@google.com}

\author{Ariel Shamir}
\authornote{Equal advising.}
\affiliation{
 \institution{Reichman University and Google}
 \country{Israel}}
\email{arik@runi.ac.il}

\author{Yedid Hoshen}
\authornotemark[1]
\affiliation{
 \institution{Hebrew University of Jerusalem and Google}
 \country{Israel}}
\email{yedid.hoshen@mail.huji.ac.il}


\begin{teaserfigure}
    \small
    \centering
     \includegraphics[width=1.\linewidth]{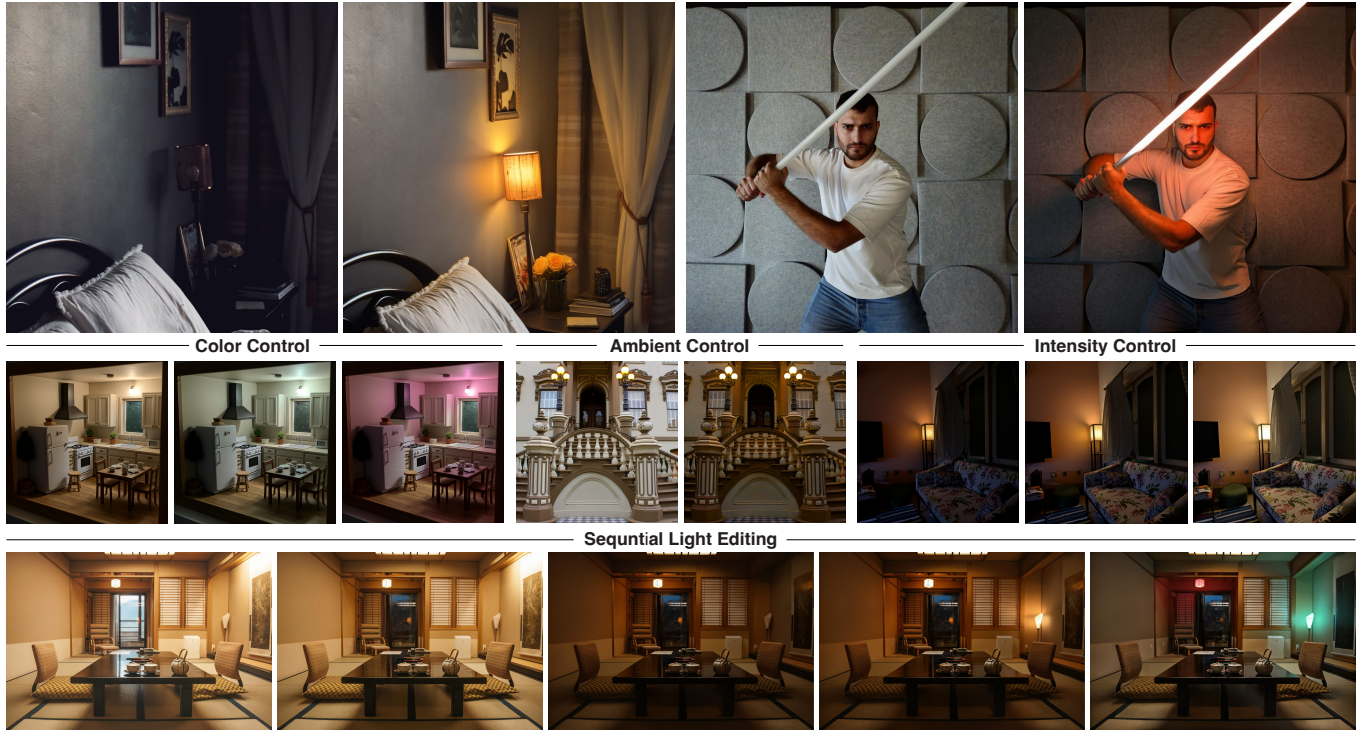}
    \caption{
    \textbf{Light editing results of \ourmethod.} Our method enables explicit parametric control over light sources in an image, while producing physically plausible shadows and environmental effects (top). The method can manipulate the intensity of light sources, change their color, and adjust ambient illumination (middle). \ourmethod{} can be used for intricate sequential editing of lighting in images (bottom from left to right): starting with an input image, turning off even outside light coming from a window, turning off an inside light source, turning on another light source, and changing lights colors. Note how reflections and shadows are plausibly handled in all these cases (please zoom in for better view and examine the tabletop).}
    \myvspace{+0.5em}
    \label{fig:teaser}
\end{teaserfigure}

\begin{abstract}
We present a simple, yet effective diffusion-based method for fine-grained, parametric control over light sources in an image.
Existing relighting methods either rely on multiple input views to perform inverse rendering at inference time, or fail to provide explicit control over light changes.
Our method \hl{fine-tunes a diffusion model on} a small set of real raw photograph pairs, supplemented by synthetically rendered images at scale, to elicit its photorealistic prior for the relighting task.
We leverage the \hl{linearity of light} to synthesize image pairs depicting controlled light changes of either a target light source or ambient illumination. 
Using this data and an appropriate fine-tuning scheme, \hl{we train a model for} precise illumination changes with explicit control over light intensity and color.
Lastly, we show how our method can achieve compelling light editing results, \hl{and outperforms existing methods based on user preference.}

\end{abstract}
\maketitle

\section{Introduction}
\label{sec:introduction}

A simple flick of a light switch can transform the appearance of an image. 
Turning on a light does not only increase an image's brightness, but can also add depth, contrast, and shift a viewer's focus.
Similarly, changing the temperature of a light source can set the mood of an image, warm light creates a comfortable and inviting atmosphere, whereas cool light induces a cold and pristine feeling.
While turning a light on is easy at the time of capturing a photo, doing so after capture is far more challenging.

Traditionally, 3D graphics methods model scene geometry and intrinsics from multiple input captures, then simulate new lighting conditions with a physical illumination model.
These methods enable precise and explicit control over light sources, however recovering a 3D model from a single image is an ill-posed problem, and often fails to produce appealing results. 
Recently, diffusion-based image editing methods have been applied to various relighting tasks, utilizing their strong statistical prior to bypass the need for an explicit physical model of the image.
However, the stochastic nature of diffusion models, and their common reliance on textual conditioning, makes fine-grained parametric control of physical image properties challenging.

In this work, we present \ourmethod, a diffusion-based method for explicit, parametric control over light sources in an image. 
Specifically, we target two key properties of a light source: its intensity and color. 
To complement these abilities, we also enable control over the scene's ambient illumination, and over the resulting tone mapping effects.
Combined, these capabilities provide end users a set of rich light editing tools, and allow control over the look and feel of an image, by means of manipulating its illumination.

Editing the illumination in an image presents unique challenges. 
Firstly, creating physically plausible re-lit images requires an understanding of light transport.
The distribution of light in a scene is highly dependent on its geometry, material properties, and light source characteristics, making relighting an extremely ill-conditioned problem.
A second challenge arises in preserving intrinsic scene properties, such as geometry, diffuse and specular reflectance, and roughness, while changing illumination. 
These properties are intertwined in the image pixels, and disentangling them is a complex task, even when solved explicitly. 
Furthermore, editing illumination in Standard Dynamic Range (SDR), requires also disentangling changes to the physical scene from changes originating from the tone-mapping and imaging pipeline.

To overcome these challenges, we take a data-centric approach, using paired images representing the complex illumination changes induced by turning on a visible light source (\sect{sec:datasets}).
As acquiring a large scale dataset of paired examples depicting a diverse set of scenes and light configurations is infeasible, we propose a combination of three solutions.
First, we capture a small set of photograph pairs, which depict a scene while a light source is switched on or off.
Next, we use physically-based rendering of 3D scenes \cite{pharr2023physically} to create a large number of synthetic pairs, while also switching on and off several visible light sources and producing several ambient lighting configurations. To further increase the diversity of the data, we also procedurally place light sources at random locations in the scene.
Finally, we use the \hl{linearity of light} (\sect{sec:post-process}) and a suitable tone-mapping strategy (\sect{sec:tonemapping}) to synthesize image pairs for the full range of light source intensity.

Although simulated pairs cause a domain gap too large for training on their own, we show that combining \hl{a smaller set of} real photograph pairs \hl{mitigates the resulting domain drift}.
This combination mitigates the physical limitations of data capture, while allowing fine-grained control over light sources. Given such data, we fine-tune a conditional diffusion model (\sect{sec:diffusion}, \sect{sec:implementation_details}) to learn to edit the color and intensity of the light sources.%

\hl{We demonstrate our method on images that contain visible light sources, focusing on indoor settings, with additional results on outdoor images (Figure D.9), and out-of-domain examples (Figure~\ref{fig:more_results}).}
Further comparisons with other related works (\sect{sec:comparisons}) show that our method is the first to enable high-quality, precise control over visible local light sources, as confirmed by a user study.

We present several applications of our method (\sect{sec:applications}), including control over intensity and color of light sources in a photograph, applying a series of consecutive edits, changing illumination in non-photorealistic images, and generating consistent lighting across multiple animation frames.

In summary, the main contributions of our paper are:

\begin{hlbreakable}
\begin{itemize}
    \item A method for effectively fine-tuning and conditioning a diffusion model for parametric control over light sources from a single image.
    \item We show that even a small set of real, physically accurate training examples can complement large-scale synthetic rendering, where the former prevents domain drift while the latter improves physical plausibility.
    \item A resulting high-quality image-based solution that enables users to make complex and consecutive illumination edits. 
\end{itemize}
\end{hlbreakable}

\begin{figure*}[t!]
  \centering
  \includegraphics[width=0.92\linewidth]{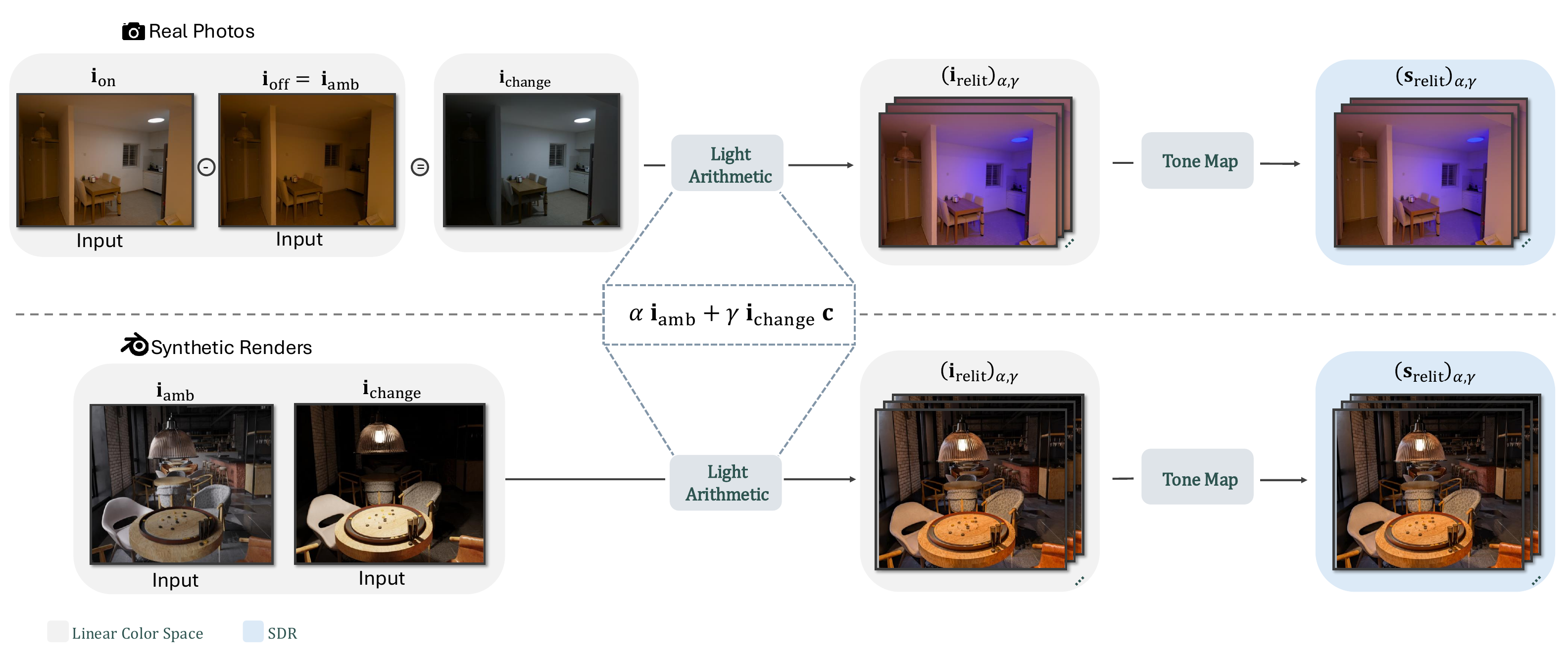}
  \caption{\textbf{Post processing pipeline.} \textbf{Top row.} From a pair of real (raw) photograph pairs, we first isolate the target light change $\textbf{i}_\text{change}$. \textbf{Bottom row.} For synthetic data, we render each light component separately. After light disentanglement both domains undergo light arithmetic to create parameterized sequences of images $\textbf{i}_\text{relit}\left( \alpha, \gamma, \textbf{c}_\text{t} \right)$, which are later tone mapped to SDR (either together or separately).}
  \label{fig:postprocess}
  \Description{The proposed post-processing pipeline}
\end{figure*}

\section{Related Works}
\label{sec:related_works}

\paragraph{\textbf{Relighting with Generative Models}}
Generative image editing methods has been applied to various image relighting tasks.
In portrait relighting, several works use images of a subject, captured by a light stage rig \cite{debevec00lightstage}, which are used to supervise a generative model \cite{nestmeyer2020physicsGuided, pandey2021totalRelighting, mei2023lightpainter, ren2024relightfulHarmonization, anonymous2024ICLight}.
\hl{\citet{jin2024neuralgaffer} enable general single image object relighting by fine-tuning a diffusion model conditioned on a normalized environment map, using synthetic relighting datasets from \citet{deitke2023objaverse}.}
While effective for portrait relighting, light stage data is object-centric, and could not be easily used to control the illumination of arbitrary light sources in the wild. 
For outdoor scenes, several works assume a single dominant light source, such as the sun, to simulate cast shadows, which can be used to condition a generative model \cite{griffiths2022outcast, Kocsis2024LightIt}.
Compared to outdoor settings, indoor scenes often present \hl{complex multi-illumination conditions.
\citet{li2020inverse} train an inverse rendering network that recovers a spatially-varying spherical Gaussian lighting from a single image.}
\citet{wang2022stylelight, bhattad24stylitgan} find edit directions in StyleGAN's latent space \cite{Karras2019stylegan2} that manipulate light.
However, such methods do not enable direct control over specific light sources, and do not work on real images.
\citet{xing2024luminet} train a ControlNet \cite{Zhang2023ControlNet} to condition a diffusion model with implicit latent intrinsics and illumination representations extracted using the method of \cite{zhang2024latent}.
More closely related to our application, \citet{choi2024scribblelight} use a ControlNet to condition lighting changes on user scribbles.
While their method enables fine-grained spatial control over illumination changes, it biases the model to make localized edits bounded by the user scribble and lacks control over light source intensity and color.

\begin{hlbreakable}
\paragraph{\textbf{Light Control under Multi-illumination}} Image based light editing under multi-illumination has been extensively studied in the context of flash-photography, with applications in white-balancing \cite{hui16@whitebalance}, automatic flash bounce capture \cite{Murmann2016ComputationalBF} and post-capture computational control of flash \cite{Maralan2023Flash}.
\citet{Murmann_2019} collect a multi-illumination dataset using a custom motorized capture device that fires flash at different directions.
They use the linearity of light to synthesize mixed-illuminant data, and train a U-Net \cite{ronneberger2015unet} to change illuminant direction.
\citet{hui2017illuminant} use a flash/no-flash raw image pair to disentangle and manipulate scene illuminants based on their spectral differences.
Notably, when their assumptions hold they can control the intensity and color of particular lights (the flash and joint ambient illumination).
Our goal is to allow similar complex illumination edits, in a more general and accessible setting where only a single no-flash SDR image is given.

\citet{aksoy2018@flashambient} tackle this setting for the case of flash illumination. 
By curating a crowd-sourced dataset of raw flash/no-flash pairs, from which they decompose illuminants to ambient and flash pairs.
Notably, they demonstrate that with ambient illumination variation augmentations, this dataset can be used to train a Pix2Pix model \cite{pix2pix2017}, to decompose ambient and flash illuminations from a single flash photograph.
We also take a data-driven approach and train a diffusion model on a curated dataset of controlled illumination changes, synthesized from raw image pairs.
Specifically, our setting is closer to that of \cite{Haeberli92SyntheticLighting} in which an arbitrary visible light source is turned on/off.
\end{hlbreakable}

\begin{figure*}[t]
    \centering
    \includegraphics[width=1.\linewidth]{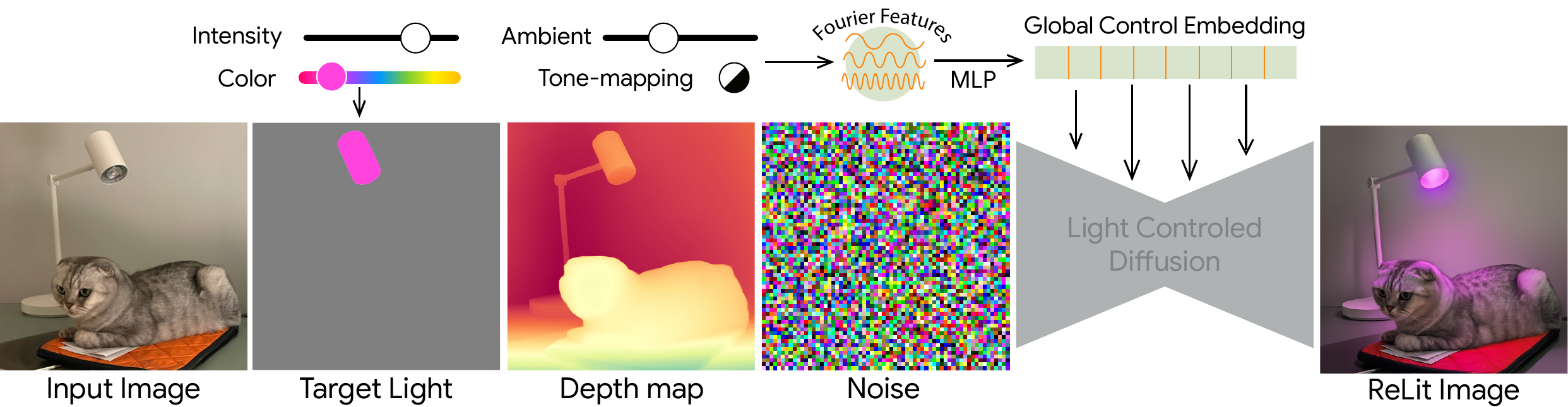}
    \myvspace{-2em}
    \begin{hlbreakable}
    \caption{\textbf{Conditioning Signals.} Spatial conditions (input image, target light mask and depth map) are embedded to the latent dimensions and concatenated to the input noise. Light intensity and color control are applied by scaling the intensity and color of the target light mask.  Global controls (ambient light intensity and tone-mapping value) are projected to text embedding dimension and inserted through cross-attention.}
    \end{hlbreakable}
    \label{fig:diffusion-diagram}
\end{figure*}

\section{Method}
Our method uses paired images to implicitly model controlled light changes in image space, which are used to train a diffusion model. 
In Section~\ref{sec:datasets}, we describe our data collection procedure, which includes a small set of paired raw photographs, supplemented by a large set of synthetically rendered 3D indoor scenes. 
Sections~\ref{sec:post-process} and \ref{sec:tonemapping} continue by explaining how we control the light source and ambient light parameters in post-processing.
Finally, Section~\ref{sec:diffusion} explains how we fine-tune a pre-trained diffusion model for light source control using this mixture of collected datasets.

\subsection{Datasets}
\label{sec:datasets}

\paragraph{\textbf{Photography capture.}}
We capture a set of ~600 raw photograph pairs, using off-the-shelf mobile devices, a tripod, and triggering equipment. 
Each pair depicts the same scene, in which the only physical change is switching on a visible light source.

\hl{To ensure the captured images are well-exposed, we use each device's default auto-exposure setting, and calibrate them using the raw image meta-data during post-capture (see supplementary Section~B.1).}
This dataset provides intricate details of geometry, material appearance, and complex light \hl{phenomena} that may not be found in synthetically rendered data (see below). 

\begin{hlbreakable}
Following previous work \citep{Haeberli92SyntheticLighting, aksoy2018@flashambient, hui2017illuminant}, we regard the "off image" as ambient illumination $\mathbf{i}_\text{amb} \vcentcolon= \mathbf{i}_\text{off}$ and extract the illumination from the target light: $\mathbf{i}_\text{change} = \mathbf{i}_\text{on} - \mathbf{i}_\text{off}$.
This difference may have negative values due to captured noise, errors in the post-capture calibration process, or slight differences in ambient illumination conditions between the two images.
To avoid the resulting unintended dimming, we clip the difference to be non-negative: $\mathbf{i}_\text{change} = \text{clip}(\mathbf{i}_\text{on} - \mathbf{i}_\text{off},\, 0)$. We show statistics for this clip operation in Section B.1 in the supplementary.
\end{hlbreakable}

As we show in Section~\ref{sec:effect_of_domain}, incorporating real data helps to disentangle the intended illumination changes from the style of synthetically rendered images, which do not include visual artifacts introduced by a real physical camera sensor such as lens distortion or chromatic aberrations, among others.
In post-processing (Section~\ref{sec:post-process}), we inflate each real image pair by a factor of 60 to encompass a range of intensities and colors. 
The complete dataset after post-processing includes approximately $36\text{K}$ images.

\paragraph{\textbf{Synthetic rendering}}
To complement the captured dataset, we use a larger set of procedurally generated synthetic images of 3D scenes, rendered using a physically-based renderer.  
We begin with 20 synthetic indoor scenes constructed by 3D artists using Blender \cite{blender}, 
from which we procedurally render images. Our rendering pipeline randomly samples camera views around view target objects, and procedurally sets virtual light sources  parameters (e.g. intensity, color temperature, area size, cone angle, etc.).
We further expand this data by randomly sampling plausible light fixture locations, and adding them to the scenes.
Furthermore, we apply different environment maps of varying strengths, and randomly configure background illumination.
\hl{See the supplementary Section~A.1 for further details.}

\hl{We render synthetic images with each light component $\mathbf{i}_\text{amb}$ and $\mathbf{i}_\text{change}$ separately, which are later combined in post-processing (Section~\ref{sec:post-process})}.
\hl{Images are created in linear RGB color space, using a Monte-Carlo path tracing renderer. Commonly, this may create pixels with unbounded outlier sampled values, when a path' sampling probability is very low.
To avoid subsequent noisy tone-mapping behavior, we apply a bound $E_{max}$, set to the top $5\times10^{-4}$ percentile of pixel values, computed on a subset of $2000$ random renders.}

This dataset spans 16,000 combinations of target light source, camera view, and environmental lighting, which are then also inflated by a factor of 36 in post-processing totaling approximately $600\text{K}$ images. 
Despite the relatively low diversity in scene geometries and materials, in Section~\ref{sec:effect_of_domain} we show that the synthetic data steers the model to create physically plausible, view-consistent lighting.

\subsection{Image-based Relighting}
\label{sec:post-process}

\hl{Given the disentangled linear RGB images $\mathbf{i}_\text{amb}$ and $\mathbf{i}_\text{change}$}, we generate a parametric sequence of images with varying target light source intensity and color and ambient illumination.

To change the target light color, we multiply it by a change coefficients $\textbf{c} \in \mathbb{R}^3$ in linear RGB space.
To compute this coefficient, we first estimate the original light RGB $\textbf{c}_\text{o}$, and then multiply by the desired RGB coefficients $\textbf{c}_\text{t}$ to obtain $\textbf{c}= \textbf{c}_\text{t} \odot {\textbf{c}_\text{o}}^{-1}$, where $\odot$ denotes element-wise multiplication and ${\textbf{c}_\text{o}}^{-1}$ element-wise inverse.

\begin{hlbreakable}
For a relative ambient illumination intensity $\alpha \in [0,1]$, a relative target light intensity $\gamma\in [0,1]$ and the target light RGB $\textbf{c}_\text{t}$ the relit image is computed by:
\begin{equation} \label{eq:relit}
    \mathbf{i}_\text{relit}\left( \alpha, \gamma, \textbf{c}_\text{t} ;\; \mathbf{i}_\text{amb}, \mathbf{i}_\text{change}\right) = \alpha \, \mathbf{i}_\text{amb} + \gamma \, \mathbf{i}_\text{change} \textbf{c}
\end{equation}
\end{hlbreakable}
See examples of post-processed samples in the supplementary.

\subsection{Tone-Mapping Strategy}
\label{sec:tonemapping}
\hl{The relit image sequences (Equation~\ref{eq:relit}) need to be tone-mapped to match the training distribution of the base diffusion model.}
\hl{One approach is to tone-map each relit image separately, ensuring it is well exposed.}
\begin{figure}[hb!]
    \centering
    \setlength{\tabcolsep}{1.5pt}
    {\small
    \begin{tabular}{c c c c}
        \textbf{0.00} & \textbf{0.25} & \textbf{0.50} & \textbf{1.00} \\
        
        \includegraphics[width=0.22\linewidth]{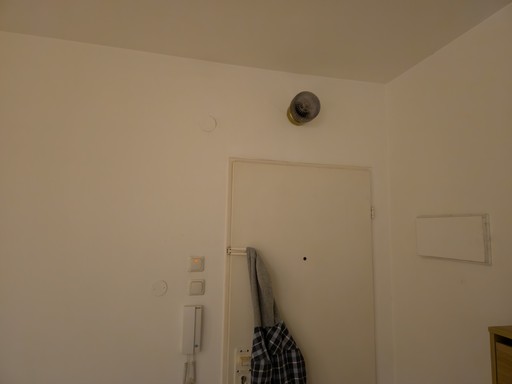} &
        \includegraphics[width=0.22\linewidth]{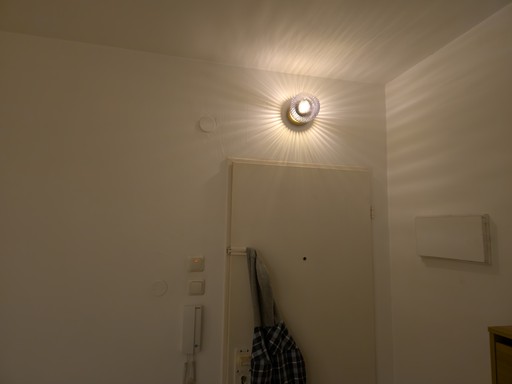} &
        \includegraphics[width=0.22\linewidth]{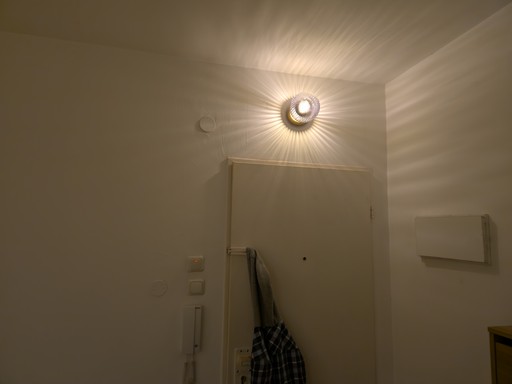} &
        \includegraphics[width=0.22\linewidth]{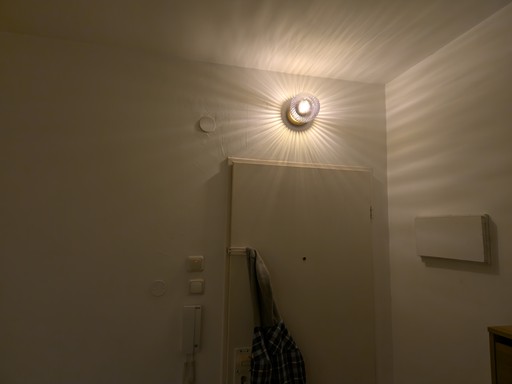} \\   
        
        \includegraphics[width=0.22\linewidth]{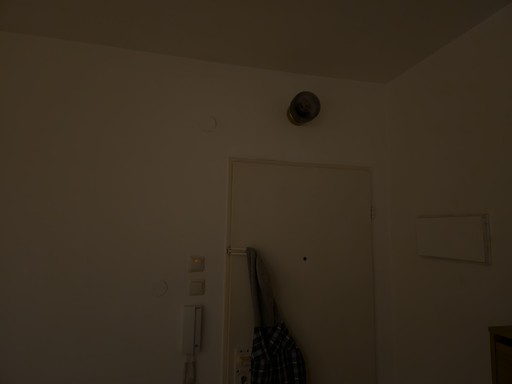} &
        \includegraphics[width=0.22\linewidth]{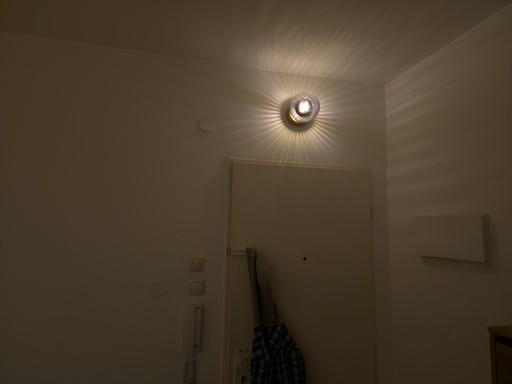} &
        \includegraphics[width=0.22\linewidth]{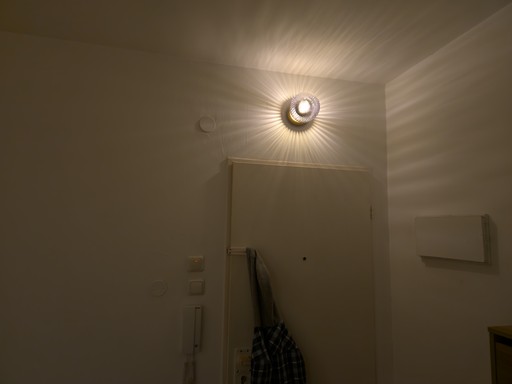} &
        \includegraphics[width=0.22\linewidth]{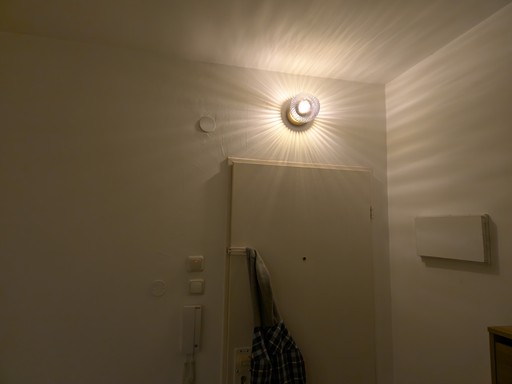} \\  
    \end{tabular}
    }
    \caption{\textbf{Tone mapping strategy.} A sequence of images of increasing light intensity, tone mapped either separately or together.
    \textbf{Top row.} The images tone mapped separately, notice how the light source intensity appears constant when lit, while ambient light appears to be dimmed.
    \textbf{Bottom row.} Tone mapped together.
    }
    \label{fig:tonemapping_artifcats}
\end{figure}

We find that this approach produces inconsistent image sequences, where the perceived changes in light intensity do not align with the physical change.
\hl{For example, when the target light source dominates the dynamic range, it appears constant for increasing intensities, while ambient light is dimmed instead (Figure~\ref{fig:tonemapping_artifcats}).}

\hl{We address this problem by tone-mapping image sequences together using the same fixed exposures,} based on \cite{hasinof16HDR, mertens07ExpsureFusion}.
Given an image pair in linear space: $\mathbf{i}_\text{off},\, \mathbf{i}_\text{on}$, we heuristically choose \emph{deciding intensities} $\gamma_d$, $\alpha_d$ for the target light source and ambient light, respectively. 
The synthetic exposures used in \cite{hasinof16HDR} are computed with respect to the relit image $\mathbf{i}_\text{relit}\left( \alpha_d, \gamma_d, \textbf{c}_\text{t} ;\; \mathbf{i}_\text{amb}, \mathbf{i}_\text{change}\right)$ and are applied across all different light source, and ambient light intensity combinations.

\begin{hlbreakable}
Although using fixed exposures produces intuitive sequences of varying light intensity, it can cause problems when solely used to train a relighting model.
Firstly, at inference time we expect input images to be captured using auto-exposure, and that are tone-mapped individually.
Secondly, we want to allow the user to decide how the model's output should be tone-mapped, providing control over the trade-off between a well-exposed output and intuitive light changes.
As such, we tone-map our data using both strategies and expose the used strategy as an input condition to the diffusion model.
We show the effect of this condition on the model's output in Figure~\ref{fig:more_results} (bottom row) and Figure D.11 in the Supplement.
\end{hlbreakable}
 
\begin{figure}[tb]
    \centering
    \setlength{\tabcolsep}{1.1pt}
    {\small
    \begin{tabular}{c c c c}
    \textbf{Input} & \textbf{+0.25} & \textbf{+0.50} & \textbf{+1.00} \\
    \includegraphics[width=0.242\columnwidth]{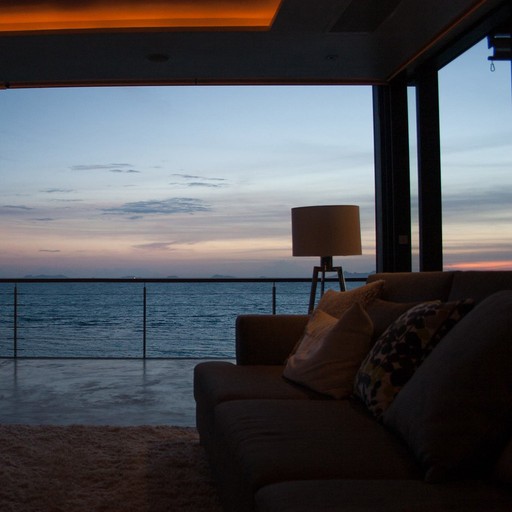} & 
    \includegraphics[width=0.242\columnwidth]{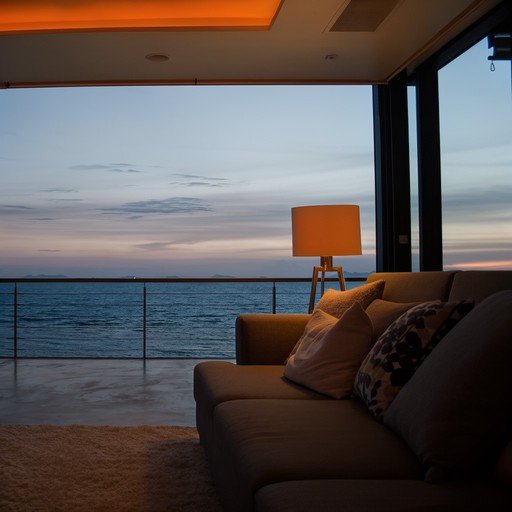} &
    \includegraphics[width=0.242\columnwidth]{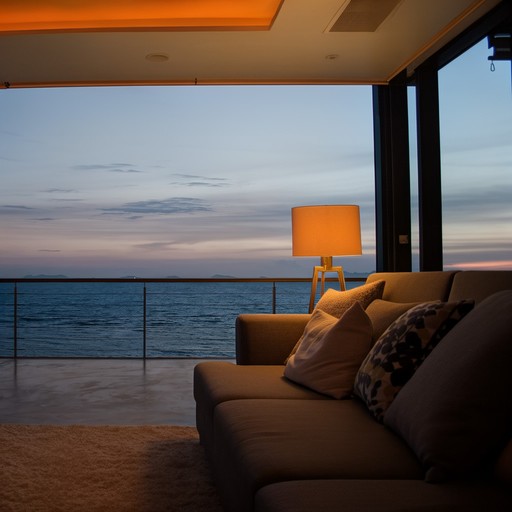} &
    \includegraphics[width=0.242\columnwidth]{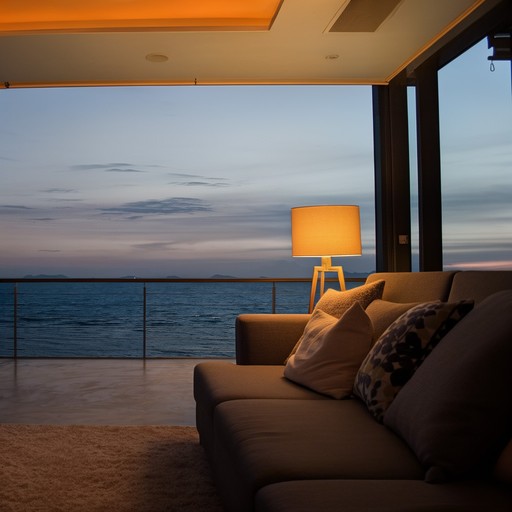} \\
    \textbf{Input} & \textbf{-0.50} & \textbf{-0.75} & \textbf{-1.00} \\
    \includegraphics[width=0.242\columnwidth]{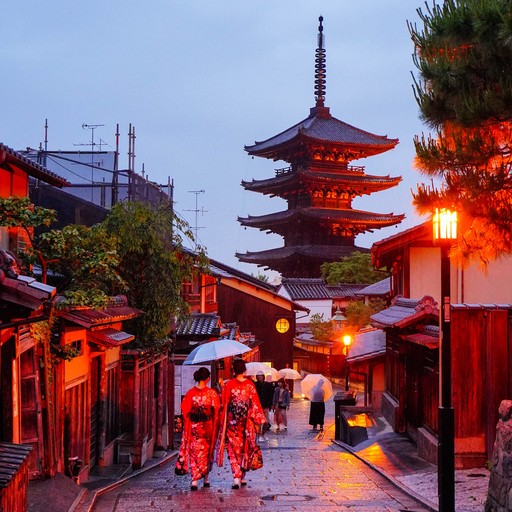} & 
    \includegraphics[width=0.242\columnwidth]{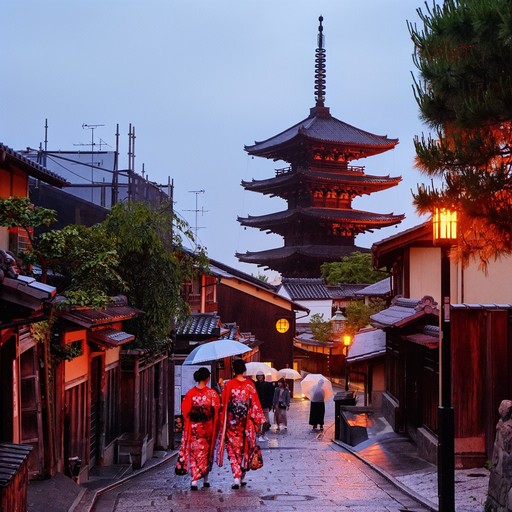} &
    \includegraphics[width=0.242\columnwidth]{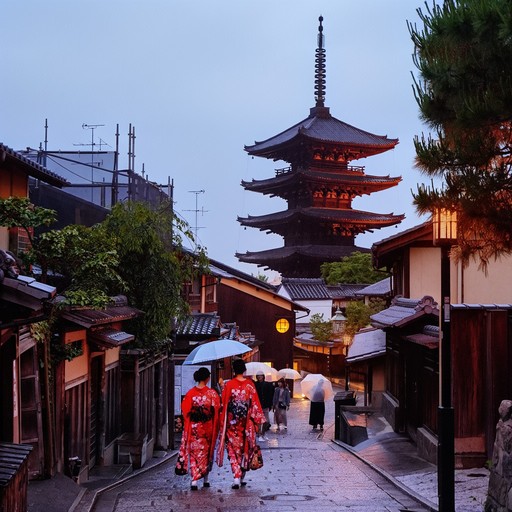} &
    \includegraphics[width=0.242\columnwidth]{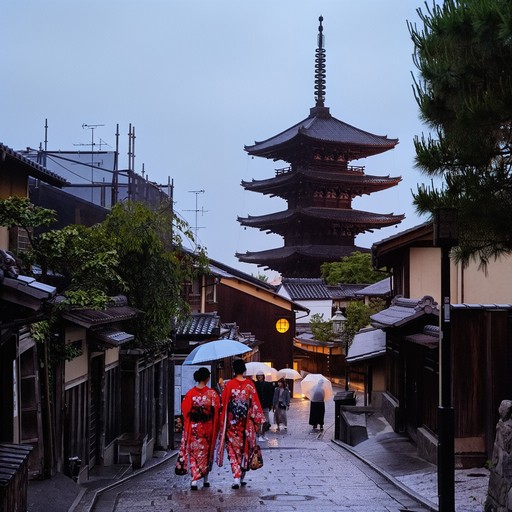} \\
    \end{tabular}
    }
    \caption{\textbf{Intensity control}. Fine-grained control over a target light's intensity using our method. Values represent the relative intensity change with respect to the source image.}
    \label{fig:results_intensity}
\end{figure}

\subsection{Light Source Control with Diffusion Models}
\label{sec:diffusion}
We fine-tune a pre-trained latent diffusion model~\cite{rombach2022high, saharia2022photorealistic, ramesh2022hierarchical} to enable parametric control over visible light source parameters, ambient light intensity, and tone mapping effects. 
\begin{hlbreakable}
We use different conditioning schemes for localized spatial signals and for global controls.

\paragraph{\textbf{Spatial conditions.}} 
Spatial conditions include the input image, an automatically extracted depth map of the input image, and two spatial segmentation masks for the target light source intensity change and color respectively.
To represent the target light source, we use a semantic segmentation mask acquired using \cite{ravi2024sam2segmentimages} conditioned on a user-given bounding box.
The mask is used twice to condition the model on both the target light intensity change and on the target color. 
For the intensity condition, the mask is multiplied by the relative light intensity change scalar (corresponding to $\gamma$ in Equation~\ref{eq:relit}). 
For the target color condition, the mask is expanded to 3 channels and scaled to the requested target RGB color (corresponding to $\textbf{c}$ in Equation~\ref{eq:relit}).
The input image is encoded using the model's Variational Auto-Encoder (VAE), while other conditions are resized to match the spatial latent dimensions. Following this, a learned $1\times1$ convolution is used to match the number of channels of the spatial conditions tensor. The resulting tensor is then concatenated to the input noise \cite{nichol2021glide, wang2023imageneditoreditbenchadvancing}.
We provide additional details regarding the conditioning and sampling scheme of light change parameters during training in the supplementary Sections C.1 and C.3.

\begin{figure}[t!]
    \footnotesize
    \centering
    \includegraphics[width=1.\linewidth]{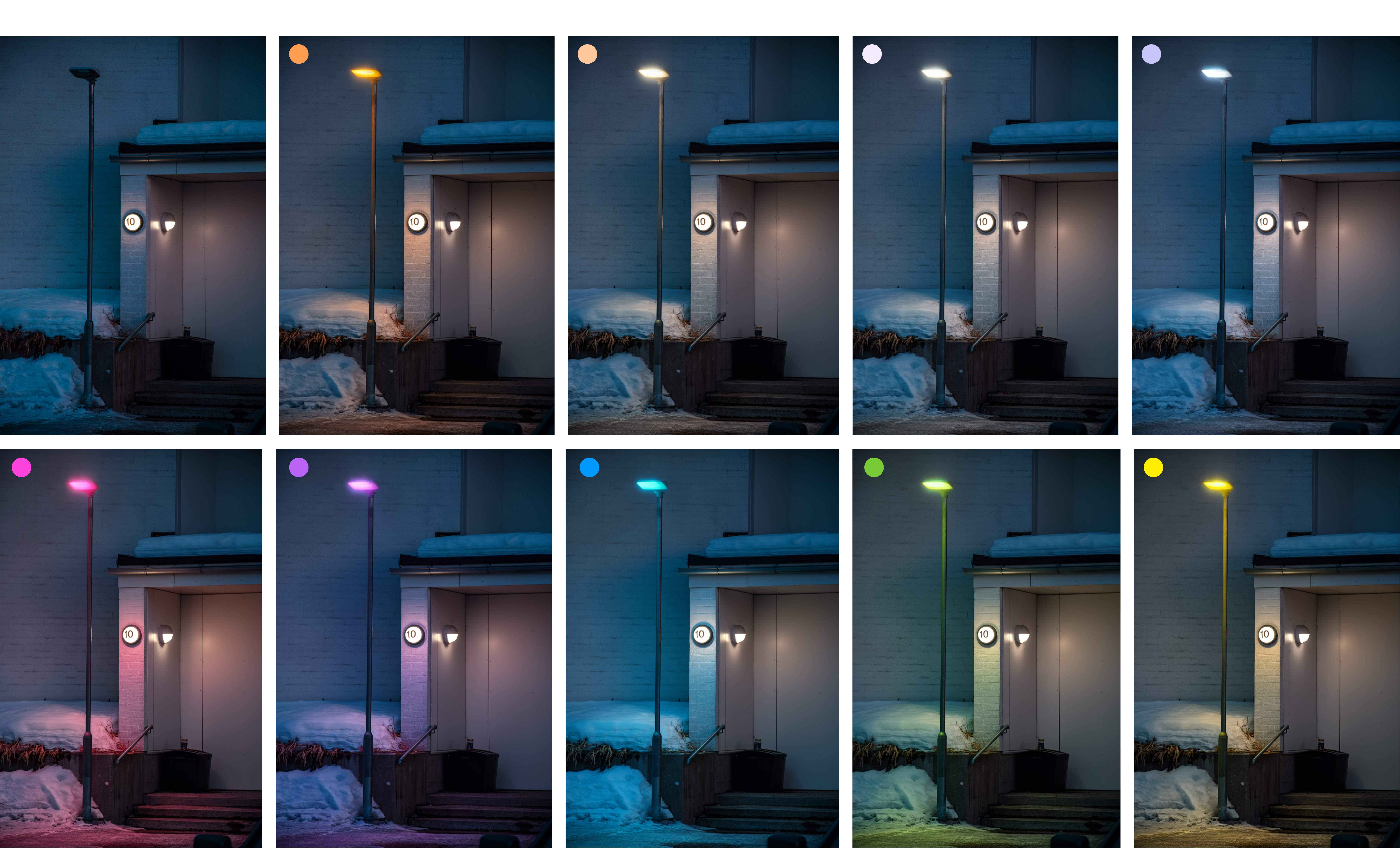}
    \caption{\textbf{Color Control.} We turn on the street lamp in the input image (top left) with different colors. \textbf{Top row.} artificial light blackbody temperatures. \textbf{Bottom row} arbitrary non-natural RGB colors.
    }
    \label{fig:results_colors}
\end{figure}

\paragraph{\textbf{Global conditions.}} These include the ambient light change as a scalar value in the range $[-1,1]$, and the requested tone-mapping strategy as a binary scalar value, which affects the exposure of generated images during inference as described in Section~\ref{sec:tonemapping}.
\end{hlbreakable}
Global controls are encoded into Fourier features ~\cite{tancik2020fourfeat} followed by a multi-layer perceptron (MLP) that projects them to the text embedding dimension. 
The projections are concatenated to the text embeddings and then inserted through the text-to-image cross-attention layers.

\section{Experiments}
\subsection{Implementation Details}
\label{sec:implementation_details}

\paragraph{\textbf{Model and training.}} We fine-tune a text-to-image latent diffusion model with the \hl{same architecture layout and similar hidden dimensions as Stable Diffusion-XL \cite{podell2023sdxlimprovinglatentdiffusion}, which was trained on a subset of \citet{chen2022pali}.}
In Section~\ref{sec:effect_of_domain} we train each model for 45,000 steps with a learning rate of $10^{-5}$, and a batch size of $128$ at $1024 \times 1024$ resolution. Training took $\sim$12 hours on 64 v4 TPU's. During training, we drop the depth and color conditions $10\%$ of the time to allow unconditional inference.
We use \cite{ravi2024sam2segmentimages} light source segmentation and \cite{yang2024depthv2} to create depth maps. 
\hl{See Section B.2 in the supplementary for the masking procedure.
Running inference with 15 denoising steps took approximately $5$ seconds on a single v4 TPU.}

\paragraph{\textbf{Evaluation datasets.}} 
For quantitative ablations and comparisons, we evaluate trained models on \emph{paired} datasets, curated using the procedure described in Section \ref{sec:datasets}. 
The real photograph dataset contains ~200 photo pairs of different scenes and light sources, which is expanded by a factor of 60 during post-processing.
The synthetic evaluation dataset includes renders from two held out scenes, containing unique light sources, objects, and materials.
For qualitative evaluations, where a ground truth target is not needed, we evaluate the model on a collection of 100 images from \cite{bell14intrinsic}.
For these images, we manually annotate the target light sources in each image and compute their respective segmentation masks and depths. 
\hl{Throughout the evaluation and to generate all results in the paper, the tone-mapping condition was set to "together", unless stated otherwise.}
\paragraph{\textbf{Evaluation metrics.}}
\label{par:eval_metrics}
We measure the model's performance on paired images with two common metrics: Peak Signal to Noise Ratio (PSNR) and Structural Similarity Index Measure (SSIM) \cite{ren2024relightfulHarmonization, xing2024luminet, zeng2024rgbx, zeng2024dilightnet}.
To supplement these measurements, we provide qualitative results in Figures~\ref{fig:domains_qualitative_desk}, \ref{fig:comparisons_iiw} and in the supplementary materials. Furthermore, we verify that these results align with user preference by conducting a user study to compare with other methods in Table~\ref{tab:comparisons}. More information about our user study can be found in the supplementary materials.

\begin{table}[t!]
    \centering
    \footnotesize
    \caption{\textbf{Effect of Training Domain.} Ground truth similarity metrics for models trained on different weightings between domains.
    \emph{Binary.} Image pairs that represent either completely turning on or off a light source.
    }
    \setlength{\tabcolsep}{1pt}
    \begin{tabular}{
        l@{\hspace{2pt}} 
        l@{\hspace{2pt}} 
        c@{\hspace{4pt}} 
        c@{\hspace{4pt}} 
        c@{\hspace{6pt}} 
        c@{\hspace{4pt}} 
        c@{\hspace{4pt}} 
        c@{\hspace{4pt}}
    }
    \toprule
    \multicolumn{2}{l}{\multirow{2}{*}{\textbf{Training Datasets}}} & \multicolumn{3}{c}{\textbf{PSNR}$\uparrow$} & \multicolumn{3}{c}{\textbf{SSIM}$\uparrow$} \\
    & & Binary & Intensity & Color & Binary & Intensity & Color \\
    \midrule

    \multirow{2}{*}{Real + \synthetic{}} & {\tiny w/ depth} & \colorbox{tabfirst}{23.2} & \colorbox{tabfirst}{28.6} & \colorbox{tabfirst}{24.2} & \colorbox{tabfirst}{0.818} & \colorbox{tabfirst}{0.879} & \colorbox{tabfirst}{0.874} \\
    & {\tiny w/o depth} & 23.2 & 28.5 & 24.1 & 0.816 & 0.876 & 0.871 \\
    
    \midrule

    \multicolumn{2}{l}{Real + \synthetic{} (Binary)} & 22.7 & - & \colorbox{tabsecond}{23.8} & 0.818 & - & 0.864 \\

    \multicolumn{2}{l}{Real Only} & \colorbox{tabsecond}{22.9} & \colorbox{tabsecond}{28.3} & 23.75 & \colorbox{tabsecond}{0.815} & \colorbox{tabsecond}{0.879} & \colorbox{tabsecond}{0.870} \\

    \multicolumn{2}{l}{\synthetic{} Only} & 20.71 & 27.38 & 22.33 & 0.7947 & 0.8730 & 0.8605 \\

    \bottomrule
    \end{tabular}
    \label{tab:domain}
\end{table}

\subsection{The Effect of Different Domains}
\label{sec:effect_of_domain}
Empirically, we find that using a mixture of both a small set of real captures and synthetic 3D renders yields the best results.
Nevertheless, understanding the effect of each domain dataset, and its corresponding augmentations, on the trained models performance provides insights for further research.
In this section we isolate these effects we evaluate models fine-tuned from the same base model, with different data distributions. 

\input{assets/figures/effect_of_synth_desk}
\paragraph{\textbf{Generalization across domains}} We observe that the model trained only on synthetically rendered data does not generalize well to real images (Table~\ref{tab:domain} third row). 
We attribute this generalization error to differences in style, e.g. lack of complex and intricate geometries, fidelity of textures and materials, absence of camera artifacts such as glare, which are not present in the synthetic dataset.
Empirically, this domain gap causes a knowledge drift in the base diffusion model, as can be seen in Figure D.3 in the supplementary.
\label{sec:domain_generalization}

\paragraph{\textbf{Using multiple domains.}} We train three models using the same procedure on three mixtures of the data domains: real captures only, synthetic renders only, and their weighted mixture. The results in Table~\ref{tab:domain} show that using a weighted mixture of data from both domains achieves the best results for all settings. 
Notably, we observe a small quantitative relative difference between the mixture dataset and real captures only, despite the significant difference in their size. For example, adding synthetic data results in only $2.2\%$ averaged improvement in PSNR.
This is likely due to perceivable local changes in illumination, such as small instance shadows, and specular reflections, are obscured by image-wide, low-frequency details \hl{in these metrics}.
\begin{table}[t]
    \centering
    \caption{\textbf{Comparison with other works.} \textbf{Left} ground truth similarity in PSNR and SSIM on our paired evaluation set Section~\ref{par:eval_metrics}.
    \textbf{Right} user study preference rates for our method.
    Our method outperforms previous results both in physical plausibility and in user satisfaction.
    }
    \begin{tabular}{
        l@{\hspace{9pt}}
        c@{\hspace{9pt}}
        c@{\hspace{9pt}}
        c@{\hspace{9pt}}
    }
    \toprule
    {\textbf{Method}} & {\textbf{PSNR}$\shortuparrow$} & {\textbf{SSIM}$\shortuparrow$} & {\textbf{Ours Win Rate}$\shortuparrow$} \\
    \midrule
    RGB $\leftrightarrow$X & 12.0  & 0.518 & 89.3 \%  \\
    IC-Light & 12.2 & 0.507 & 83.0 \% \\ %
    OmniGen  & 15.1  & 0.584 & 83.0 \%  \\ %
    ScribbleLight  & 13.8  & 0.418 & 84.8 \% \\ %
    \midrule
    Ours & \colorbox{tabfirst}{23.2}  &  \colorbox{tabfirst}{0.818} & - \\
    \bottomrule
    \end{tabular}
    \label{tab:comparisons}
\end{table}

We corroborate this effect with qualitative comparisons in Figures~\ref{fig:domains_qualitative_desk}, and Figure C.2 in the supplementary, which demonstrate how adding synthetic data encourages the model to produce intricate local shadows which are not present in the real-only model.

\subsection{Comparisons}
\label{sec:comparisons}
\begin{figure*}   
    \footnotesize
    \setlength{\tabcolsep}{1.0pt}
    \begin{tabular}{c c c c c c}
        \textbf{Input (On/Off)} & \textbf{OmniGen} & \textbf{RGB $\leftrightarrow$ X} & \textbf{ScribbleLight} & \textbf{IC-Light} & \textbf{Ours} \\
        \includegraphics[width=0.162\textwidth]{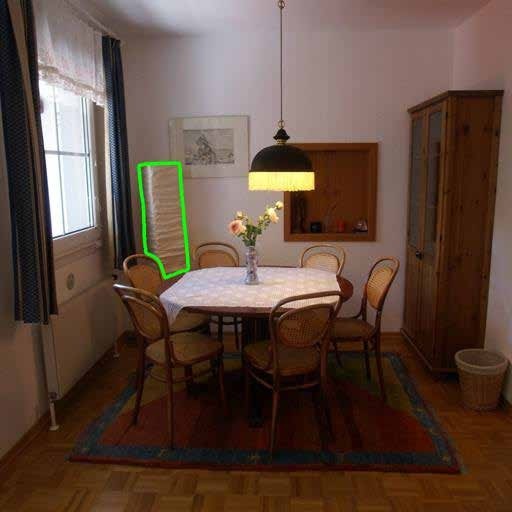} &
        \includegraphics[width=0.162\textwidth]{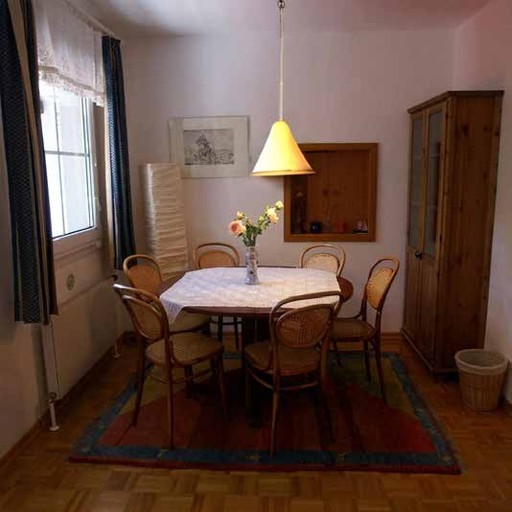} &
        \includegraphics[width=0.162\textwidth]{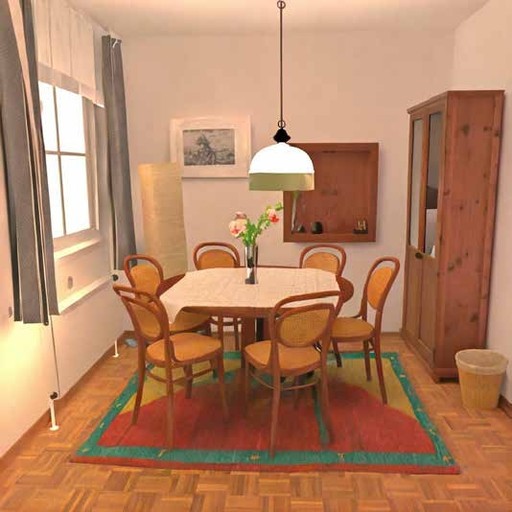} &
        \includegraphics[width=0.162\textwidth]{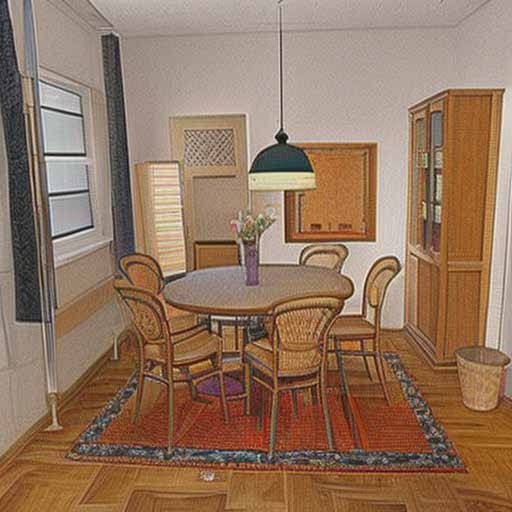} &
        \includegraphics[width=0.162\textwidth]{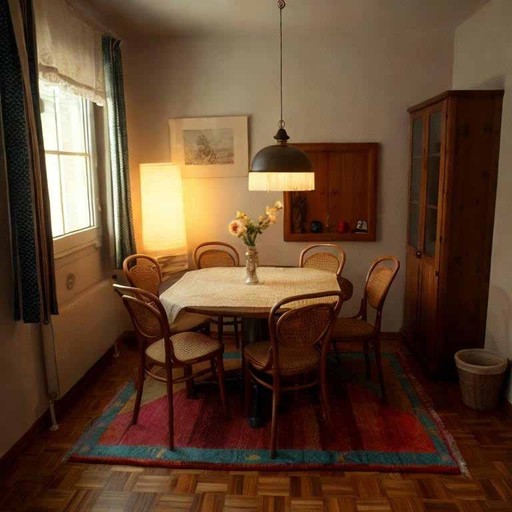} &
        \includegraphics[width=0.162\textwidth]{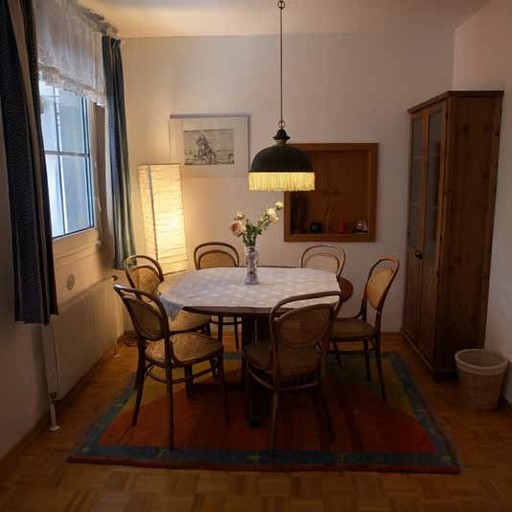} \\
        \includegraphics[width=0.162\textwidth]{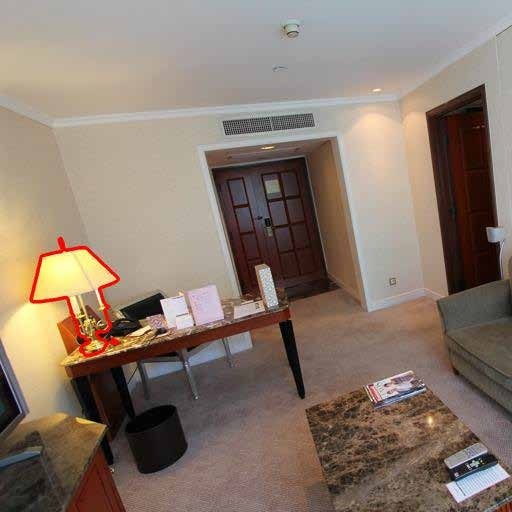} &
        \includegraphics[width=0.162\textwidth]{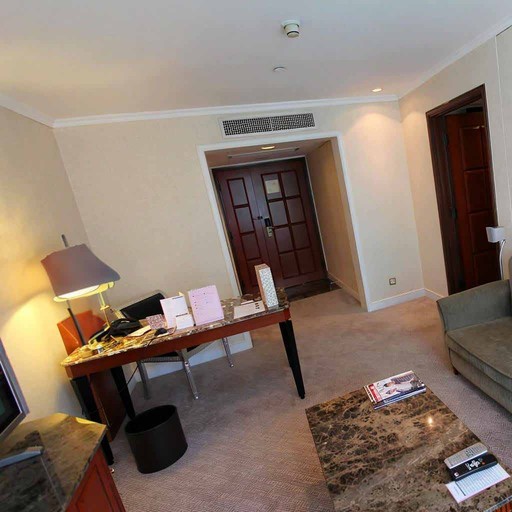} &
        \includegraphics[width=0.162\textwidth]{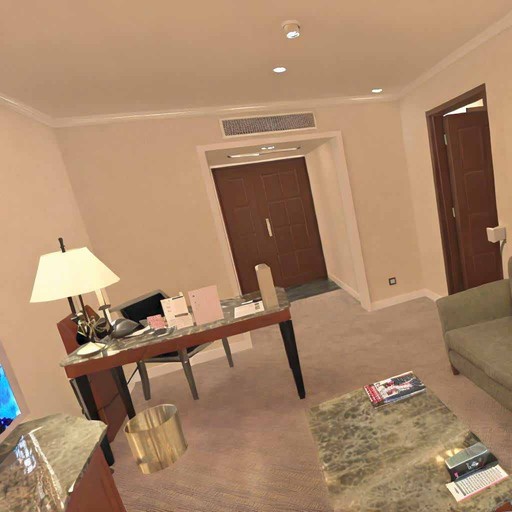} &
        \includegraphics[width=0.162\textwidth]{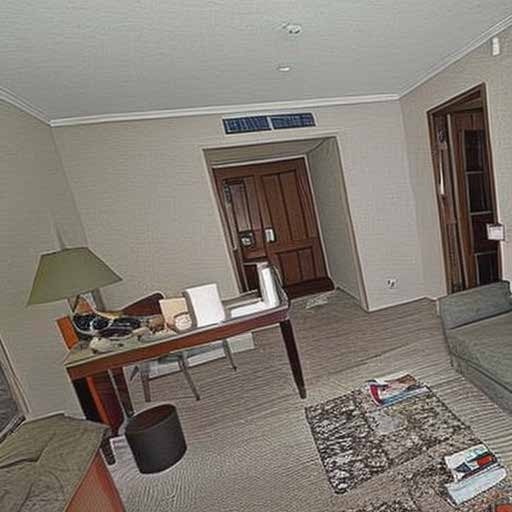} &
        \includegraphics[width=0.162\textwidth]{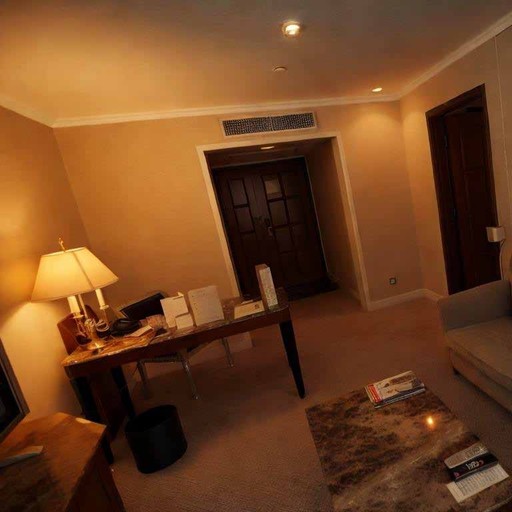} &
        \includegraphics[width=0.162\textwidth]{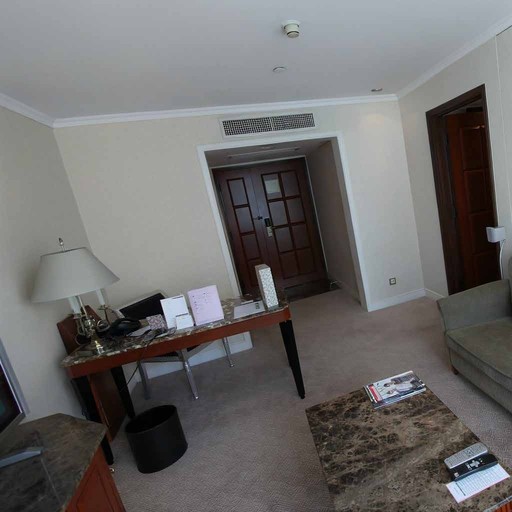} \\
    \end{tabular}
    \\[-0.65\baselineskip]
    \caption{\textbf{Qualitative Comparison.} Comparison with other works on images from IIW dataset. \textbf{Left.} the input image where \textcolor{green}{green} / \textcolor{red}{red} contours specify which light source(s) should be turned \textcolor{green}{on} / \textcolor{red}{off} respectively. See Figures D.5, D.6 in the supplementary for more examples.}
    \label{fig:comparisons_iiw}
\end{figure*}

To the best of our knowledge, our method is the first to provide fine-grained control over light sources in a real single image. Therefore, for a fair comparison, when comparing to other works, we evaluate only on the binary task. As baselines, we adapt four diffusion-based editing methods: OmniGen \cite{xiao2024omnigen}, \rgbx{} \cite{zeng2024rgbx}, ScribbleLight~\cite{choi2024scribblelight}, and IC-Light \cite{anonymous2024ICLight}. 
These methods use text prompts describing the light change and light source location in the input images, and various other scene intrinsics.
The \rgbx{} model is conditioned on an multiple pre-computed normals, albedo, roughness, and metallically maps of the input images. 
ScribbleLight receives the albedo and a mask layer that indicates where light should be turned on/off (as opposed to the light source mask for our method).
Lastly, to use IC-Light to control light sources, we input the entire image as the foreground and provide our light source segmentation masks as the environment light source condition.

As can be seen from Table~\ref{tab:comparisons}, our method outperforms previous methods by a significant margin.
We also provide a qualitative comparison on evaluation images from \cite{bell14intrinsic} in ~\ref{fig:comparisons_iiw}. 
Notably, OmniGen fails to turn the target light sources on/off, and introduces local geometry changes. 
\rgbx{}, ScribbleLight, and IC-Light can succeed in changing input lighting conditions, but usually results in additional unwanted illumination changes or color distortion. 
With respect to prior works, our method faithfully controls the target light source, and generates physically-plausible lighting.

\begin{hlbreakable}
\subsection{Qualitative Results} \label{sec:qualtiative_discussion}
\input{assets/figures/qualitative_discussion}
We present additional qualitative results, demonstrating our method's ability to generate different illumination phenomena, and its representative failure cases.
\paragraph{\textbf{Out-of-domain light sources.}} \label{sec:ood_ls} Figure~\ref{fig:qualitative_results} (b - red) shows an example where our model lights candles as light-tubes. 
This is likely due to bias in the light sources represented in the fine-tuning datasets.
Interestingly, this bias is less prominent when turning off the light, as demonstrated in Figure~\ref{fig:qualitative_results} (d) where the model turns off the paper lantern.
Furthermore, Figure~\ref{fig:more_results} (second row) shows that the model can faithfully turn on a well-represented light source in out-of-domain images (non-photorealistic cartoon image). Note how the desk lamp is lit to match the image (zoom in on the slits at the top of the lamp shade).

\paragraph{\textbf{Reflections.}} Figure~\ref{fig:qualitative_results} (b - green) shows an example where the model lights the mirror reflection of the chandelier, another example can be seen in Figure~\ref{fig:teaser} (bottom row, second image from the left), where the model preserves the reflection of the ceiling lamp on the glass door when turning off ambient day light entering through it.
Figures~\ref{fig:qualitative_results} (d - right box) and (f - green) show examples of removed / generated specular reflections from metallic materials, respectively. 

\paragraph{\textbf{Perspective failures.}} Our dataset contains mostly indoor images with bounded depth of field. 
This bias can create physically inaccurate results in cases such as forced perspective, as seen in Figure~\ref{fig:results_intensity} (bottom row) where turning off the street light near the camera removes the red light projected on the distant pagoda. 

\end{hlbreakable}

\section{Applications}
\label{sec:applications}
\begin{figure*}[tb]
    \small
    \centering
    \includegraphics[width=1.\linewidth]{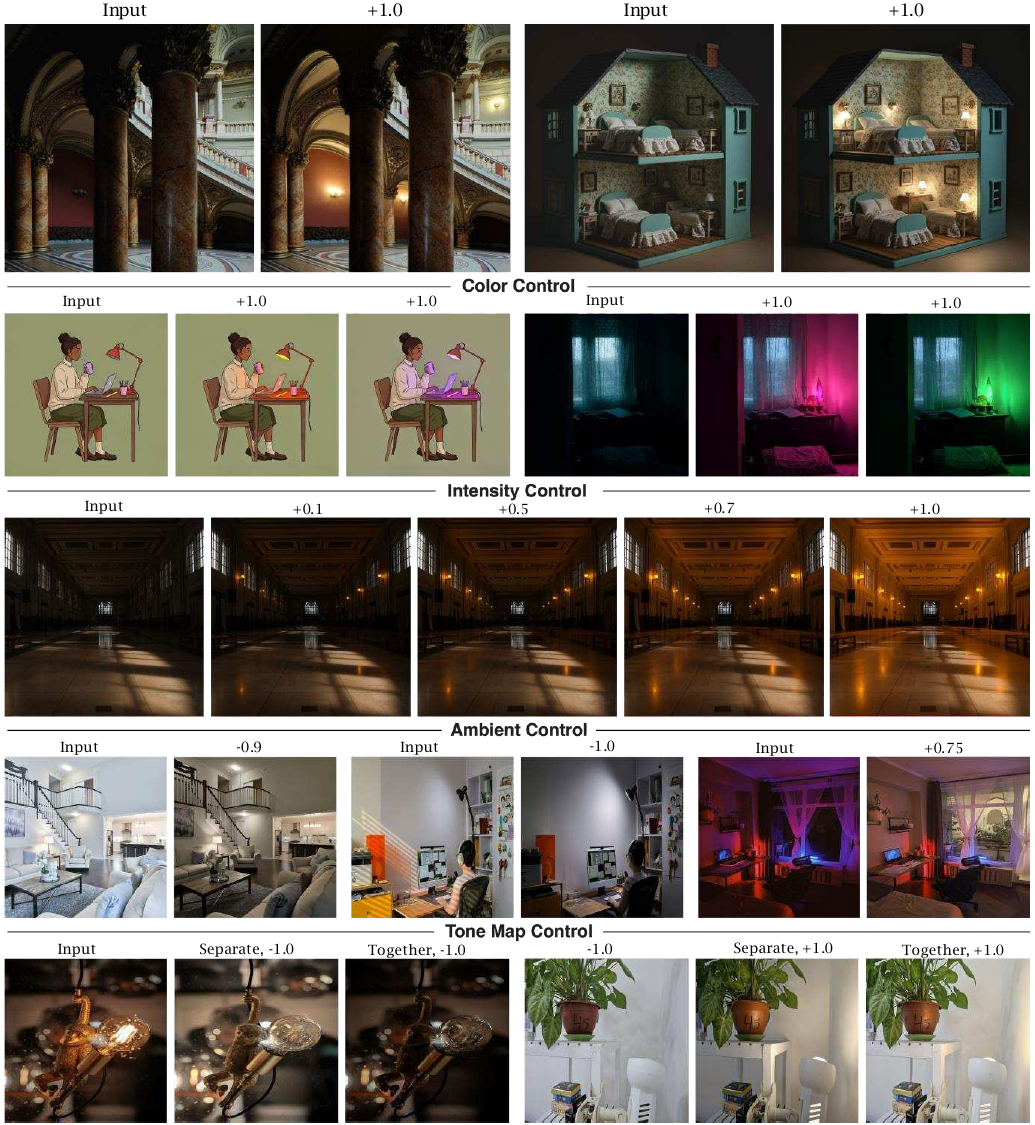}

    \caption{\textbf{Additional results.}. Our method can control a target light intensity (third row), light color (second row) and a scene's ambient light (fourth row) \hl{and tone-mapping effects (fifth row)} across diverse scenes and image styles.
    The number above the results indicate the relative light intensity change used for the target light or ambient.
    See the complete inputs used to generate results in the supplementary.
    }
    \label{fig:more_results}
\end{figure*}

We present several possible applications of our method in various settings. 
The primary one is the ability to control lights in a photograph post-capture. Examples of turning lights on or off can be seen in Figures ~\ref{fig:teaser}, \ref{fig:comparisons_iiw}, \ref{fig:more_results}. More accurate control of light intensity can be seen in Figures ~\ref{fig:results_intensity}, control of color can be seen in Figures~\ref{fig:results_colors}, and control of ambient light can be seen in Figure~\ref{fig:more_results}.

 In addition, we showcase how our method can be used for consistent light editing that combines light intensity, color and ambient changes. For example, notice how, in the \textit{tatami room} image in Figure~\ref{fig:teaser}, modification of different light sources causes disentangled light changes over the input image. For more results, see our Supplemental Video and our interactive demo.

\begin{figure*}[h]
    
\includegraphics[width=\textwidth]{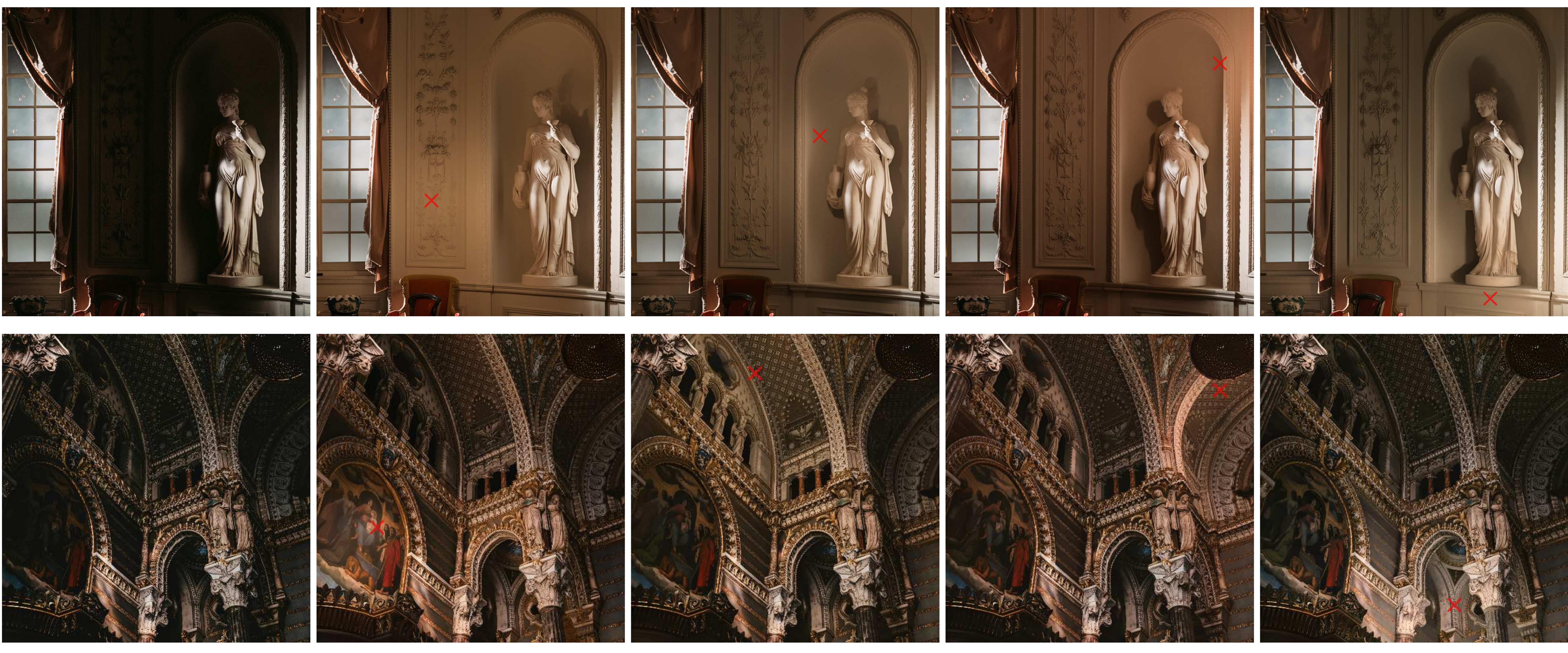}
\includegraphics[width=\textwidth]{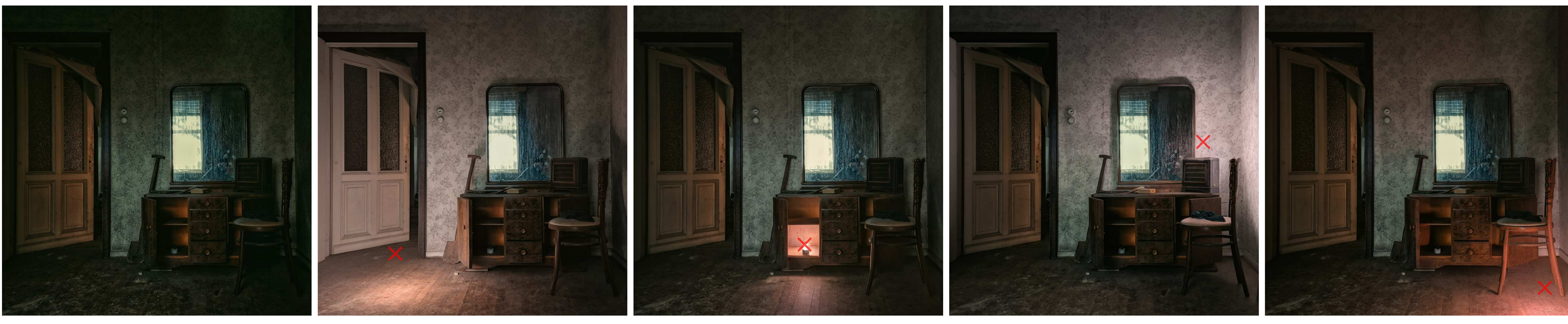}
\\[-0.65\baselineskip]
\caption{\textbf{Virtual Point Lights.} By using synthetic renderings of point lights without geometry we can insert an "invisible" light source into the scene. On the left of each sequence is the input image, red "\textcolor{red}{X}" marks the insertion point. Notice how the localization of the light source.
}
\label{fig:virtual_light}
\end{figure*}

In Figure~\ref{fig:virtual_light} we demonstrate another possible application of using synthetic data, inserting virtual light sources - ones without any geometry into the scene. 
The images were rendered as in Section~\ref{sec:datasets}, without the light source geometry. This application alleviates the requirement of a visible light source in the image. 

The ability to turn on lights anywhere in a 2D image also opens up possibilities for animation. In figure~\ref{fig:stop_motion} we present an example where we move the position of a turned-off lamp on a table, and in post-processing we turn it on. As can be seen, LightSwitch correctly interprets the geometry of the scene and creates plausible shadows and lighting throughout the sequence - allowing one to create a stop-motion type animation.

\section{Limitations}
\label{sec:limitations}
\begin{hlbreakable}
While fine-tuning on a low-diversity dataset can generate pleasing relighting results in many settings, the resulting model suffers from bias in the dataset. %
This bias is most evident on the type of target light source as demonstrated in Section~\ref{sec:ood_ls}.
We believe that combining our method with an unpaired fine-tuning method could mitigate this bias and is an interesting direction for future research.

Secondly, we use a simplistic data capture process, using consumer mobile devices and calibrating exposure post-capture. Although using this method allows us to easily collect the captured dataset, it does not enable precise relighting in physical units.

\end{hlbreakable}

\section{Conclusion}
\label{sec:conclusion}
We presented \ourmethod - a diffusion-based method for fine-grained control of light sources in images. 
Using the linearity of light and synthetic 3D data, we create a high-quality data set of paired images, which implicitly model complex and controlled illumination changes.
\begin{hlbreakable}
We show that fine-tuning a diffusion model using physically-based data, one can achieve physically-plausible control over light sources after capture.
We believe that leveraging classical computational photography techniques and physics-based simulations to generate training data for generative models, is a promising direction for physically-based image editing.
\end{hlbreakable}

\begin{acks}
We thank Kiran Murthy for his valuable contribution and guidance in tone-mapping linear color images. We are also very grateful to Alberto García García, Erroll Wood, Jesús Pérez and Iker J. de los Mozos, for their help and contributions to the synthetic rendering effort. Finally, we also thank Dani Lischinski, Andrey Voynov, Bar Cavia, Tal Reiss, Daniel Winter, Yarden Frenkel, Asaf Shul, Matan Cohen, Julian Iseringhausen, Francois Bleibel, Chloe LeGendre, Dani Cohen-Or and Or Patashnik for discussion and their support that aided and improved our work.
\end{acks}

\input{assets/figures/stop_motion}
\
\clearpage
\bibliographystyle{ACM-Reference-Format}
\bibliography{99_bibliography}

\appendix
\renewcommand\thefigure{\thesection.\arabic{figure}}
\setcounter{figure}{0}

\hl{\section{Generating Data for Illumination Control}}

\hl{\subsection{Synthetic Rendering}}

\begin{figure*}[ht]
    \centering
    \setlength{\tabcolsep}{1.5pt}
    {\small
        \includegraphics[width=0.4\textwidth]{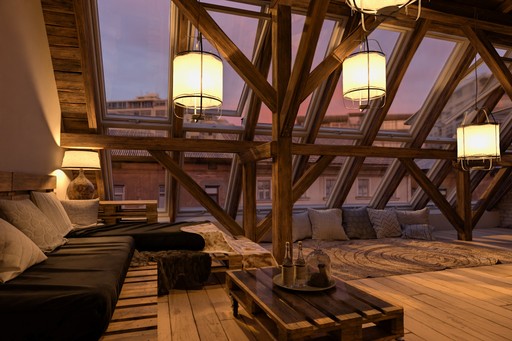}
        \includegraphics[width=0.4\textwidth]{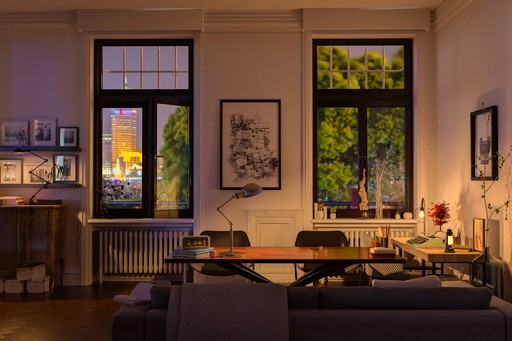}
        \includegraphics[width=0.4\textwidth]{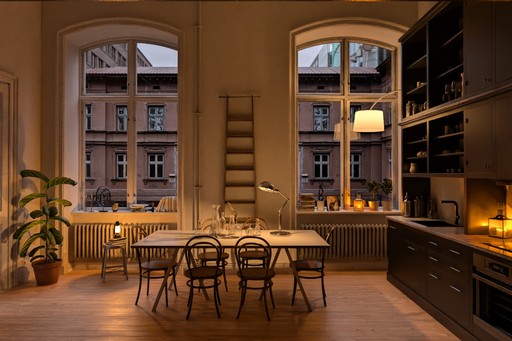}
        \includegraphics[width=0.4\textwidth]{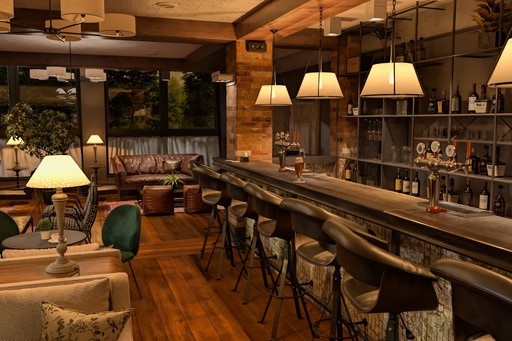}
    }   
    \caption{\hl{\textbf{3D scenes.} Indoor scene examples used to generate our synthetic dataset.}}
    \label{fig:3d_scenes}
\end{figure*}

\hl{To procedurally generate synthetic images as described in Section 3.1, we use a small set of indoor scenes from publicly available 3D asset stores. The set of scenes we use represent common indoor environments, e.g. living room, kitchen, bedroom, bar, restaurant, office, etc., complete with furniture and accessories. The scenes have physically plausible textured materials and can be rendered in Blender / Cycles, producing photorealistic results, as shown in Figure~\ref{fig:3d_scenes}.}

\hl{We start by manually annotating the scenes, identifying specific scene objects that are plausible camera view targets, i.e. kitchen countertop, coffee table, bar counter, couch, side table, chair, etc. We also identify existing light fixture geometry (light bulbs’ glass, spot lights’ lens) and meshes onto which additional light fixture models can be randomly placed (floor, tables, shelves, furniture top). We also identify the floor geometry to validate the randomized camera placement described next.}

\hl{To create randomized and plausible camera views with targeted content, we sample an object from the set of annotated view target objects and randomly place the camera view around that object given a range of valid camera distances, field of view angle, and polar and azimuth angles w.r.t. the center of the object and the up axis. The sampled candidate camera position is then tested to be above the floor geometry and additional visibility tests are performed to make sure the camera is not randomly placed outside of the scene.}

\hl{Once a valid camera view is defined, the lighting is setup procedurally, creating virtual light sources corresponding to existing annotated light fixtures and to the randomly placed additional light fixtures. Light sources that are not seen directly by the camera are then discarded and we sample the remaining light sources' parameters (intensity, color temperature, area size, spot angle, etc.) within pre-defined ranges.}

\hl{Finally, we render the scene once per light source, with only the corresponding light source switched on and all other lights switched off. We also render the scene several more times, with all light sources switched off, each time using a different distant HDRI environment map. This produces several images of that same scene, lit with one light at a time and with a diverse set of outdoors lighting scenarios casting ambient lighting through windows and other openings.}

\hl{The images are saved using a full float precision image format, in linear color space, without applying any tone-mapping, and passed down to the image post-processing pipeline described in Section 3.2.}

\vspace{1em}
\section{Image Based Relighting}
\vspace{-2em}

\hl{\subsection{Post-Capture Calibration}} \label{appendix:disentanglement}
In this section we provide additional details on the processing of the raw photograph pairs. 
We demosaic each image pair with respect to the capturing camera color filter array. 
\hl{Our captured dataset was comprised of raw image pairs, captured without any specific photography expertise, using standard off-the-shelf mobile devices (Pixel 9 and Samsung S24)}.
\hl{The images were captured using each device's auto-exposure} to ensure they are well-exposed and to match the distribution of source images captured in-the-wild.

\hl{To correctly perform manipulation on pairs of images captured on the same device but under different exposure setting, we calibrate them using the raw image meta-data during post-capture.}
We empirically find that linearly interpolating white balance gains according to the target light source intensity produces good results.
For color correction, we opt to use the camera's color correction matrix extracted from the raw image with the light on.
Since each image in a pair is captured with different exposures, we calibrate it according to the product $P$ of the exposure value $E$, the applied analog and digital gain $G$, $D$ respectively.
To un-normalize the relit images exposure, we linearly interpolated the two products element-wise, according to the chosen target light strength $\gamma$ and ambient strength $\alpha$:
\begin{multline}
    P_\text{relit} = \left( \left( 1 - \gamma \right) \alpha E_\text{off} + \gamma E_\text{on} \right) 
    \left( \left( 1 - \gamma \right) \alpha G_\text{off} + \gamma G_\text{on} \right) \\
    \left( \left( 1 - \gamma \right) \alpha D_\text{off} + \gamma D_\text{on} \right)
\end{multline}

\begin{figure*}
    \centering
    {\small
        \includegraphics[width=0.87\textwidth]{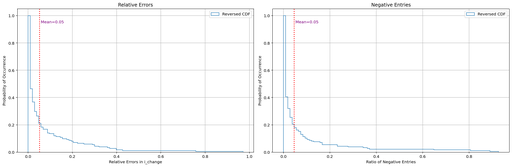}
    }   
    \caption{\hl{\textbf{Calibration Errors.} Error magnitudes in the target light residual $\mathbf{i}_\text{change} = \mathbf{i}_\text{on} - \mathbf{i}_\text{off}$ for captured photograph pairs in our dataset.}}
    \label{fig:calibration_err_hist}
\end{figure*}

\begin{hlbreakable}
As explained in Section 3.2, under a theoretically perfect capture process the target light residual $\mathbf{i}_\text{on} - \mathbf{i}_\text{off}$ is non-negative, corresponding with an addition of light to the scene. In Figure~\ref{fig:calibration_err_hist} we consider the negative component $\textbf{e} = \text{clip}(\mathbf{i}_\text{on} - \mathbf{i}_\text{off}, -\infty, 0)$ as the error residual and the positive component $\textbf{d} = \text{clip}(\mathbf{i}_\text{on} - \mathbf{i}_\text{off}, 0, \infty)$ as a clean approximation. We measure both the relative error $\frac{\lVert e \rVert_2}{\lVert d \rVert_2}$, and the percentage of negative entries in $\mathbf{i}_\text{on} - \mathbf{i}_\text{off}$ across our captured dataset, and report the statistics in Figure~\ref{fig:calibration_err_hist}.
\end{hlbreakable}

\begin{figure*}[ht]
    \centering
    \setlength{\tabcolsep}{1.5pt}
    {\small
    \begin{tabular}{c c c c c c}
         \multirow{4}{*}[6mm]{\rotatebox{90}{\textbf{Ambient Light Intensity}}} &
        \raisebox{0.112\textwidth}{1.0} &
        \includegraphics[width=0.23\textwidth]{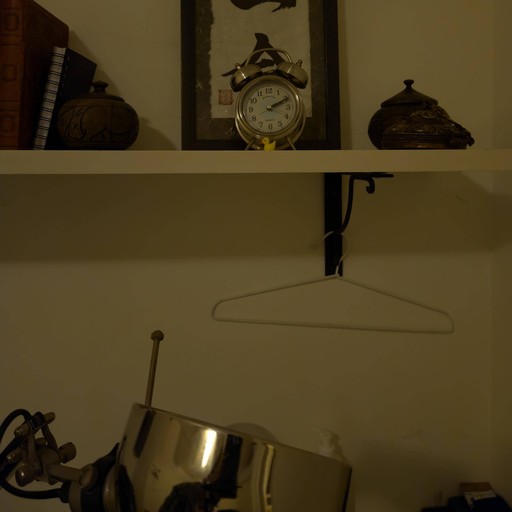} &
        \includegraphics[width=0.23\textwidth]{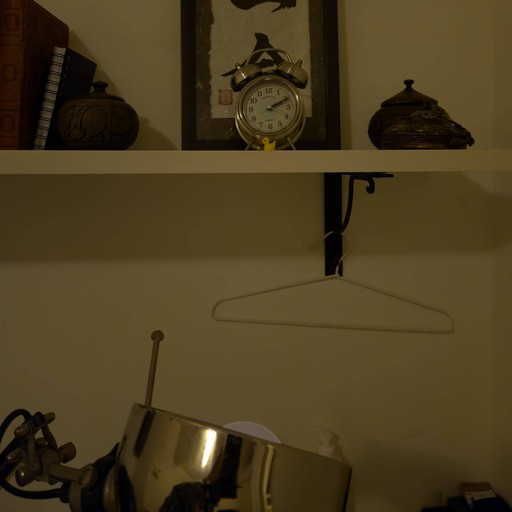} &
        \includegraphics[width=0.23\textwidth]{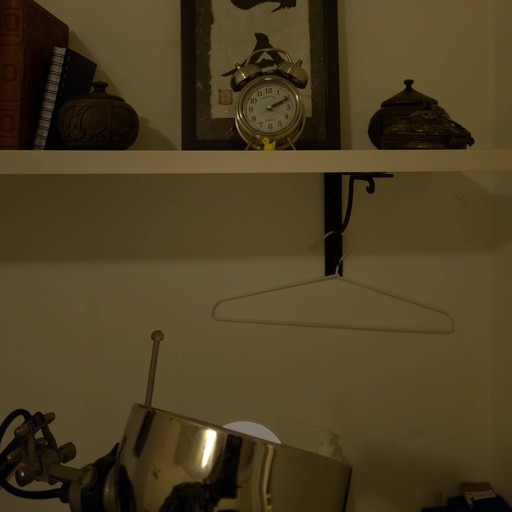} &
        \includegraphics[width=0.23\textwidth]{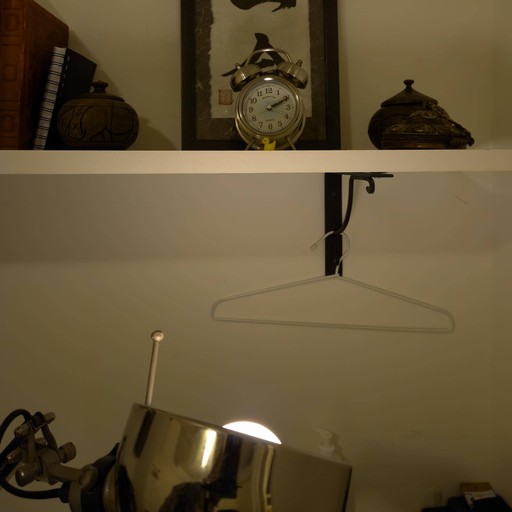} \\

        ~ & \raisebox{0.112\textwidth}{0.5} &
        \includegraphics[width=0.23\textwidth]{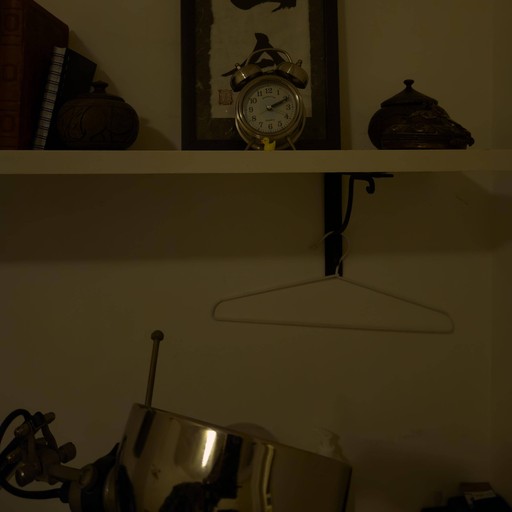} &
        \includegraphics[width=0.23\textwidth]{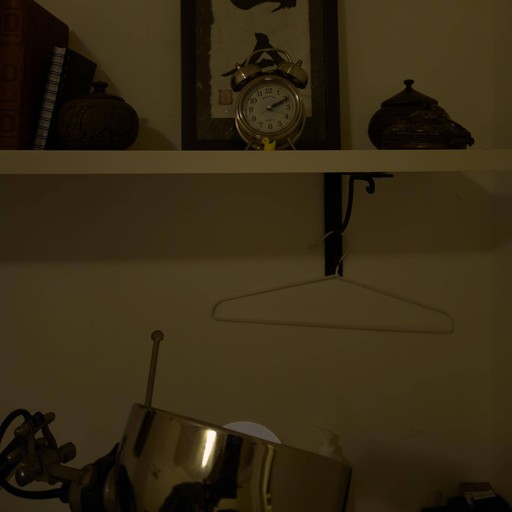} &
        \includegraphics[width=0.23\textwidth]{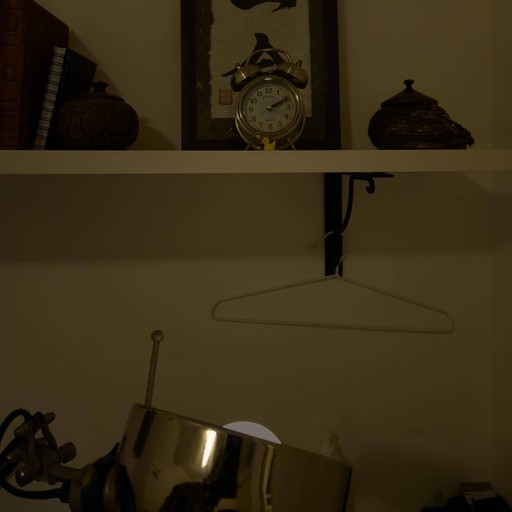} &
        \includegraphics[width=0.23\textwidth]{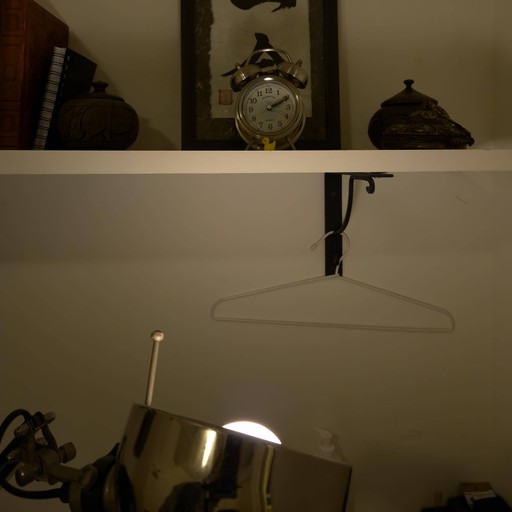} \\

        ~ & \raisebox{0.112\textwidth}{0.14} &
        \includegraphics[width=0.23\textwidth]{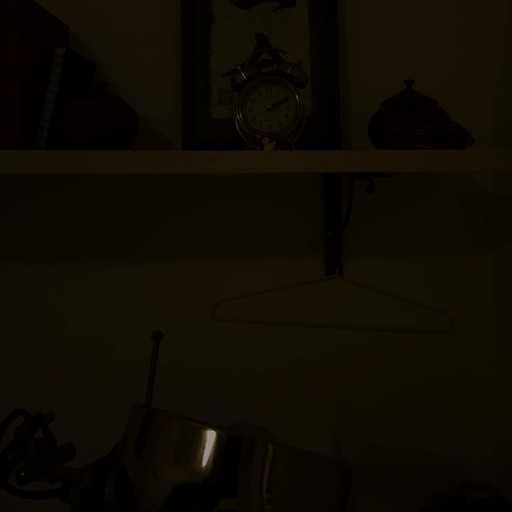} &
        \includegraphics[width=0.23\textwidth]{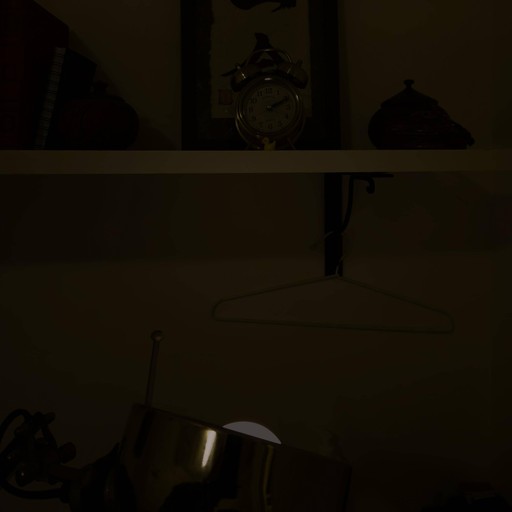} &
        \includegraphics[width=0.23\textwidth]{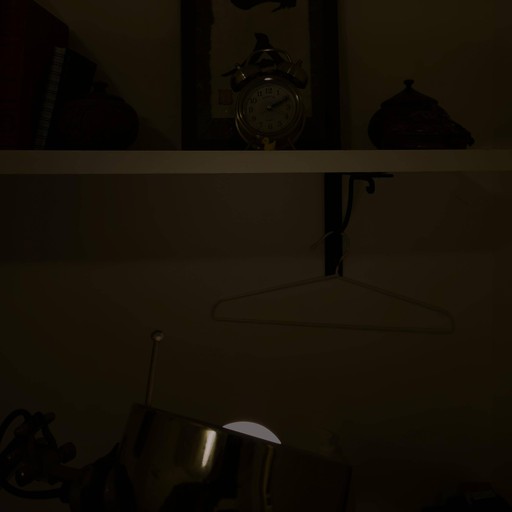} &
        \includegraphics[width=0.23\textwidth]{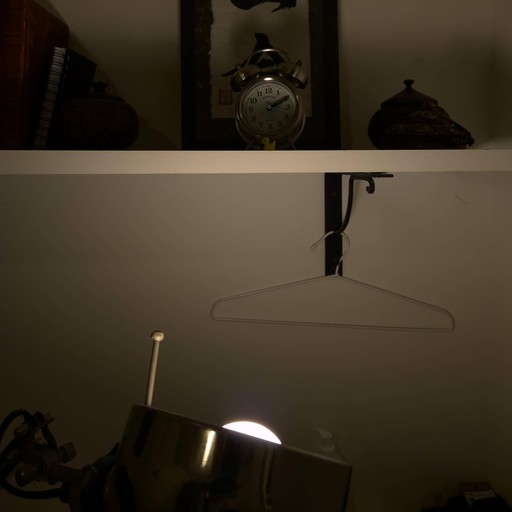} \\
        
         ~ & ~ & 0.0 & 0.3 & 0.7 & 1.0 \\
         ~ & ~ & \multicolumn{4}{c}{\textbf{Light Source Intensity}} \\
        
    \end{tabular}
    }   
    \caption{\textbf{Post-process examples.} A parametric sequence of images generated from a single raw photograph pair in the post-processing procedure in Section 3.2}
\end{figure*}

\section{Light Source Control with Diffusion Models}

\begin{hlbreakable}
\subsection{Parametric Conditioning}
This section provides additional details on the conditioning scheme introduced in Section 3.4.
We divide condition signals to spatial (localized) inputs and global conditions (defined to have spatially constant value). 
\paragraph{\textbf{Spatial Conditions.}} These include the input (source) image, a depth map, a segmentation mask of the target light source(s), relative light intensity change and the target light's RGB color.
The input image is passed through the model's VAE, reducing it's spatial dimensions to the diffusion models latent spatial dimensions $H \times W \times 4$. 
The scaled intensity change mask, the target color RGB mask and the depth map are resized to match the spatial dimensions of the latent input noise.
Spatial conditions are stacked together, overall forming a tensor with dimensions $H\times W\times 9$ merged via a learned $1\times1$ (zero-initialized) convolution layer, which reduces the number of channels to the base model's $H \times W \times 4$. This tensor is concatenated to the input noise.
\end{hlbreakable}

\begin{hlbreakable}
\subsection{Masking Procedure}
To obtain semantic segmentation masks for the photography capture training set, the "ON" image from each pair was manually annotated with axis aligned bounding boxes which were used to condition the segmentation model \cite{ravi2024sam2segmentimages}.
During inference the same procedure is applied, where a user provides a segmentation mask of the target light source (in the examples presented in the paper a bounding box condition was given to the segmentation model as for the training data).
Note that the human annotation introduces variance to the distribution of segmentation masks on which the model is trained, which mirrors the variance expected by users at inference time.
For the synthetically generated training and evaluation sets, a separate mask is rendered showing the geometry of each light fixture.
\end{hlbreakable}

\subsection{Intensity Sampling}
\label{sec:intensity_sampling}
During training, for each sample we probabilistically choose a light component (light source or ambient lighting). Next, we sample two intensity values from the modified light component, and another intensity value for the unchanged component, all from the precomputed values in the dataset.   
Note that the model is trained in both directions, with intensity change values in the range $[-1,\, 1]$, and term light intensity changes (of either the target light source or ambient light) within ${-1,\, 1}$ as "binary".
As endpoint intensities - where the light is either "on" or "off" - are more common in-the-wild, we increase their weight during training by probabilistically setting either the source image or the target to an endpoint image. 
Importantly, using a relative intensity scale allows for iterative refinement in successive edits, and enables use without an additional radiance estimation stage at inference.

\section{Experiments}
\begin{table}[t]
    \centering
    \footnotesize
    \caption{\textbf{Synthetic augmentations.} ground truth similarity metrics for different synthetic data distributions, computed on the paired synthetic evaluation set.}
    \setlength{\tabcolsep}{1pt}
    \begin{tabular}{
        l@{\hspace{4pt}}
        c@{\hspace{4pt}}
        c@{\hspace{4pt}}
        c@{\hspace{6pt}}
        c@{\hspace{4pt}}
        c@{\hspace{4pt}}
        c@{\hspace{4pt}}
    }
    \toprule
    \multirow{2}{*}{\textbf{Training Datasets}} & \multicolumn{3}{c}{\textbf{PSNR}$\uparrow$} & \multicolumn{3}{c}{\textbf{SSIM}$\uparrow$ } \\
     & Binary           & Intensity           & Color           & Binary           & Intensity           & Color   \\ 
    \midrule

    \handcrafted{} + \procedural{}  & \colorbox{tabfirst}{27.7} & \colorbox{tabfirst}{29.4} & \colorbox{tabfirst}{28.8} & \colorbox{tabfirst}{0.869} & \colorbox{tabfirst}{0.884} & \colorbox{tabfirst}{0.882} \\
    
    \handcrafted{} + \procedural{} (reduced) & \colorbox{tabsecond}{26.4} & \colorbox{tabsecond}{28.7} & \colorbox{tabsecond}{28.5} & \colorbox{tabsecond}{0.845} & \colorbox{tabsecond}{0.872} & \colorbox{tabsecond}{0.878} \\
    
    \handcrafted{} & 25.9 & 28.3 & 27.8 & 0.842 & 0.870 & 0.870 \\

    \procedural{} & 25.8 & 28.1 & 26.4 & \colorbox{tabsecond}{0.845} & 0.867 & 0.866\\

    \bottomrule
    \end{tabular}
    \label{tab:procedural_lightsource}
\end{table}

\subsection{Random scene augmentations} 
We test whether extending the synthetic dataset by adding light sources at random plausible locations improves downstream performance, focusing on synthetic data. 
Note that this augmentation changes the scene light transport, as opposed to rendering the same scene from multiple views. 
We duplicate light sources that are already present in the original scenes in order to avoid increasing light source or material diversity for this experiment. 
Table \ref{tab:procedural_lightsource} shows that while using only augmented scenes yields a lower ground truth similarity, incorporating both original and augmented scenes improves ground truth similarity across all settings.
To test whether this improvement is a direct result of increasing the sample size, we match the size of the original and augmented datasets, using half of each. The results in Table~\ref{tab:procedural_lightsource} (second line from the top) show a noticeable improvement across tasks.

\subsection{Effect of Synthetic}
\input{assets/figures/effect_of_synth_dining}
\begin{hlbreakable}
This section provides additional qualitative examples demonstrating the effect of adding the synthetic data to the training data.
Figure~\ref{fig:domains_qualitative_appendicies} demonstrates domain leakage as indicated by synthetic-like lighting showing a lack of over-saturated halos around the light source, which are present in the model trained only on captured images (first two examples).
\end{hlbreakable}

\subsection{Comparisons}
\begin{hlbreakable}
In this section we provide additional results and details on the IIW dataset and on our paired evaluation dataset.
\paragraph{\textbf{\rgbx Implementation Details.}} The RGB-X model itself was used to compute all conditions from the input and ground truth images (RGB->X), we then fed back the input conditions (X->RGB) while replacing the irradiance condition with the ground truth irradiance
\end{hlbreakable}
\begin{figure*}[t]   
   \centering
    \includegraphics[width=\textwidth]{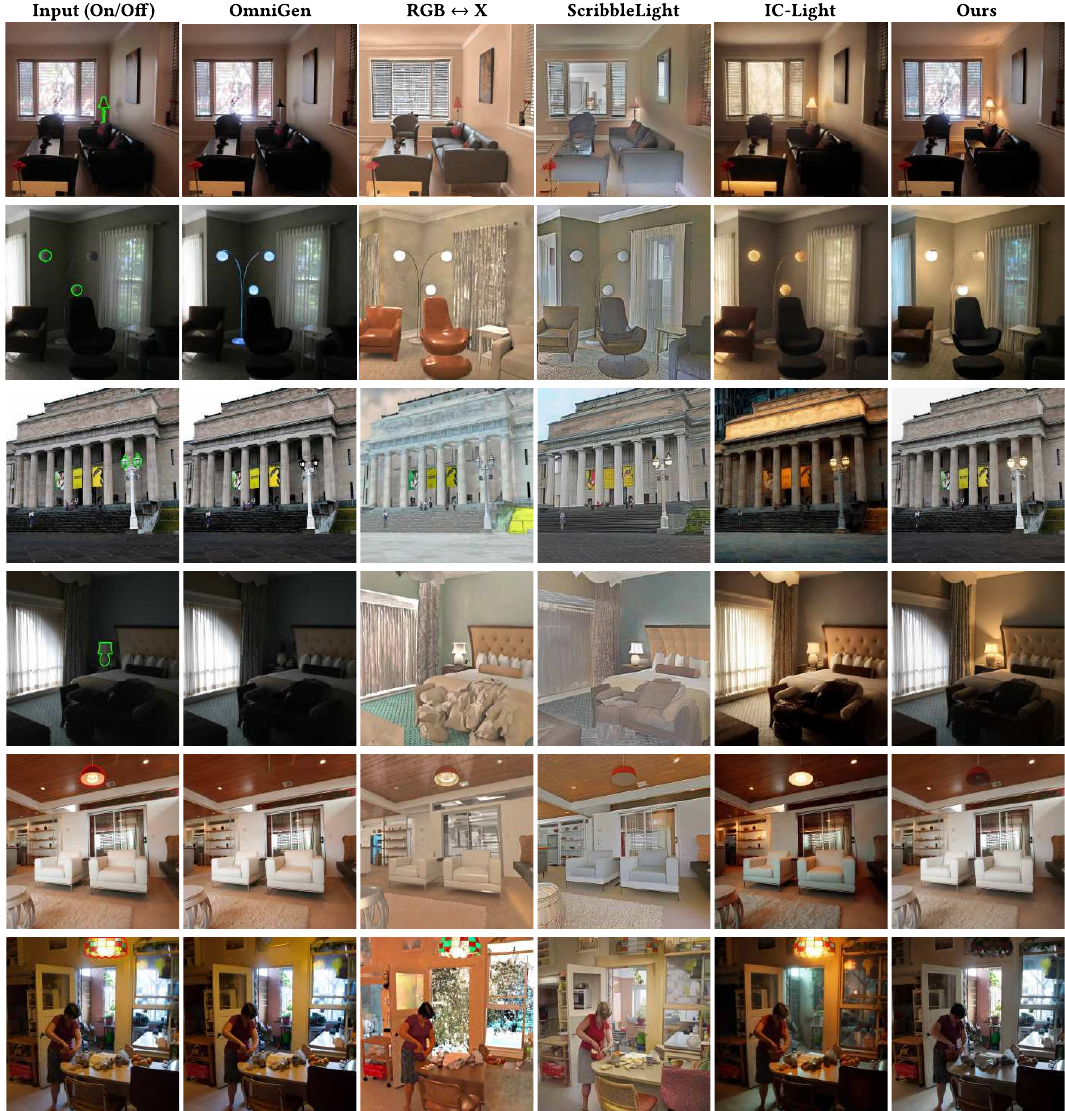}
    \caption{\textbf{Qualitative Comparisons.} Additional comparisons of binary light switch on indoor images from IIW dataset. On the left, the input image where \textcolor{green}{green} / \textcolor{red}{red} contours specify which light source(s) should be turned \textcolor{green}{on} / \textcolor{red}{off} respectively.
    \hl{The results were computed with the "together" tone-mapping condition and without color conditioning.}
    }
    \label{fig:comparisons_iiw_sup1}
\end{figure*}

\begin{figure*}[t]   
    \includegraphics[width=\textwidth]{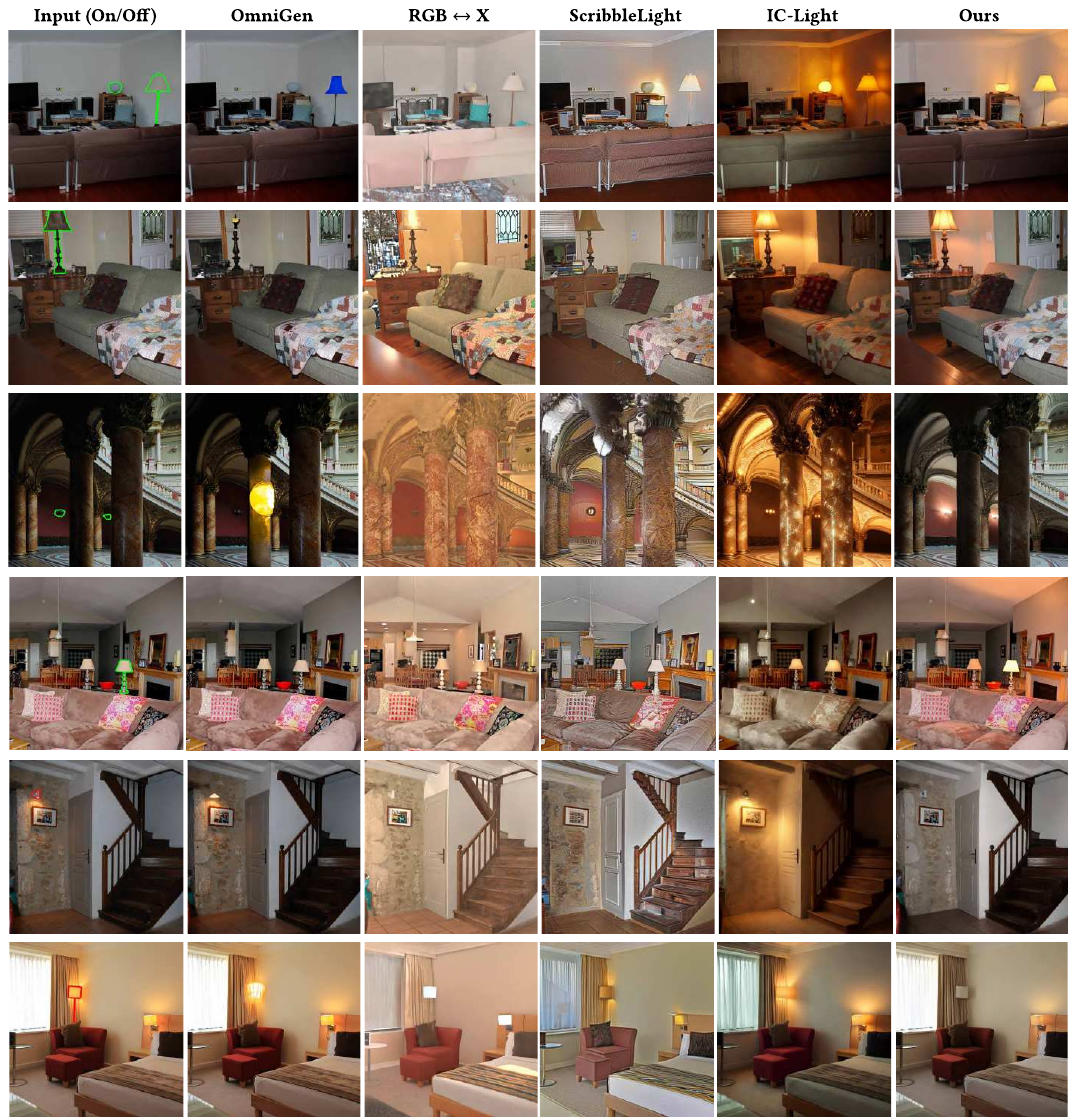}
    \caption{\textbf{Qualitative Comparisons.} Additional comparisons of binary light switch on indoor images from IIW dataset. On the left, the input image where \textcolor{green}{green} / \textcolor{red}{red} contours specify which light source(s) should be turned \textcolor{green}{on} / \textcolor{red}{off} respectively.
    \hl{The results were computed with the "together" tone-mapping condition and without color conditioning.}
    }
    \label{fig:comparisons_iiw_sup2}
\end{figure*}

\newcommand{\evalimwidth}{0.192\textwidth}

\begin{figure*}[t!]
\centering
\setlength{\tabcolsep}{1.5pt}
\begin{tabular}{c c c c c}
    \textbf{Input (On / Off)} & \textbf{Light Intensity Mask} & \textbf{Depth} & \textbf{Result} & \textbf{Target (Ground Truth)} \\
    \includegraphics[width=\evalimwidth]{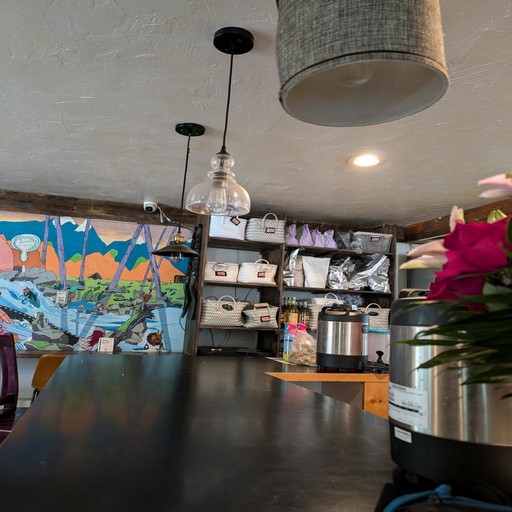} & 
    \includegraphics[width=\evalimwidth]{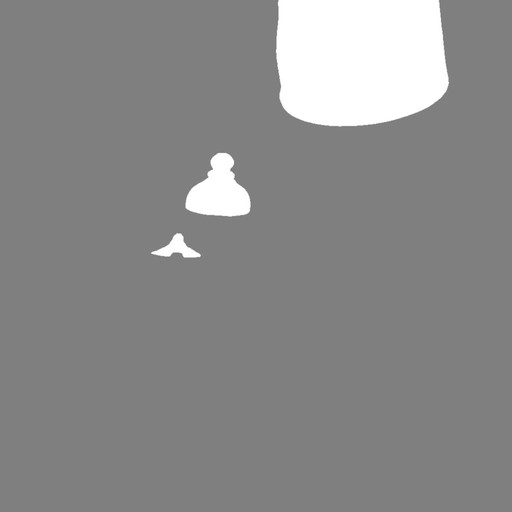} &
    \includegraphics[width=\evalimwidth]{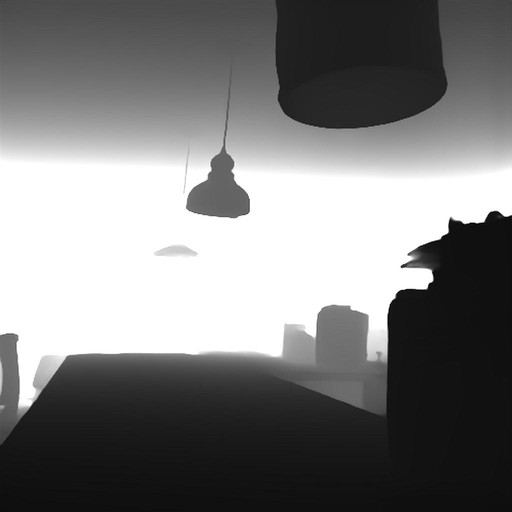} &
    \includegraphics[width=\evalimwidth]{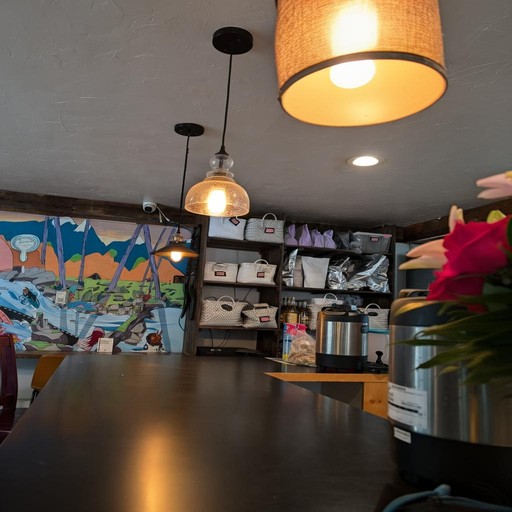} & \includegraphics[width=\evalimwidth]{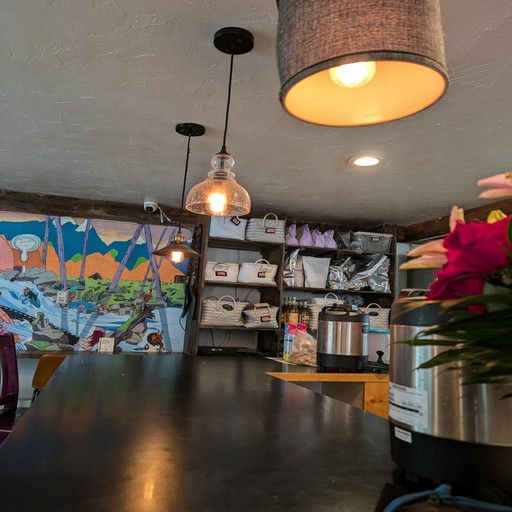} \\
    \includegraphics[width=\evalimwidth]{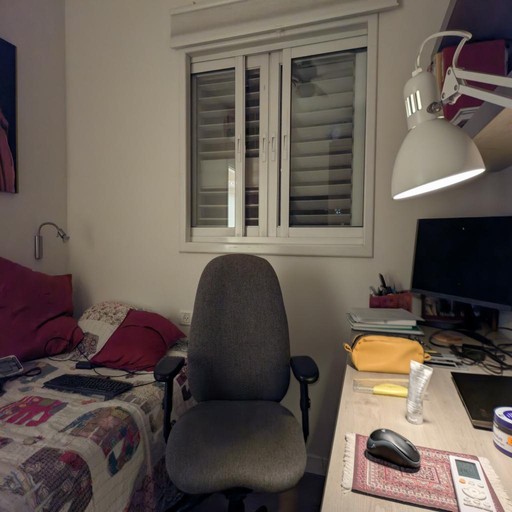} & 
    \includegraphics[width=\evalimwidth]{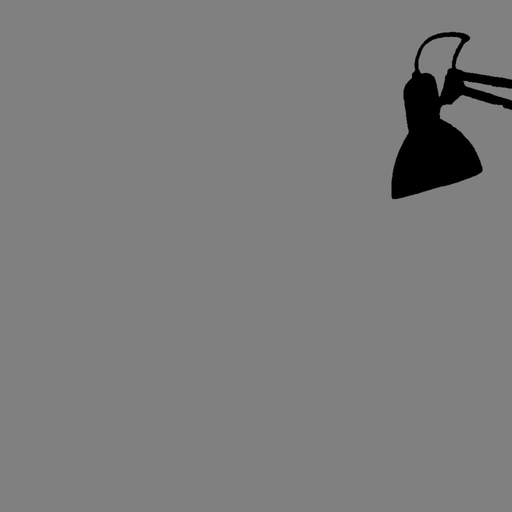} &
    \includegraphics[width=\evalimwidth]{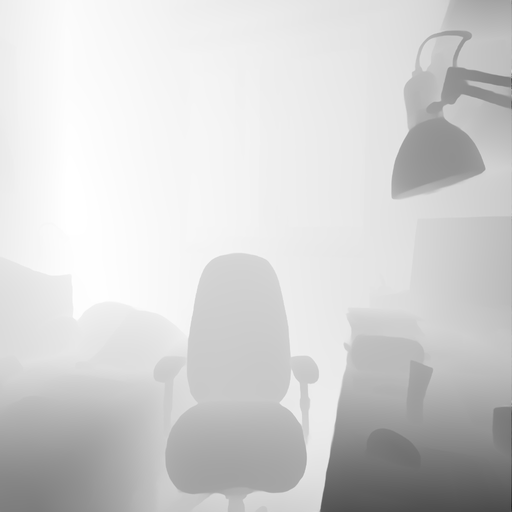} &
    \includegraphics[width=\evalimwidth]{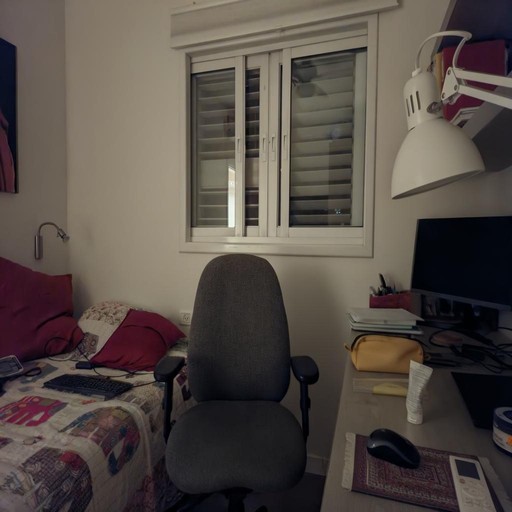} & \includegraphics[width=\evalimwidth]{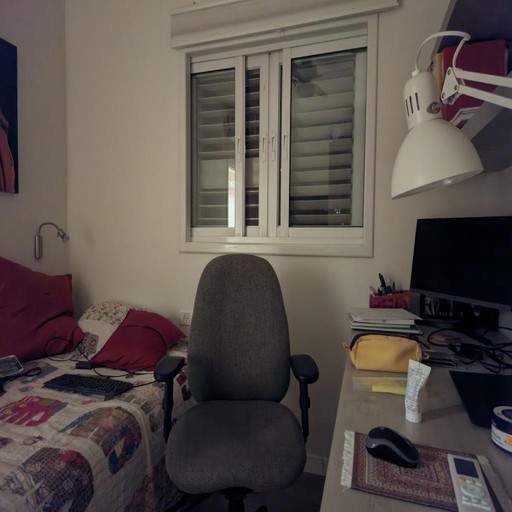} \\
    
    \includegraphics[width=\evalimwidth]{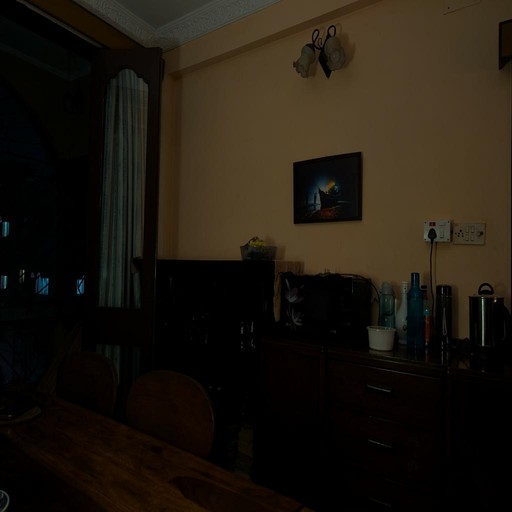} & 
    \includegraphics[width=\evalimwidth]{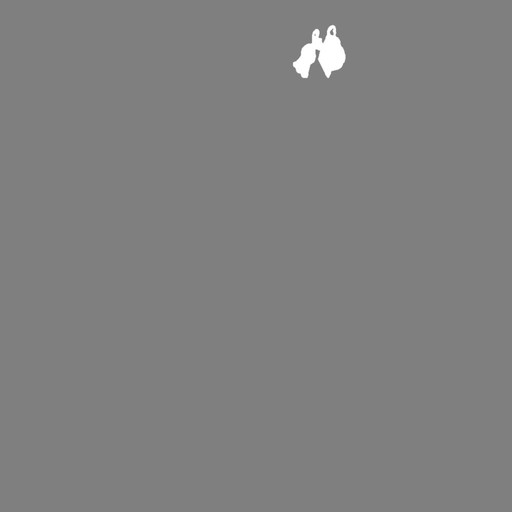} &
    \includegraphics[width=\evalimwidth]{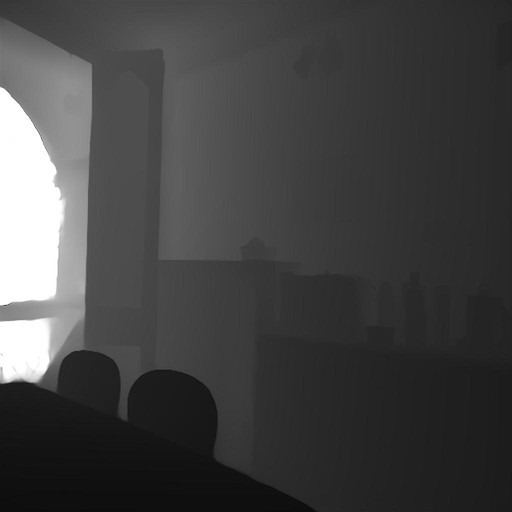} &
    \includegraphics[width=\evalimwidth]{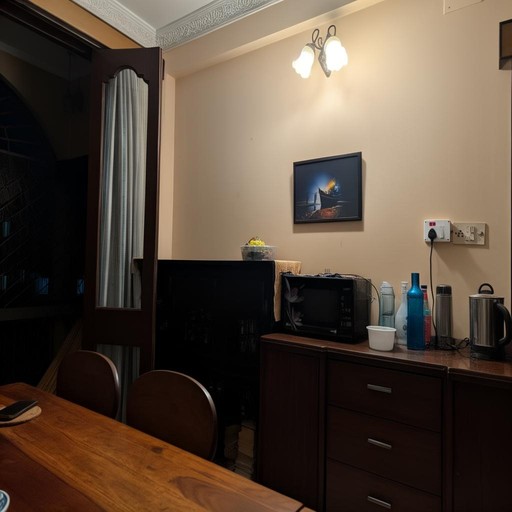} & \includegraphics[width=\evalimwidth]{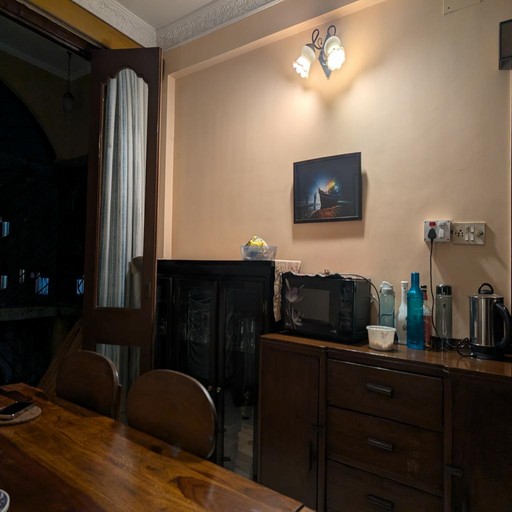} \\
    
    \includegraphics[width=\evalimwidth]{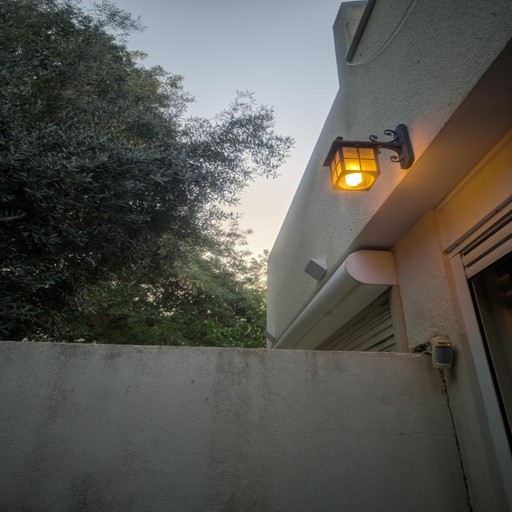} & 
    \includegraphics[width=\evalimwidth]{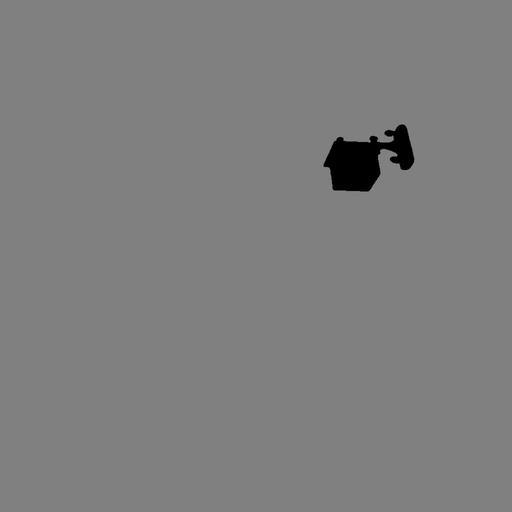} &
    \includegraphics[width=\evalimwidth]{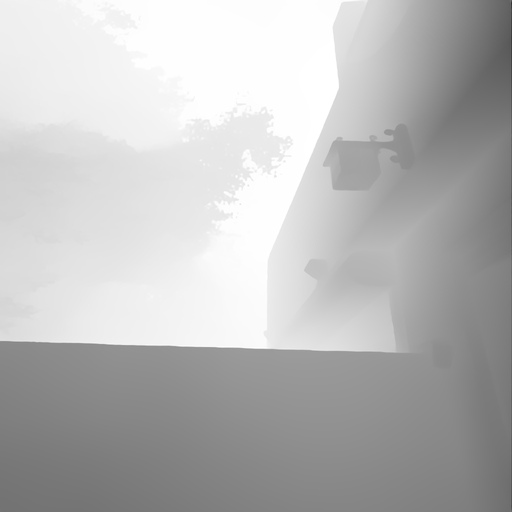} &
    \includegraphics[width=\evalimwidth]{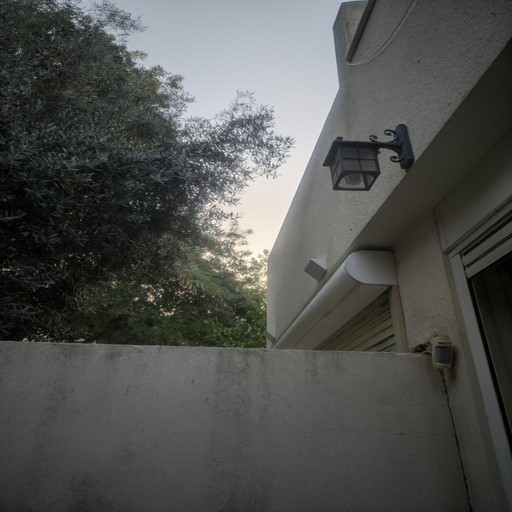} & \includegraphics[width=\evalimwidth]{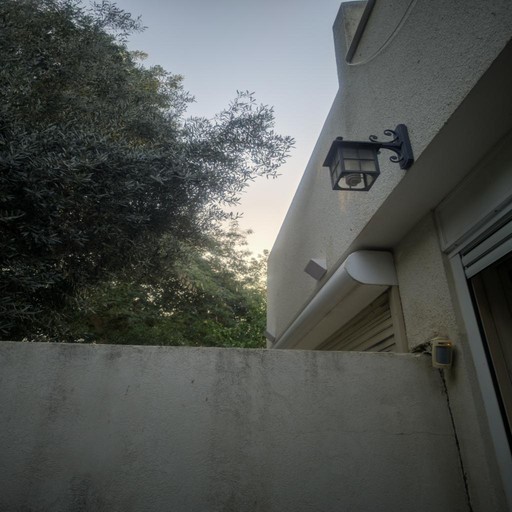} \\

    \includegraphics[width=\evalimwidth]{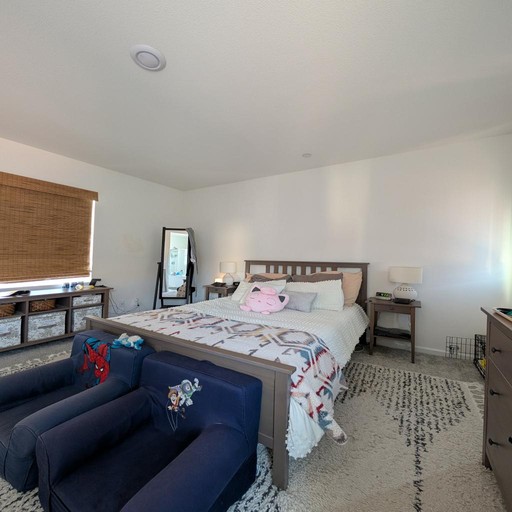} & 
    \includegraphics[width=\evalimwidth]{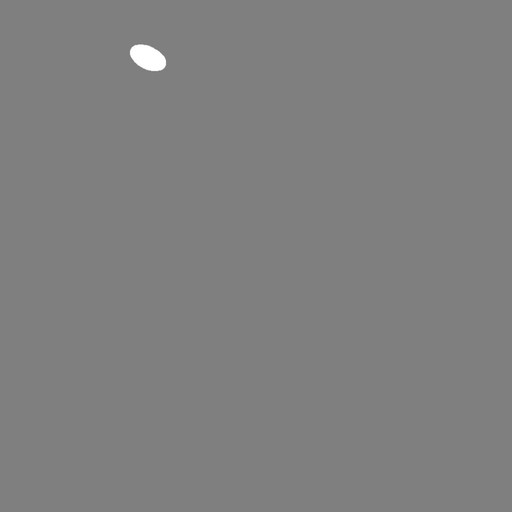} &
    \includegraphics[width=\evalimwidth]{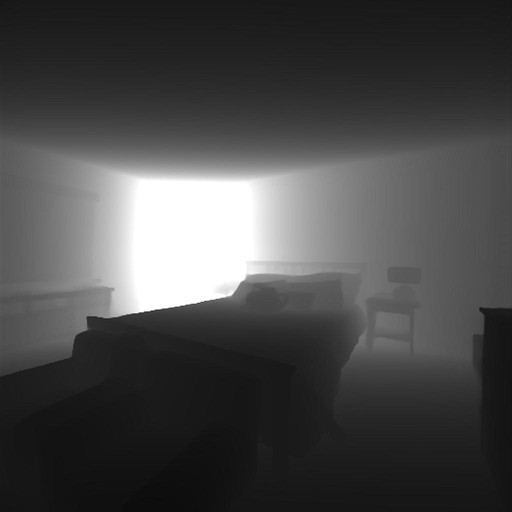} &
    \includegraphics[width=\evalimwidth]{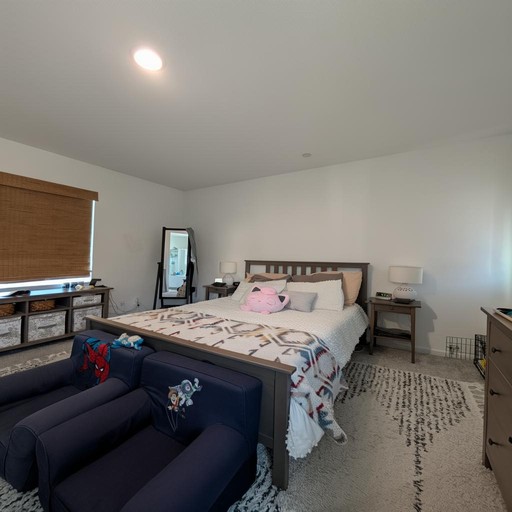} & \includegraphics[width=\evalimwidth]{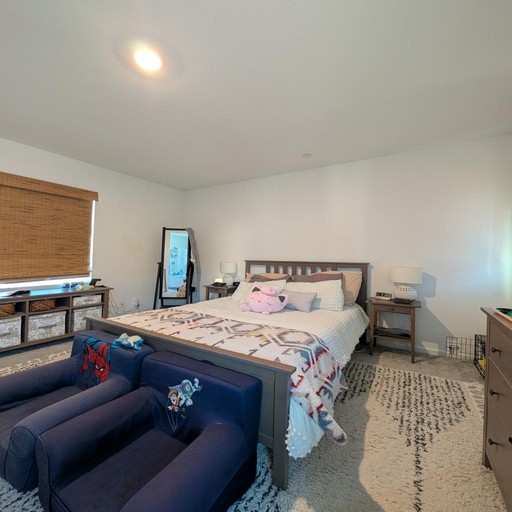} \\
\end{tabular}
\caption{\hl{\textbf{Results on Paired Captured Evaluation Dataset}: A sample of inference results on the binary task, used in the quantitative ablation in Table 1 in the main paper. \textbf{Light Intensity Mask.} the target light intensity mask was scaled and translated from the range [-1,1] to the range [0,1] for display purposes. Gray pixels represent zeros, white represent 1 (turn light on) and black represent -1 (turn light off). The results were generated with the tone-map condition "together", with no ambient change (0 value) and without color conditions (0 values).
}}
\label{fig:eval_results}
\end{figure*}

\begin{figure*}[t!]
\centering
\setlength{\tabcolsep}{1.5pt}
\begin{tabular}{c c c c c}
    \textbf{Input (On / Off)} & \textbf{Light Intensity Mask} & \textbf{Depth} & \textbf{Result} & \textbf{Target (Ground Truth)} \\
    \includegraphics[width=\evalimwidth]{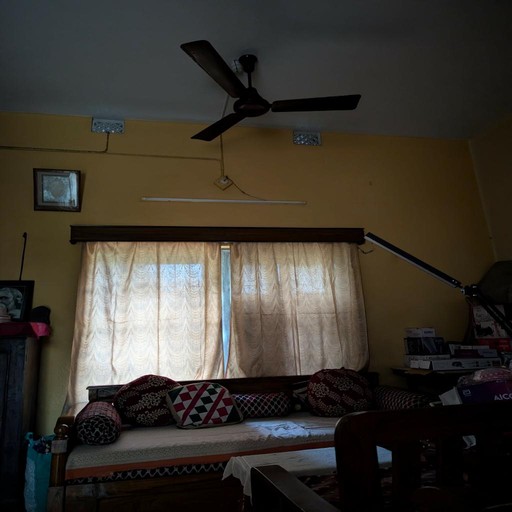} & 
    \includegraphics[width=\evalimwidth]{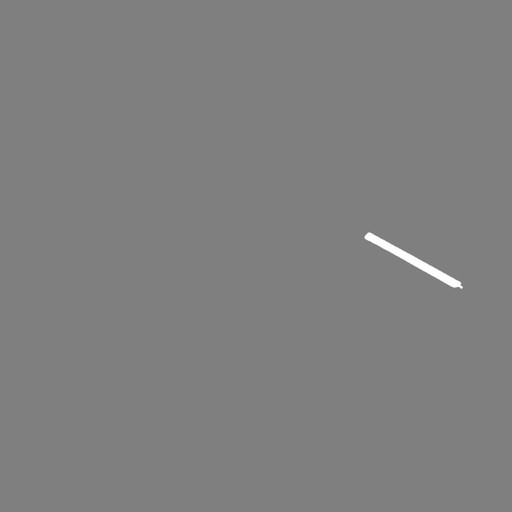} &
    \includegraphics[width=\evalimwidth]{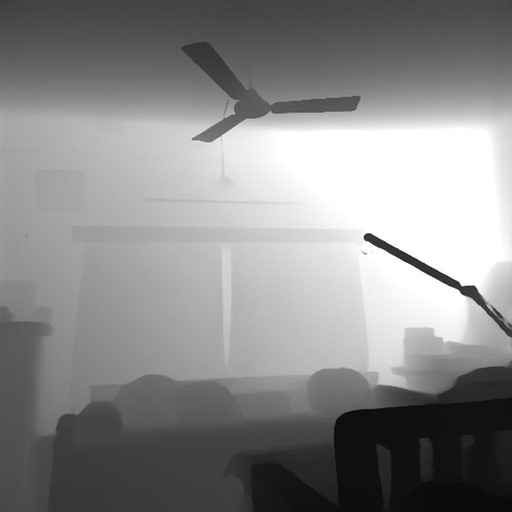} &
    \includegraphics[width=\evalimwidth]{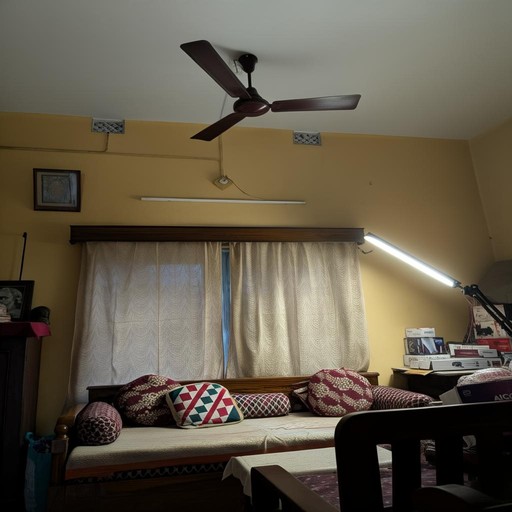} & \includegraphics[width=\evalimwidth]{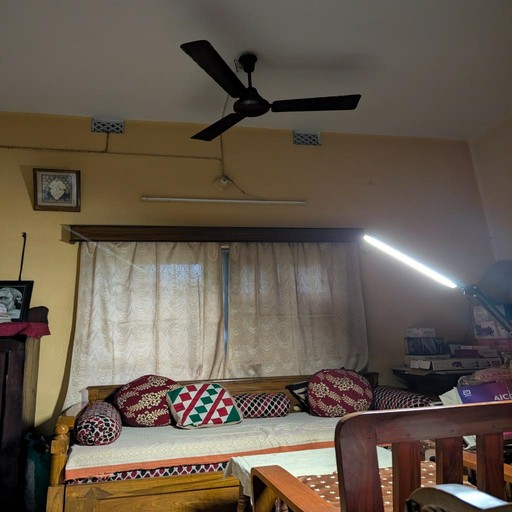} \\
    
    \includegraphics[width=\evalimwidth]{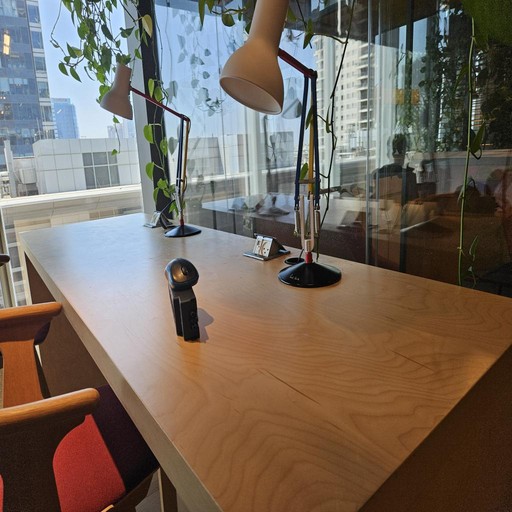} & 
    \includegraphics[width=\evalimwidth]{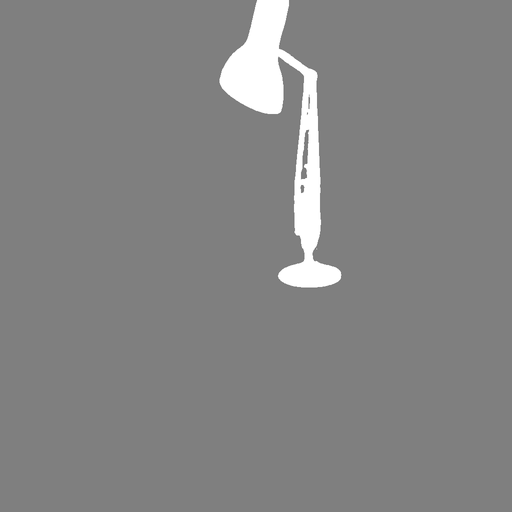} &
    \includegraphics[width=\evalimwidth]{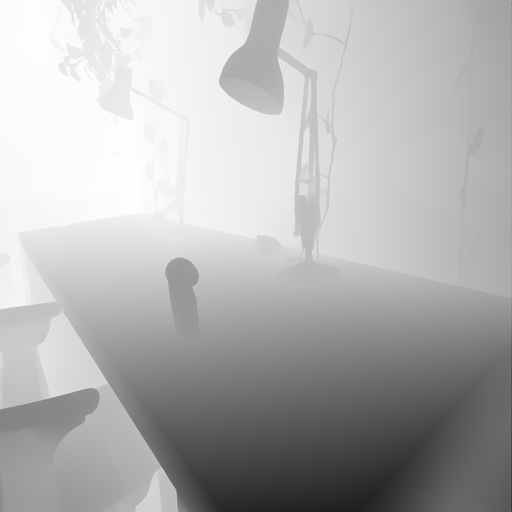} &
    \includegraphics[width=\evalimwidth]{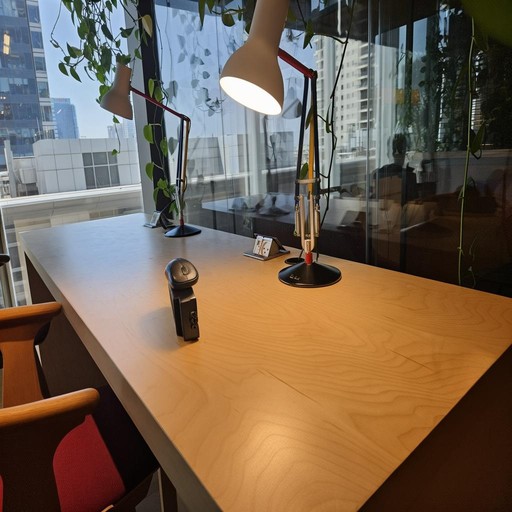} & \includegraphics[width=\evalimwidth]{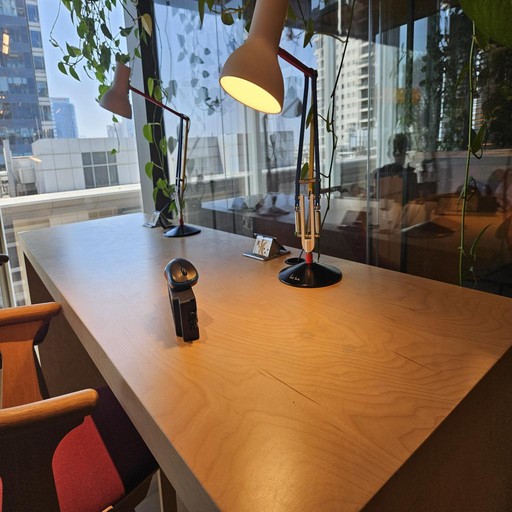} \\
    
    \includegraphics[width=\evalimwidth]{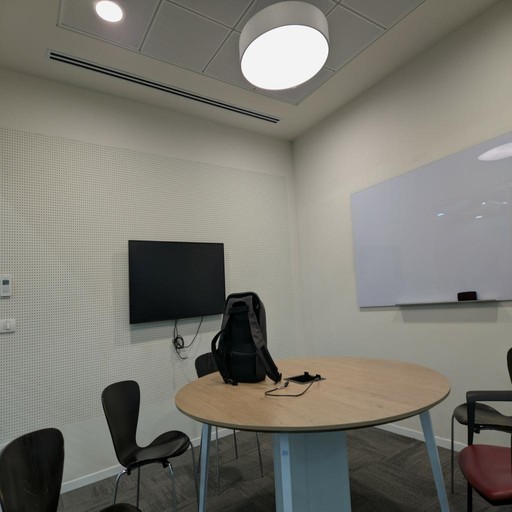} & 
    \includegraphics[width=\evalimwidth]{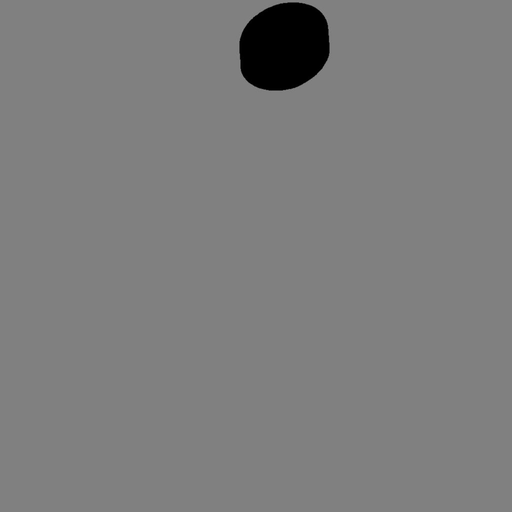} &
    \includegraphics[width=\evalimwidth]{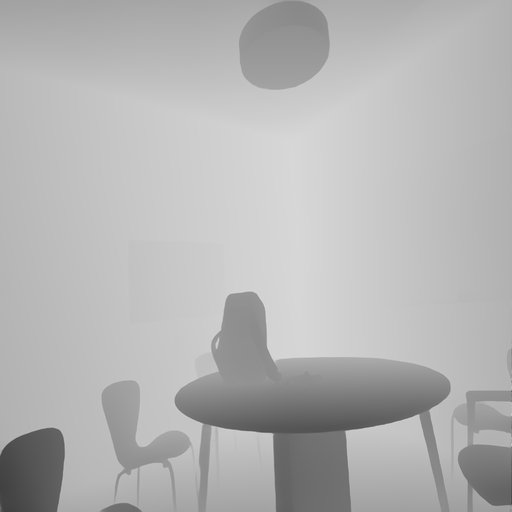} &
    \includegraphics[width=\evalimwidth]{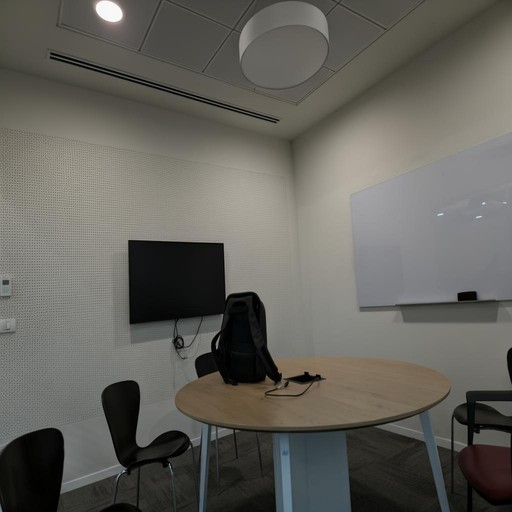} & \includegraphics[width=\evalimwidth]{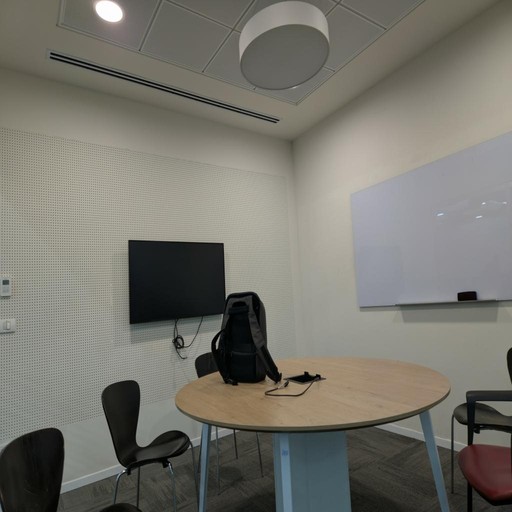} \\
    
    \includegraphics[width=\evalimwidth]{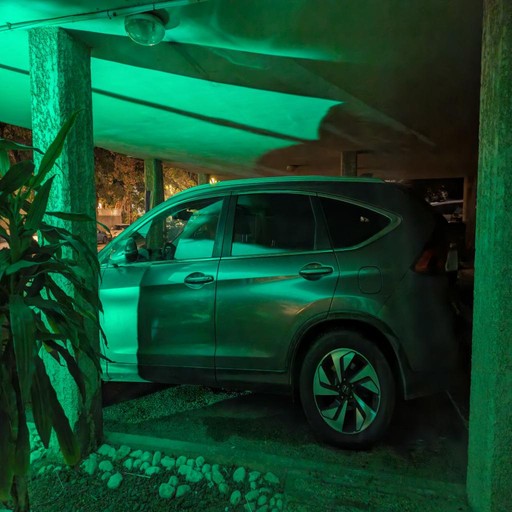} & 
    \includegraphics[width=\evalimwidth]{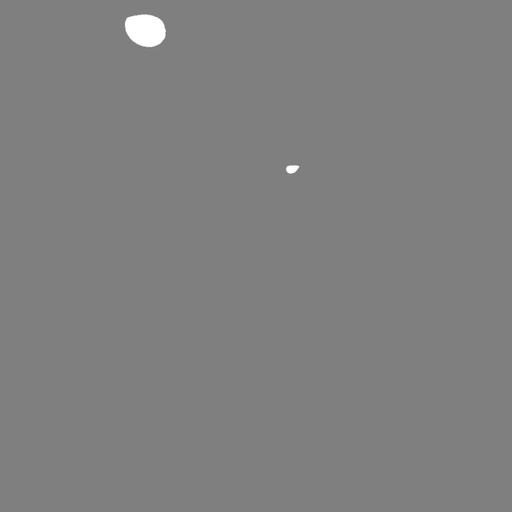} &
    \includegraphics[width=\evalimwidth]{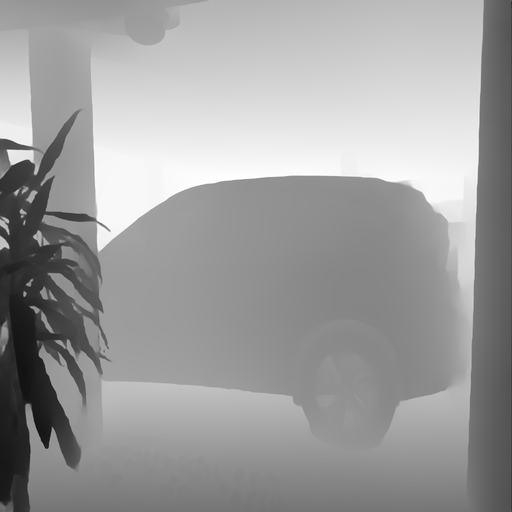} &
    \includegraphics[width=\evalimwidth]{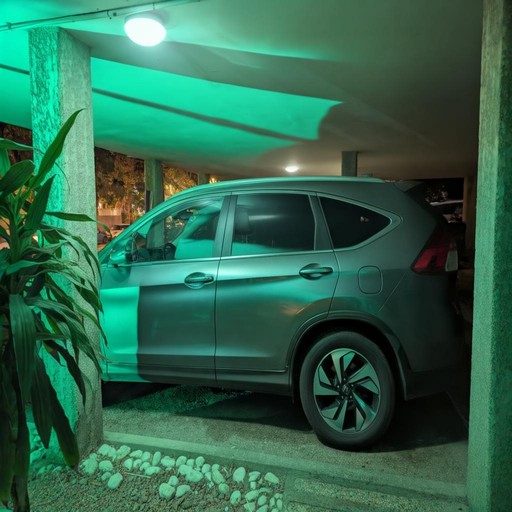} & \includegraphics[width=\evalimwidth]{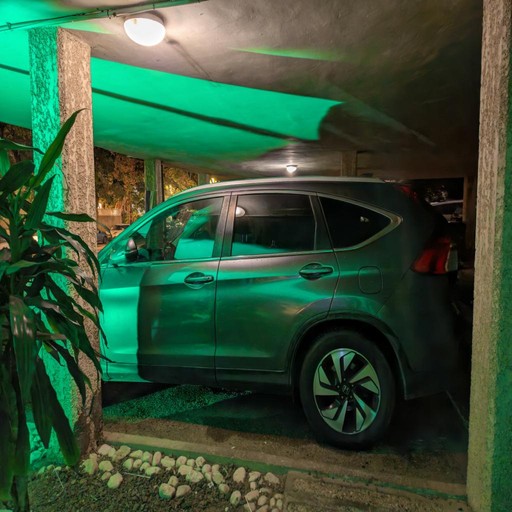} \\

    \includegraphics[width=\evalimwidth]{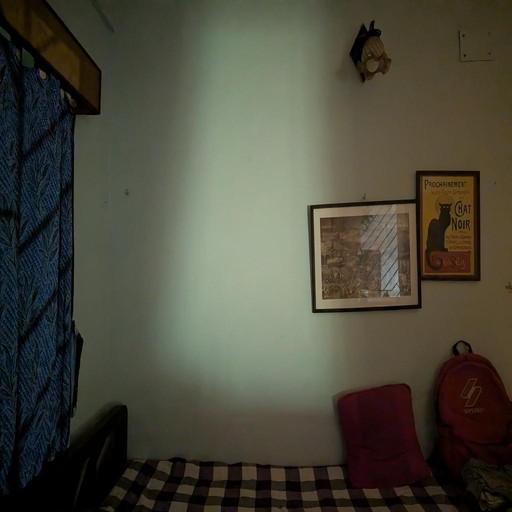} & 
    \includegraphics[width=\evalimwidth]{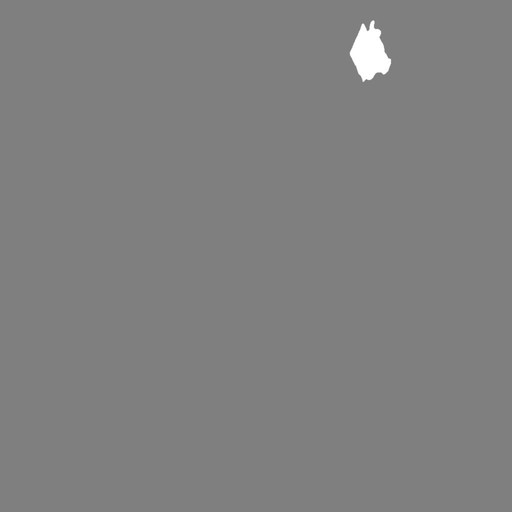} &
    \includegraphics[width=\evalimwidth]{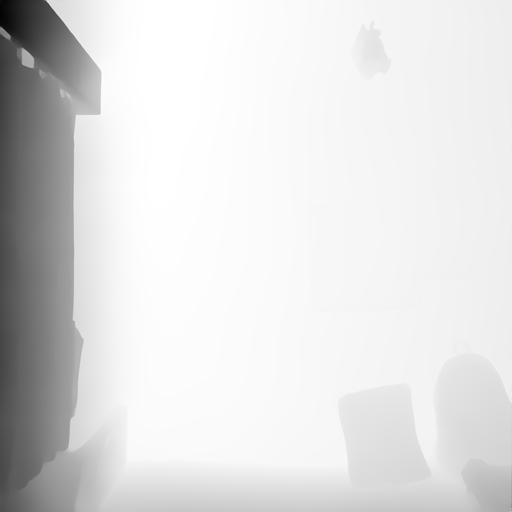} &
    \includegraphics[width=\evalimwidth]{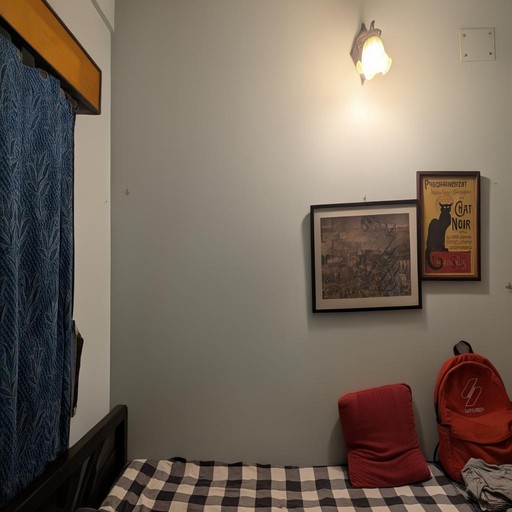} & \includegraphics[width=\evalimwidth]{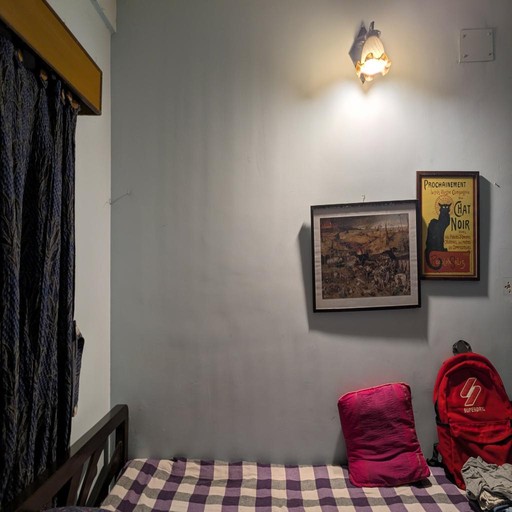} \\
\end{tabular}
\caption{\hl{\textbf{Additional Results on Paired Captured Evaluation Dataset}: A sample of inference results on the binary task, used in the quantitative ablation in Table 1 in the main paper. \textbf{Light Intensity Mask.} the target light intensity mask was scaled and translated from the range [-1,1] to the range [0,1] for display purposes. Gray pixels represent zeros, white represent 1 (turn light on) and black represent -1 (turn light off). The results were generated with the tone-map condition "together", with no ambient change (0 value) and without color conditions (0 values).
}}
\label{fig:eval_results_2}
\end{figure*}

\subsection{User Study} \label{sec:user_study}
To verify that the results of our quantitative and qualitative comparisons, we also conducted a user study over the \textit{in-the-wild} evaluation set. 
In each question, we randomly sampled one of the evaluation examples and one of the comparing methods. We presented to the user the input image, the target light source to turn either on or off, and the editing results both from our and the selected comparing method.
Users were instructed to choose which edited image result depicts a better lighting change, with respect to the input image and other visible and non-visible light sources in it.
Overall, we collected 3200 answers from 100 users using the Amazon Mechanical Turk service. 
The results are summarized in Table 2, where we report the percentage of judgments in our favor compared to each method. As can be seen, most participants favored our method by a large margin.

\newcommand{\outdoorimwidth}{0.24\textwidth}

\begin{figure*}[t!]
\centering
\setlength{\tabcolsep}{1.5pt}
\begin{tabular}{c c c c}
    \textbf{Input} & \textbf{Result} & \textbf{Input} & \textbf{Result} \\
    \includegraphics[width=\outdoorimwidth]{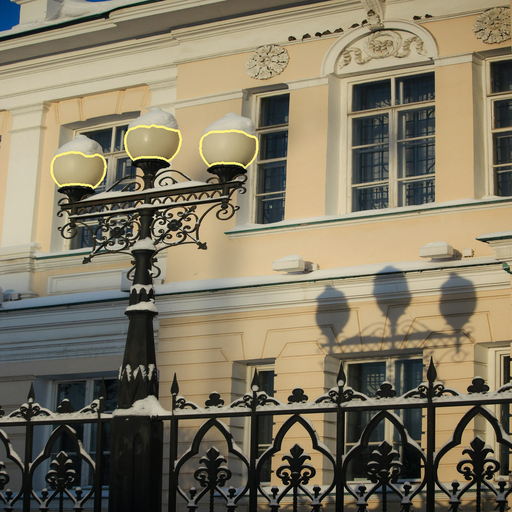} & \includegraphics[width=\outdoorimwidth]{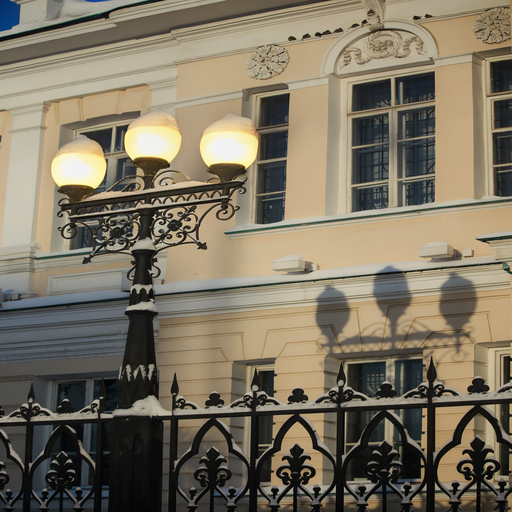} &
    \includegraphics[width=\outdoorimwidth]{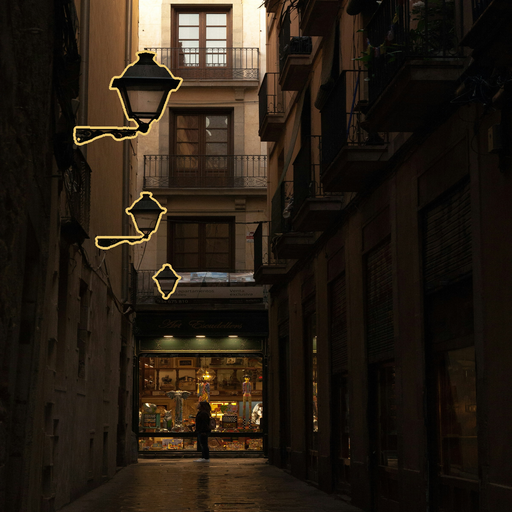} & \includegraphics[width=\outdoorimwidth]{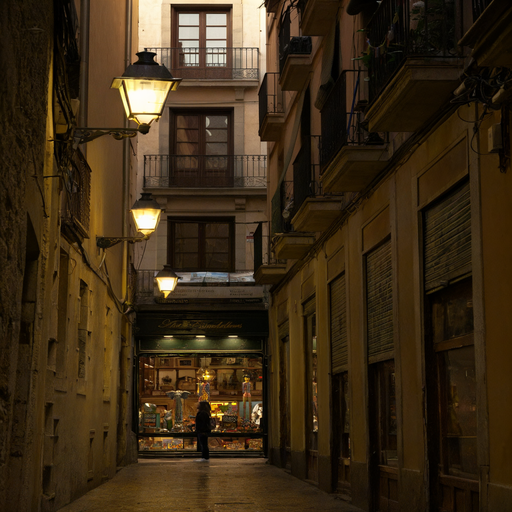} \\
    \includegraphics[width=\outdoorimwidth]{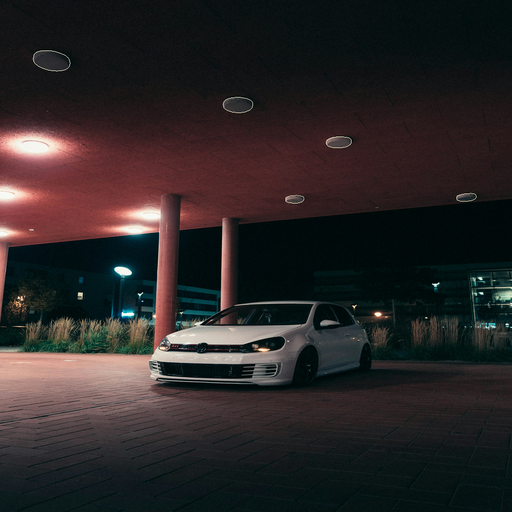} & \includegraphics[width=\outdoorimwidth]{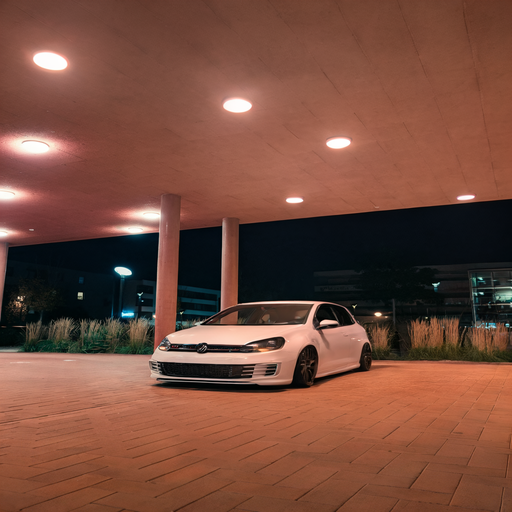} &
    \includegraphics[width=\outdoorimwidth]{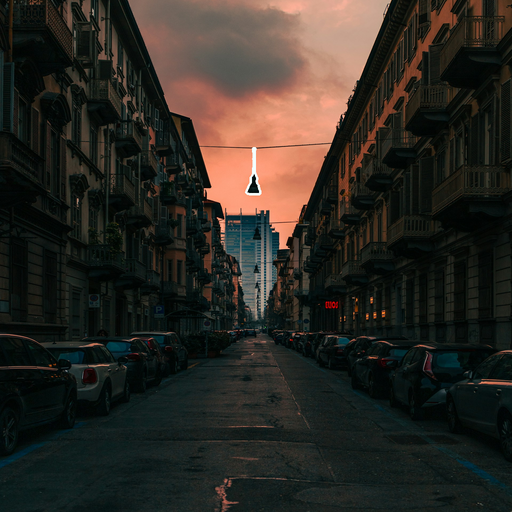} & \includegraphics[width=\outdoorimwidth]{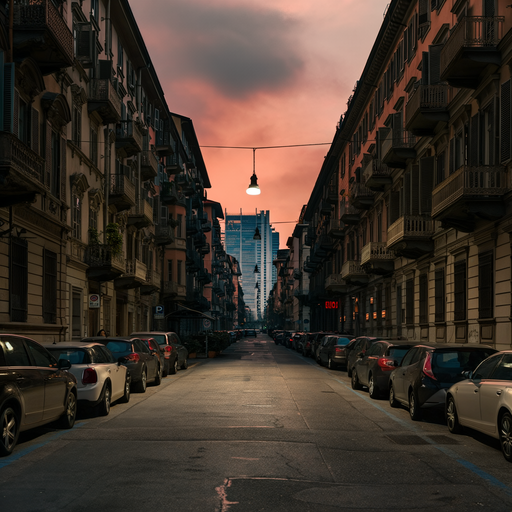} \\
    \includegraphics[width=\outdoorimwidth]{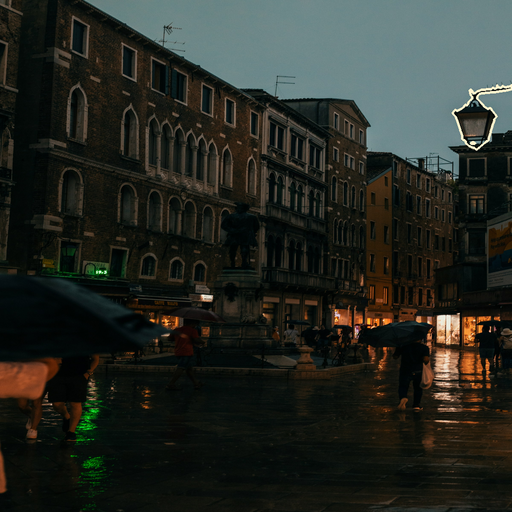} & \includegraphics[width=\outdoorimwidth]{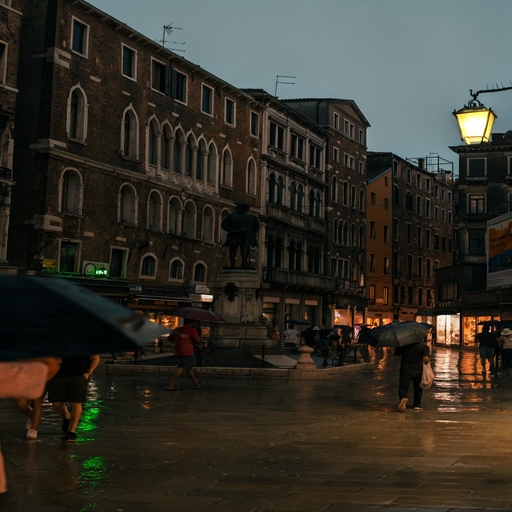} &
    \includegraphics[width=\outdoorimwidth]{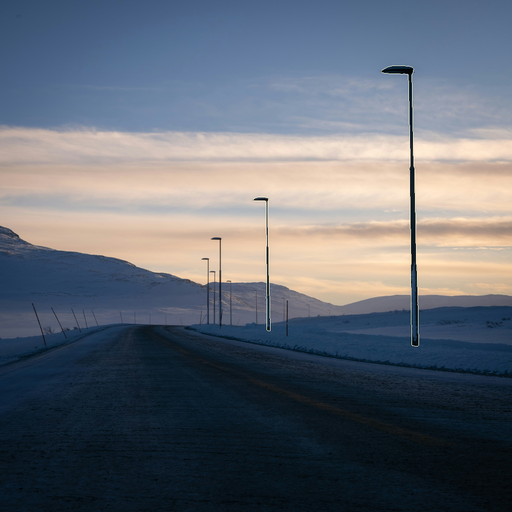} & \includegraphics[width=\outdoorimwidth]{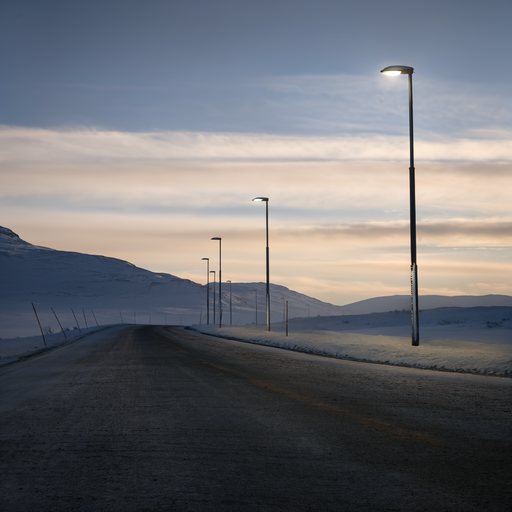} \\
    \includegraphics[width=\outdoorimwidth]{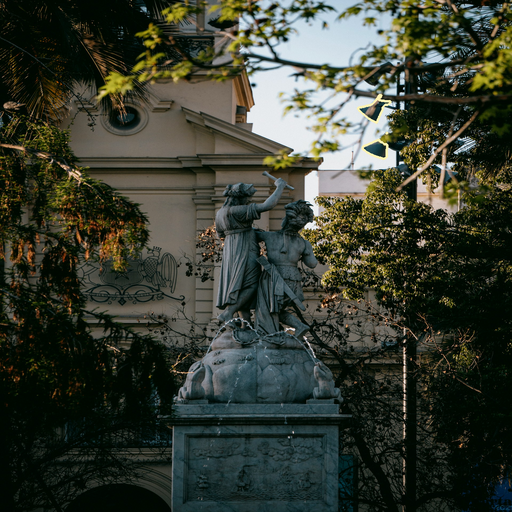} & \includegraphics[width=\outdoorimwidth]{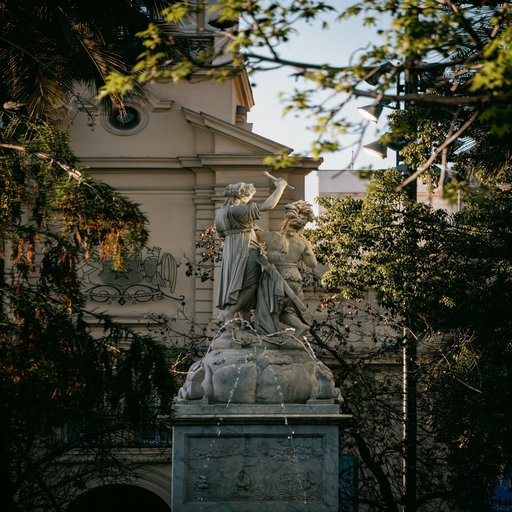} &
    \includegraphics[width=\outdoorimwidth]{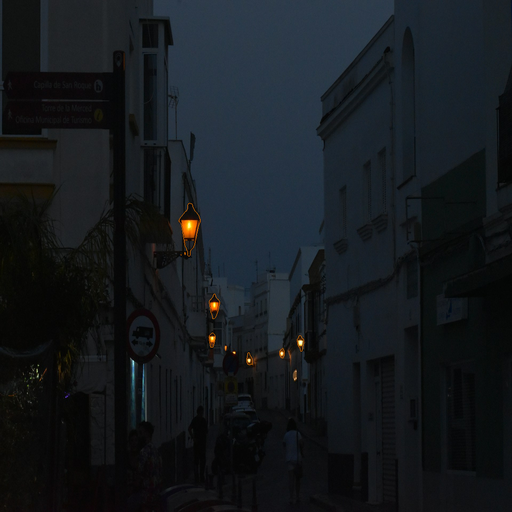} & \includegraphics[width=\outdoorimwidth]{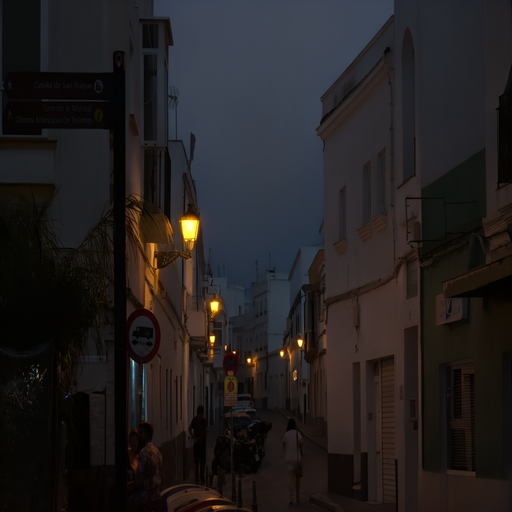} \\
\end{tabular}
\caption{\hl{\textbf{Results on Outdoor Scenes}: Input images show the contour of the target light source mask. Contour color represents the color condition given to the model. \textcolor{green}{green} contours specify negative intensity changes (turning off a light).
The input images were selected to represent of different environmental lighting conditions and different outdoor light sources. The results were generated with the tone-map condition "together", with no ambient change (0 value).
}}
\label{fig:outdoor}
\end{figure*}

\newcommand{\addintimwidth}{0.24\textwidth}

\begin{figure*}[t!]
\centering
\setlength{\tabcolsep}{1.5pt}
\begin{tabular}{c c c c}
    \textbf{Input} & \textbf{-0.34} & \textbf{-0.67} & \textbf{-1.00} \\
    \includegraphics[width=\addintimwidth]{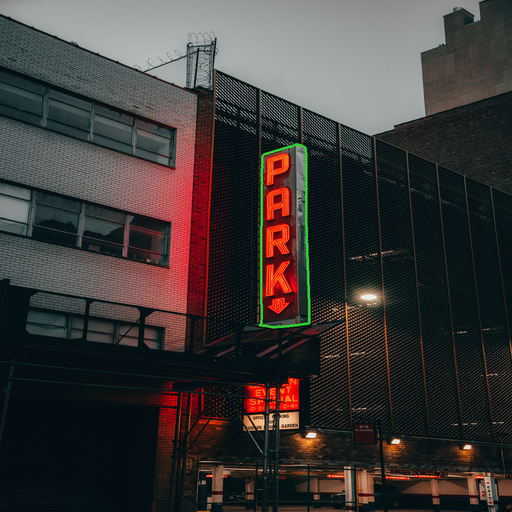} & \includegraphics[width=\addintimwidth]{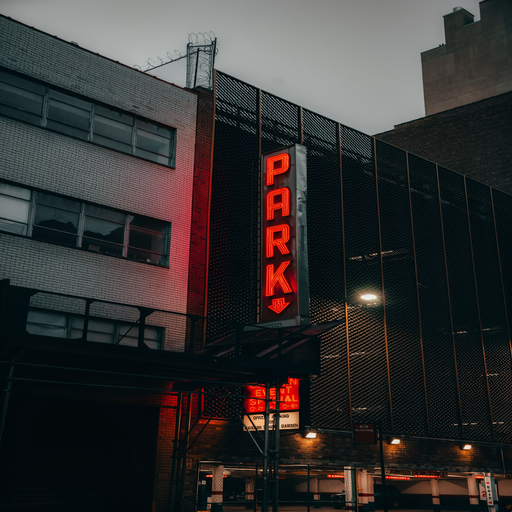} &
    \includegraphics[width=\addintimwidth]{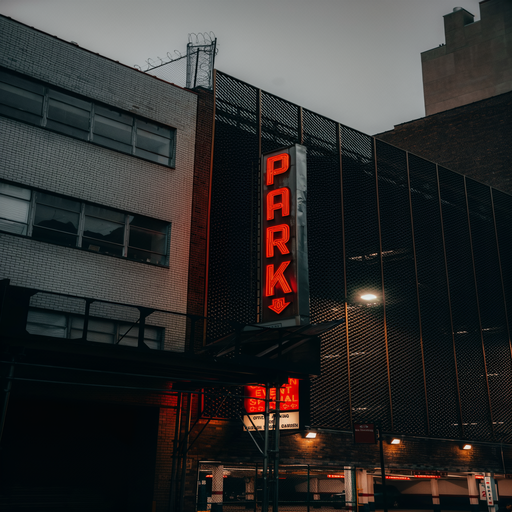} & \includegraphics[width=\addintimwidth]{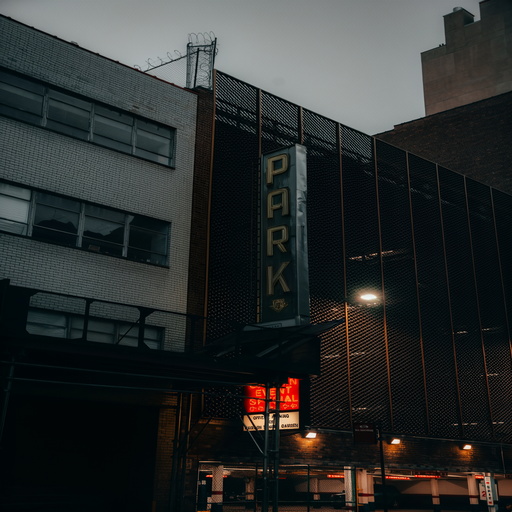} \\
    \midrule
    \textbf{Input} & \textbf{+0.20} & \textbf{+0.40} & \textbf{+0.60} \\
    \includegraphics[width=\addintimwidth]{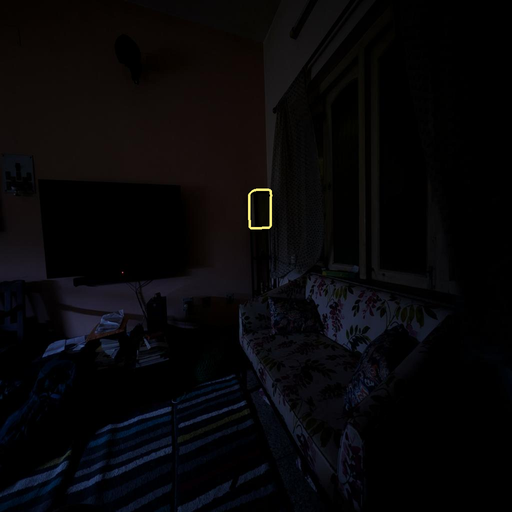} & \includegraphics[width=\addintimwidth]{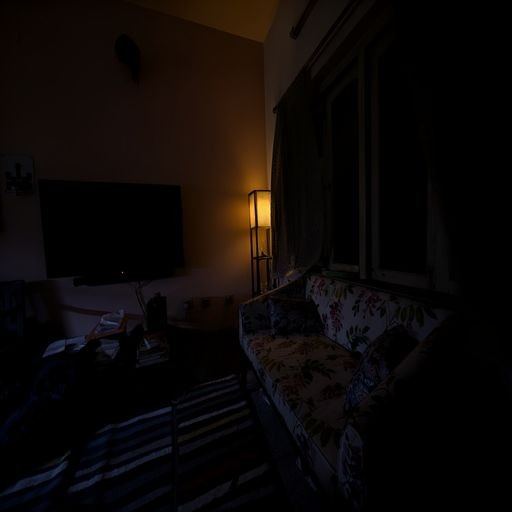} &
    \includegraphics[width=\addintimwidth]{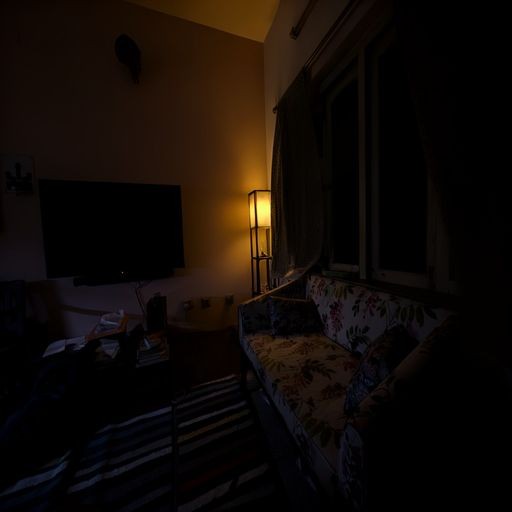} & \includegraphics[width=\addintimwidth]{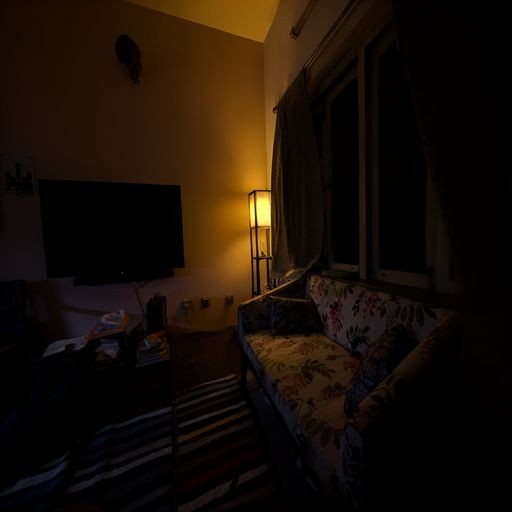} \\
    \textbf{+0.80} & \textbf{+1.00} & \textbf{+1.20} & \textbf{+1.40} \\
    \includegraphics[width=\addintimwidth]{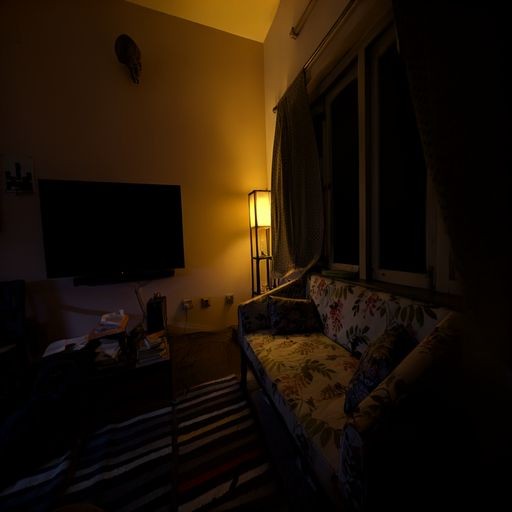} & \includegraphics[width=\addintimwidth]{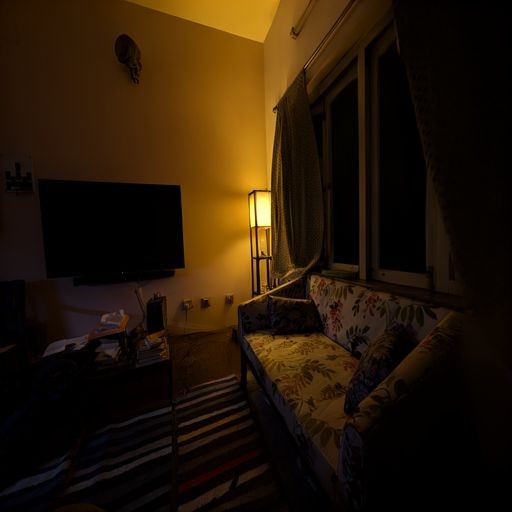} &
    \includegraphics[width=\addintimwidth]{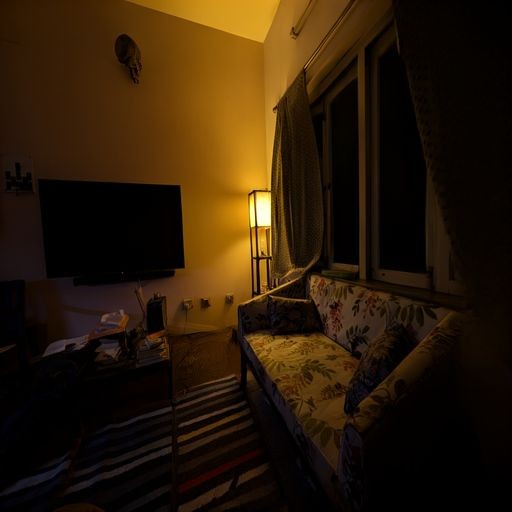} & \includegraphics[width=\addintimwidth]{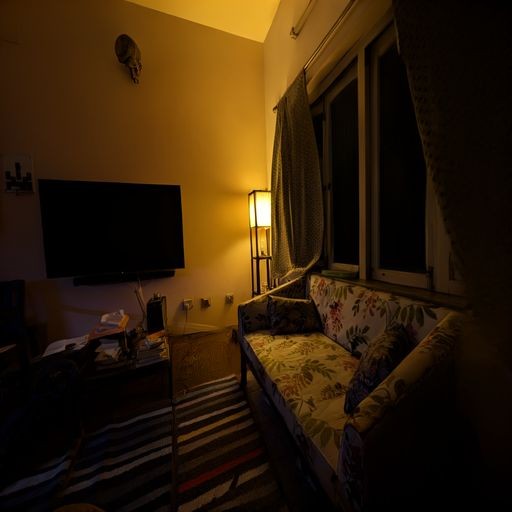} \\
    
\end{tabular}
\caption{\hl{\textbf{Effect of Tone-Mapping Condition}: The first row demonstrates gradual dimming of a colored neon light, which is not represented in the training dataset. Note how red light reflected from the adjacent brick-wall is being dimmed as-well. This figure also shows inaccuracies of the model, where red light on the lower section of the wall should not be removed. \textbf{Bottom Rows.} The bottom row shows that the model can extrapolate to intensity change values outside of the interval [-1,1].
The results were generated with the tone-map condition "together".
}}
\label{fig:additional_intensity}
\end{figure*}

\newcommand{\addambwidth}{0.24\textwidth}

\begin{figure*}[t!]
\centering
\setlength{\tabcolsep}{1.5pt}
\begin{tabular}{c c c c}
    \textbf{Input} & \textbf{+0.30} & \textbf{+0.60} & \textbf{+1.00} \\
    \includegraphics[width=\addambwidth]{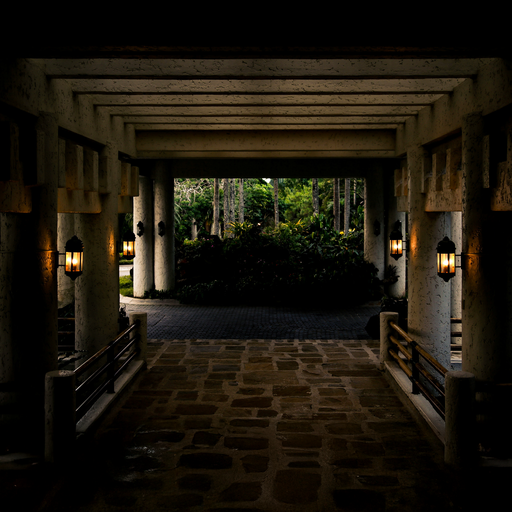} & \includegraphics[width=\addambwidth]{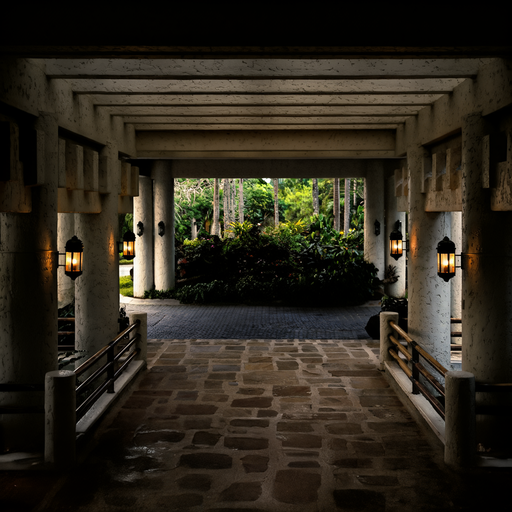} &
    \includegraphics[width=\addambwidth]{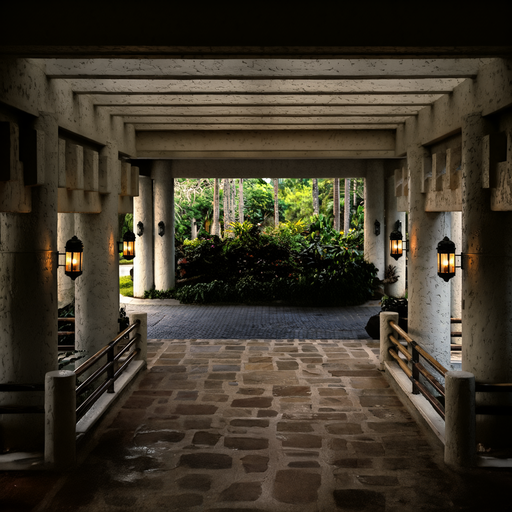} &
    \includegraphics[width=\addambwidth]{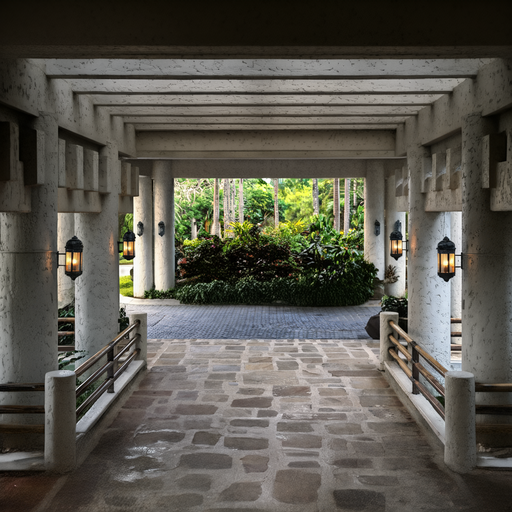} \\
    
    \textbf{Input} & \textbf{-0.40} & \textbf{-0.60} & \textbf{-1.00} \\
    \includegraphics[width=\addambwidth]{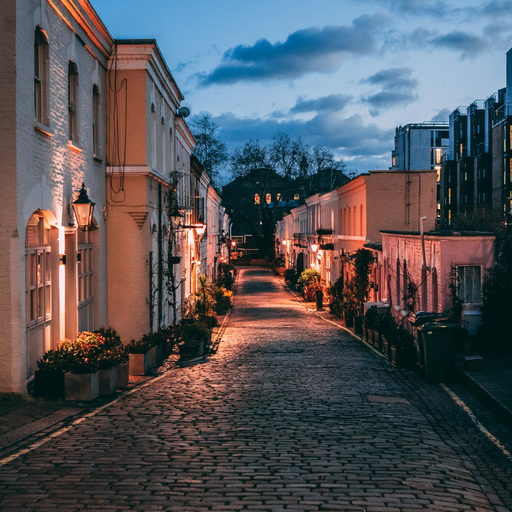} &
    \includegraphics[width=\addambwidth]{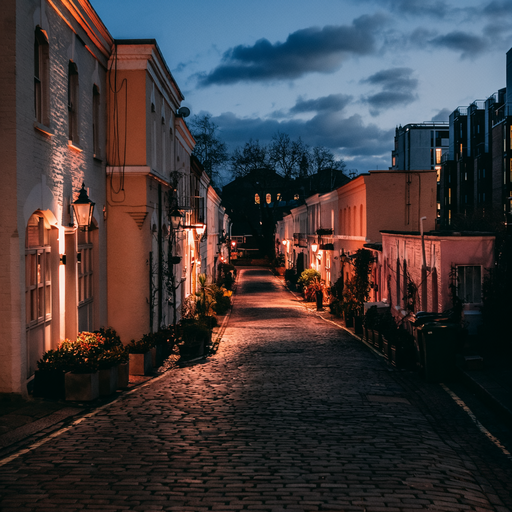} &
    \includegraphics[width=\addambwidth]{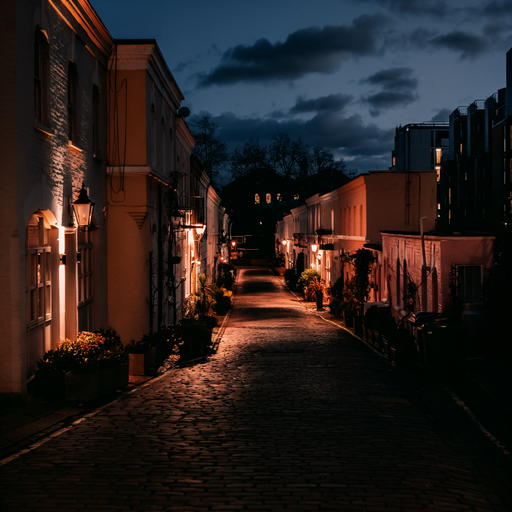} &
    \includegraphics[width=\addambwidth]{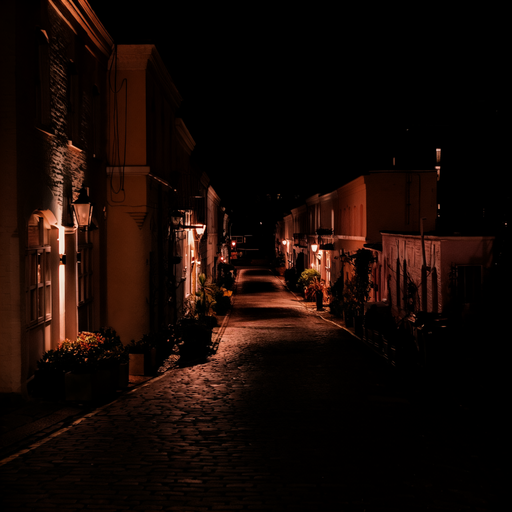} \\
    
    \textbf{Input} & \textbf{+0.30} & \textbf{+0.60} & \textbf{+1.00} \\
    \includegraphics[width=\addambwidth]{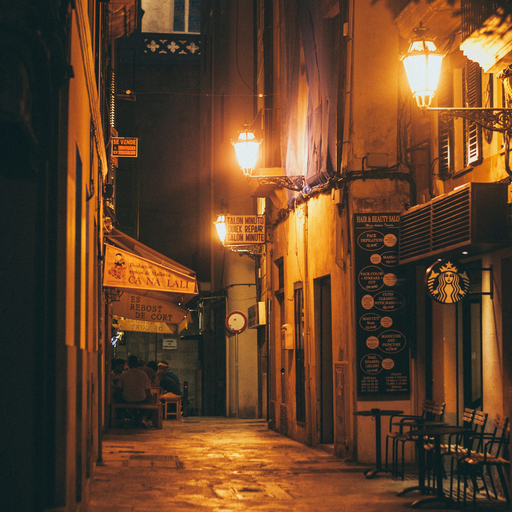} & \includegraphics[width=\addambwidth]{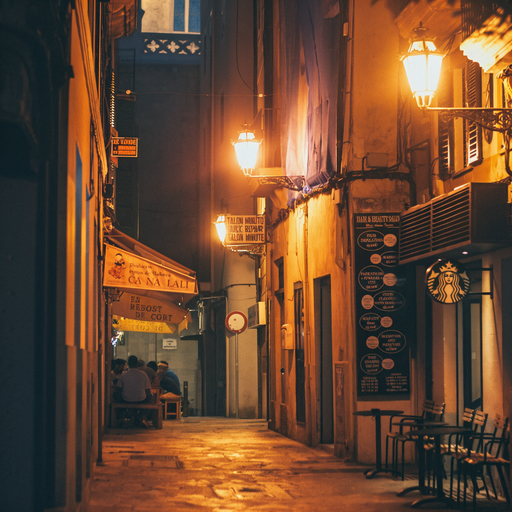} &
    \includegraphics[width=\addambwidth]{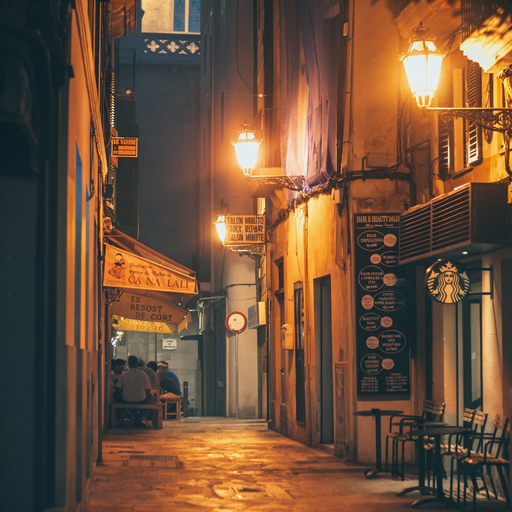} &
    \includegraphics[width=\addambwidth]{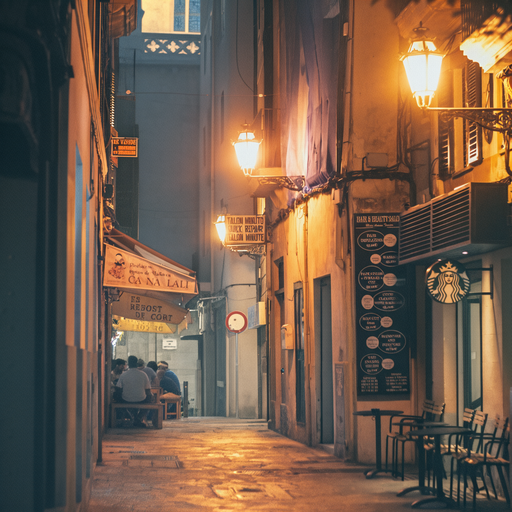} \\
    
    \textbf{Input} & \textbf{-0.30} & \textbf{-0.60} & \textbf{-1.00} \\
    \includegraphics[width=\addambwidth]{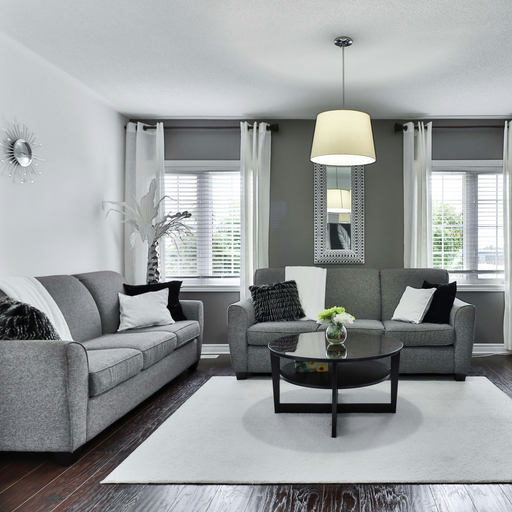} &
    \includegraphics[width=\addambwidth]{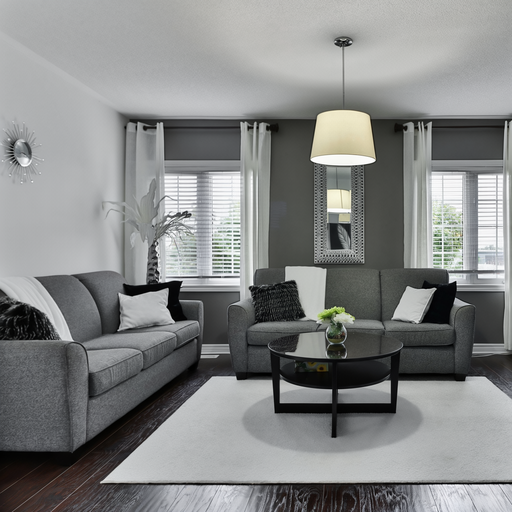} &
    \includegraphics[width=\addambwidth]{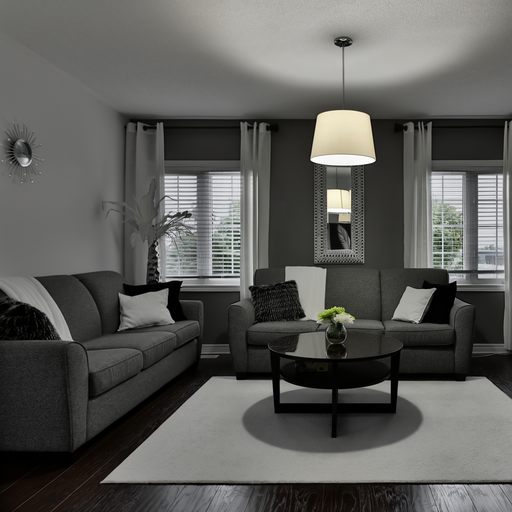} &
    \includegraphics[width=\addambwidth]{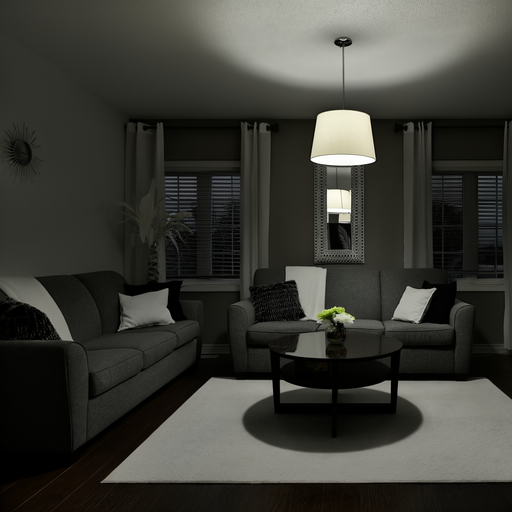} \\

\end{tabular}
\caption{\hl{\textbf{Additional Ambient Light Intensity Results}: We show ambient light edits in indoor and outdoor scenes.
In all the above images the tone-mapping condition was set to "together".
\textbf{Fourth row.} our method successfully disentangles daylight entering the room through windows from direct illumination emanating from the lamp.
\textbf{Second row.} When turning ambient light completely off the sky becomes too dark and no night sky illuminated is generated, this is a limitation arising from not using image pairs where ambient illumination changes, and from indoor bias in the training datasets.
The results were generated with the tone-mapping condition "together".
}}
\label{fig:additional_ambient}
\end{figure*}

\begin{figure*}%
    \centering
    \setlength{\tabcolsep}{1.5pt}
    \begin{tabular}{c c c c}
        
        \textbf{Input} & \textbf{0.33} & \textbf{0.67} & \textbf{1.00} \\
        \newlength{\tmrowheight}
        \settoheight{\tmrowheight}{\includegraphics[width=0.24\linewidth]{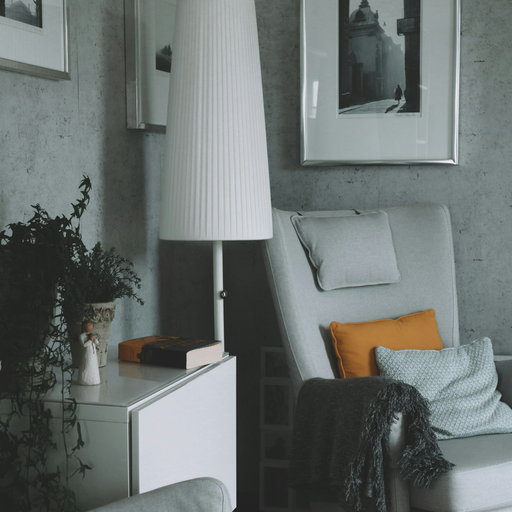}}
        \includegraphics[width=0.24\linewidth]{assets/images/intensity_tonemap_control/avechenri-NmqPnM4HJEc-unsplash/cond.png} &
        \includegraphics[width=0.24\linewidth]{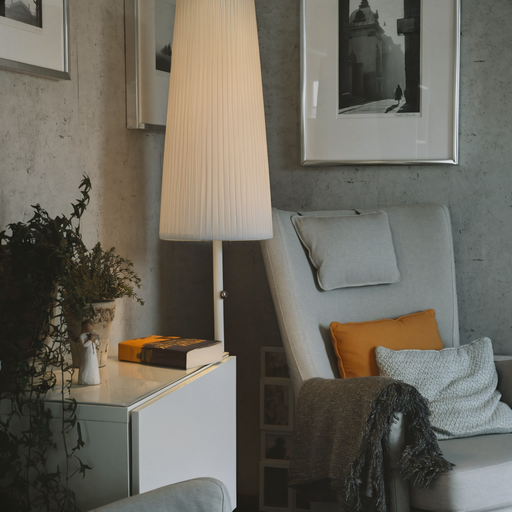} &
        \includegraphics[width=0.24\linewidth]{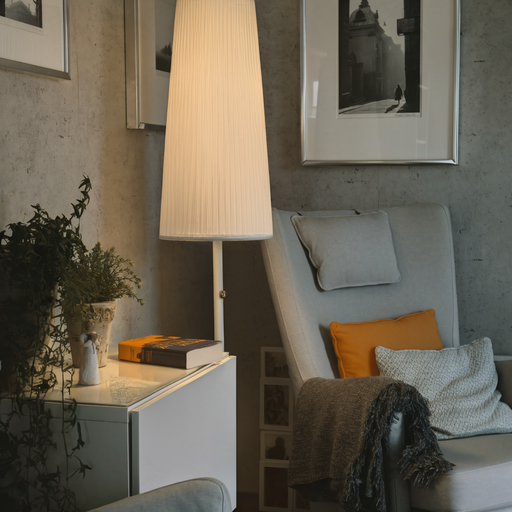} &
        \includegraphics[width=0.24\linewidth]{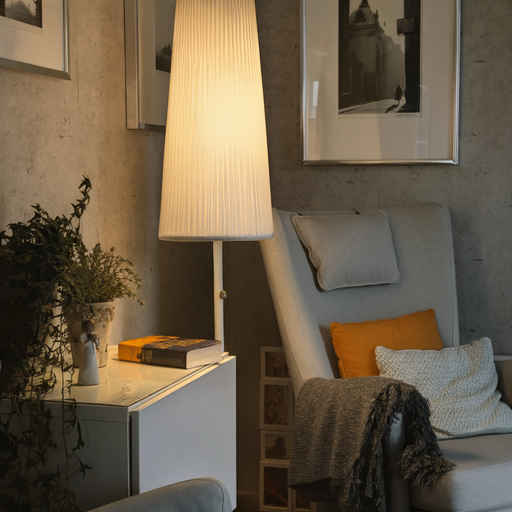} \\   
        
        \settoheight{\tmrowheight}{\includegraphics[width=0.24\linewidth]{assets/images/intensity_tonemap_control/avechenri-NmqPnM4HJEc-unsplash/cond.png}}
        \includegraphics[width=0.24\linewidth]{assets/images/intensity_tonemap_control/avechenri-NmqPnM4HJEc-unsplash/cond.png} &
        \includegraphics[width=0.24\linewidth]{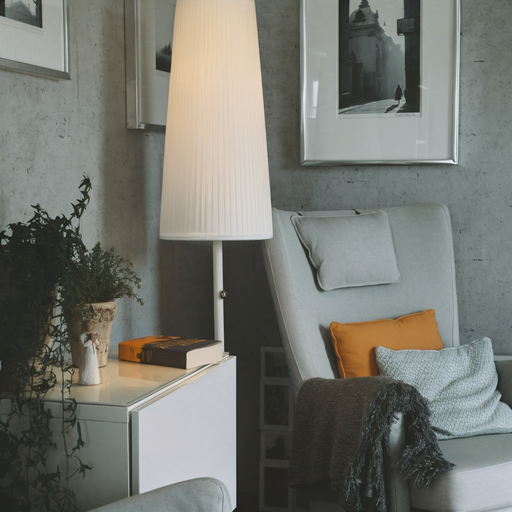} &
        \includegraphics[width=0.24\linewidth]{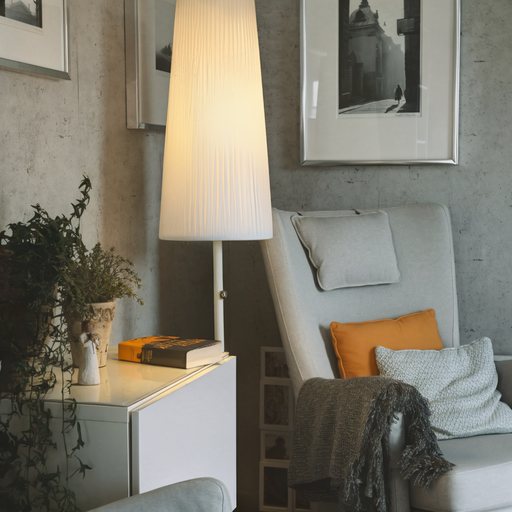} &
        \includegraphics[width=0.24\linewidth]{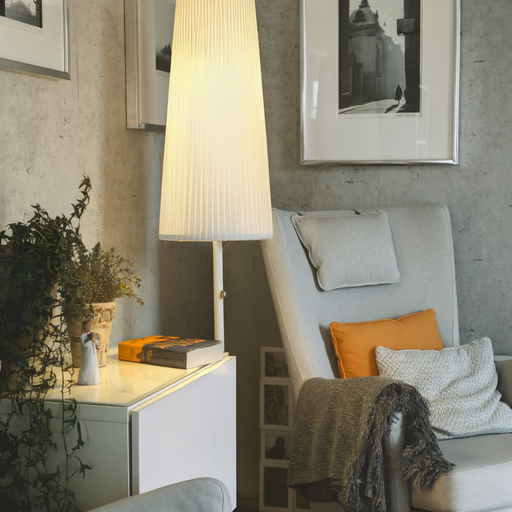} \\  

        \midrule
        
        \settoheight{\tmrowheight}{\includegraphics[width=0.24\linewidth]{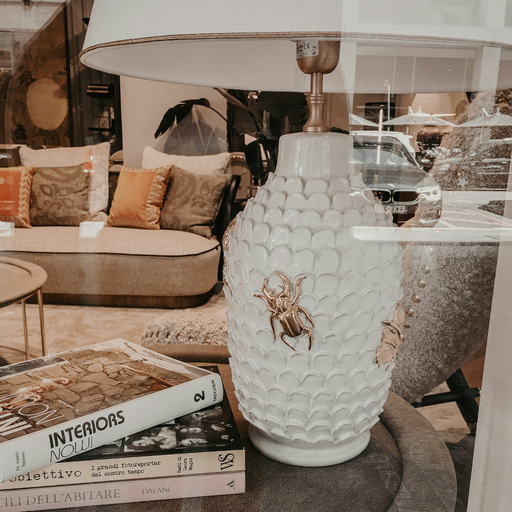}}
        \includegraphics[width=0.24\linewidth]{assets/images/intensity_tonemap_control/agata-ciosek-ffWlui8hytA-unsplash_intensity/cond.png} &
        \includegraphics[width=0.24\linewidth]{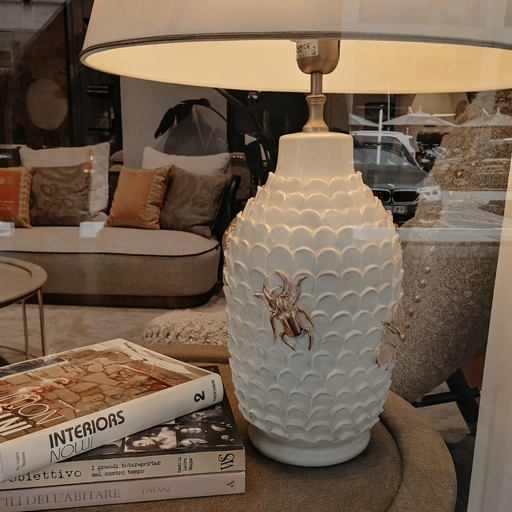} &
        \includegraphics[width=0.24\linewidth]{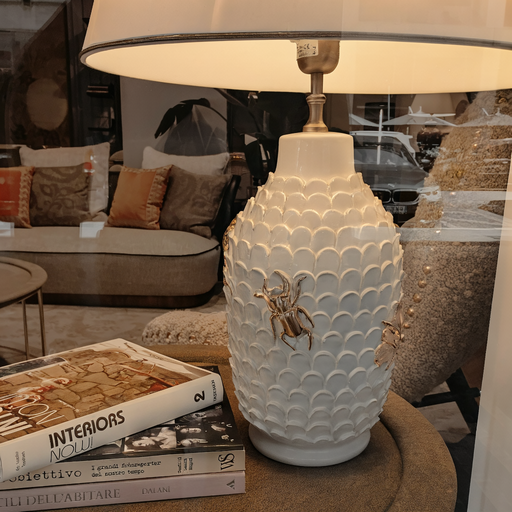} &
        \includegraphics[width=0.24\linewidth]{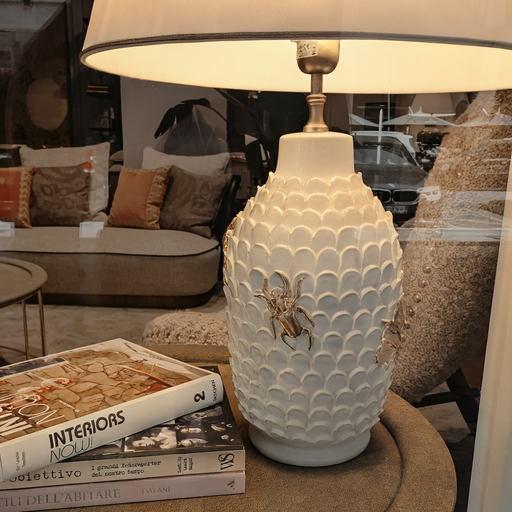} \\   
        
        \settoheight{\tmrowheight}{\includegraphics[width=0.24\linewidth]{assets/images/intensity_tonemap_control/agata-ciosek-ffWlui8hytA-unsplash_intensity/cond.png}}
        \includegraphics[width=0.24\linewidth]{assets/images/intensity_tonemap_control/agata-ciosek-ffWlui8hytA-unsplash_intensity/cond.png} &
        \includegraphics[width=0.24\linewidth]{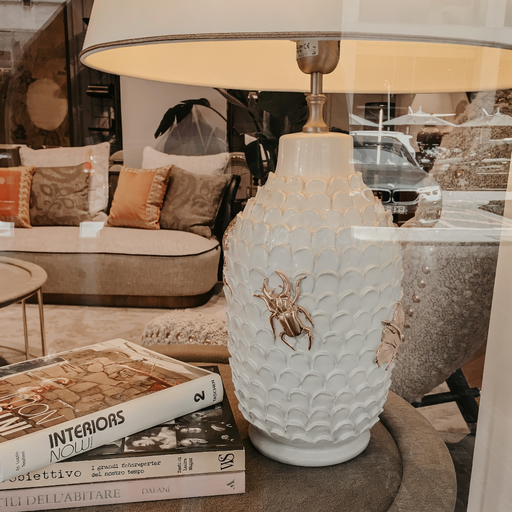} &
        \includegraphics[width=0.24\linewidth]{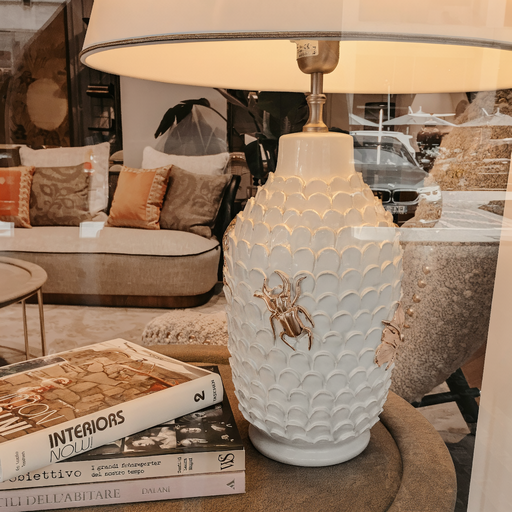} &
        \includegraphics[width=0.24\linewidth]{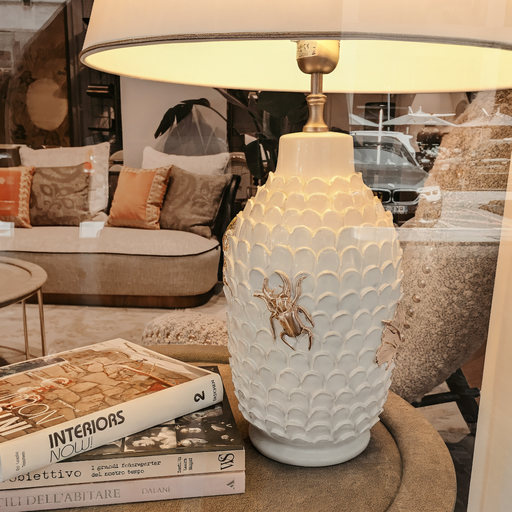} \\  
    \end{tabular}
    \caption{\hl{\textbf{Effect of Tone-Mapping Condition on Sequences.} This figure demonstrates the learned effect of the tone-mapping condition at inference time. It is complementary to Figure 4 in the main paper which shows the application of the two tone-mapping strategies on training data.
    Note that the learned tone-mapping control is not completely not disentangled, which could be seen by the perceptible variance in light intensity for the "separate" condition. \textbf{First and third rows.} the sequences tone-mapped separately. \textbf{Second and fourth rows.} the sequences tone-mapped together.
    }}
    \label{fig:tonemapping_intensity_inference}
\end{figure*}

\end{document}